\documentclass[jair,11pt]{article}
\usepackage{jair, theapa, rawfonts}
\usepackage{amssymb}
\usepackage{amsmath}
\usepackage{subcaption}
\usepackage{soul}
\usepackage{booktabs}
\usepackage{caption}
\usepackage{graphicx}
\usepackage{multirow}

\usepackage[ruled,vlined]{algorithm2e}
\usepackage{algorithmicx} 
\usepackage{threeparttable}
\usepackage{url}
\usepackage{ragged2e}

\usepackage[textsize=scriptsize,backgroundcolor=green]{todonotes}

\jairheading{78}{2023}{901-973}{11/2022}{12/2023}
\ShortHeadings{Aggregating Simple, Complex, and Multi-Object Annotations}
{Braylan, Marabella, Alonso, \& Lease}
\firstpageno{901} 

\begin{document}

\title{A General Model for Aggregating Annotations Across Simple, Complex, and Multi-Object Annotation Tasks}


\author{\name Alexander Braylan \email braylan@cs.utexas.edu \\
      \addr Dept.\ of Computer Science\\
            University of Texas at Austin
      \AND
      \name Madalyn Marabella 
      \email mmarabella@utexas.edu \\
      \addr McCombs School of Business\\
            University of Texas at Austin
      \AND
      \name Omar Alonso \email omralon@amazon.com \\
      \addr Amazon
      \AND
      \name Matthew Lease \email ml@utexas.edu \\
      \addr School of Information\\
      University of Texas at Austin
}


\maketitle

\begin{abstract}

Human annotations are vital to supervised learning, yet annotators often disagree on the correct label, especially as annotation tasks increase in complexity. A common strategy to improve label quality is to ask multiple annotators to label the same item and then aggregate their labels.
To date, many aggregation models have been proposed for simple categorical or numerical annotation tasks, but far less work has considered more complex annotation tasks, such as those involving open-ended, multivariate, or structured responses.
Similarly, while a variety of bespoke models have been proposed for specific tasks, our work is the first we are aware of to introduce aggregation methods that generalize across many, diverse complex tasks, including sequence labeling, translation, syntactic parsing, ranking, bounding boxes, and keypoints. This generality is achieved by applying readily available task-specific distance functions, then devising a task-agnostic method to model these distances between labels, rather than the labels themselves.

This article presents a unified treatment of our prior work on complex annotation modeling and extends that work with investigation of three new research questions. First, how do complex annotation task and dataset properties impact aggregation accuracy? Second, how should a task owner navigate the many modeling choices in order to maximize aggregation accuracy? Finally, what tests and diagnoses can verify that aggregation models are specified correctly for the given data? To understand how various factors impact accuracy and to inform model selection, we conduct large-scale simulation studies and broad experiments on real, complex datasets. Regarding testing, we introduce the concept of unit tests for aggregation models and present a suite of such tests to ensure that a given model is not mis-specified and exhibits expected behavior.

Beyond investigating these research questions above, we discuss the foundational concept and nature of annotation complexity, present a new aggregation model as a conceptual bridge between traditional models and our own, and contribute a new general semi-supervised learning method for complex label aggregation that outperforms prior work.

\end{abstract}

\section{Introduction}
\label{Introduction}

Human annotations provide the foundation for training and testing supervised learning models. Because label quality can greatly impact both the accuracy of a trained model and model evaluation \cite{aroyo2022data,northcutt2021pervasive}, many models and measures of annotator behavior and labels have been proposed \cite{dawid1979maximum,smyth1995inferring,artstein2008inter,kim2010improving,passonneau2014benefits}. The advent of crowd annotation \cite{su2007internet,snow2008cheap} has further stimulated a surge of modeling techniques for quality assurance with inexpert annotators \cite{sheng2008get,Sheshadri13,zheng2017truth}. However, existing annotation models typically assume relatively {\em simple} labeling tasks with small answer spaces, such as classification or rating tasks. When the answer space is small, responses can be compared by exact-match and aggregation often reduces to selecting whichever response received the most (weighted or unweighted) votes.

\begin{figure}
\centering
  \fbox{\includegraphics[scale=.7]{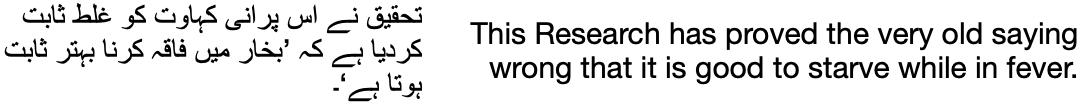}}
  \fbox{\includegraphics[scale=1,height=8.5em]{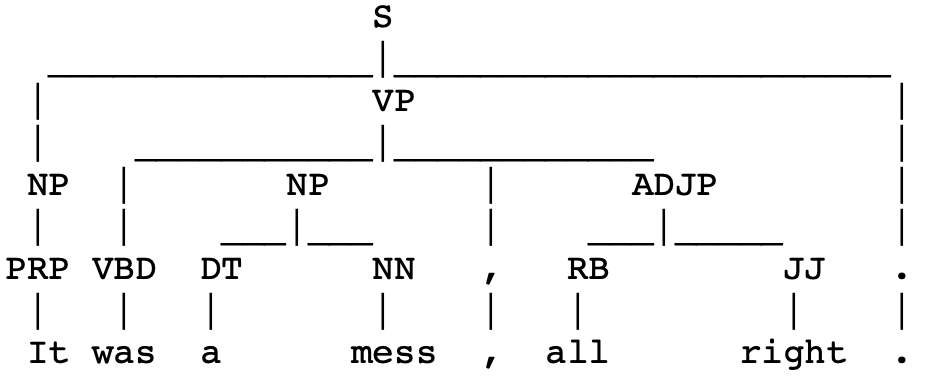}}
  \fbox{\includegraphics[scale=.8]{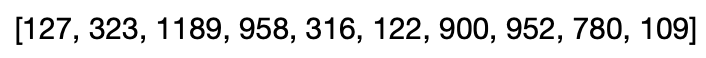}}
\caption{Examples of complex annotations: a translation between two human languages; a syntactic parse tree; and ranked elements for an item (for example, top ten most relevant document identifiers). Unlike simple annotations, these have much larger spaces of possible response values compared to typical categorical or ordinal labeling tasks.}
\label{fig:examples-selection}
\end{figure}

\begin{figure}
\centering
\includegraphics[scale=1]{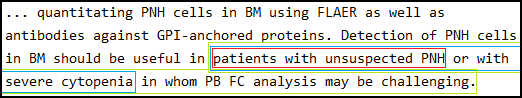}
\includegraphics[scale=0.5]{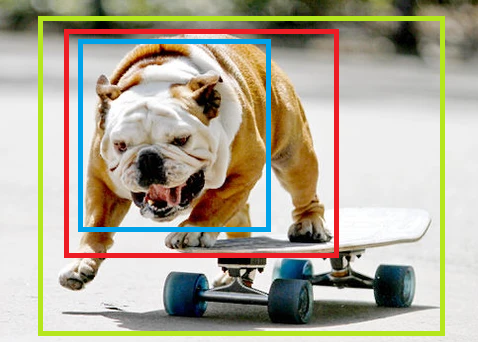}
\includegraphics[scale=0.5]{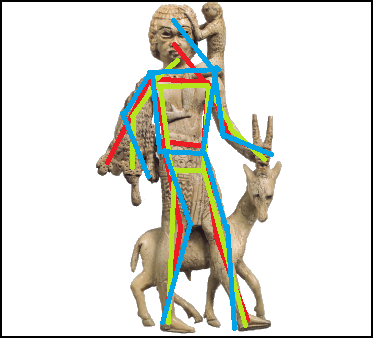}
\caption{Annotator disagreement is shown via different colors for three labeling tasks: text sequences (top) and image bounding boxes (bottom left) and keypoints (bottom right). How should we aggregate annotations to induce consensus?} 
\label{fig:examples-merge}
\end{figure}

\begin{figure}
\centering
\includegraphics[scale=1]{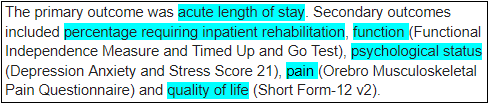}
\includegraphics[scale=0.6]{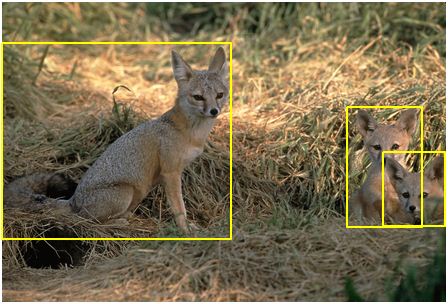}
\includegraphics[scale=0.4]{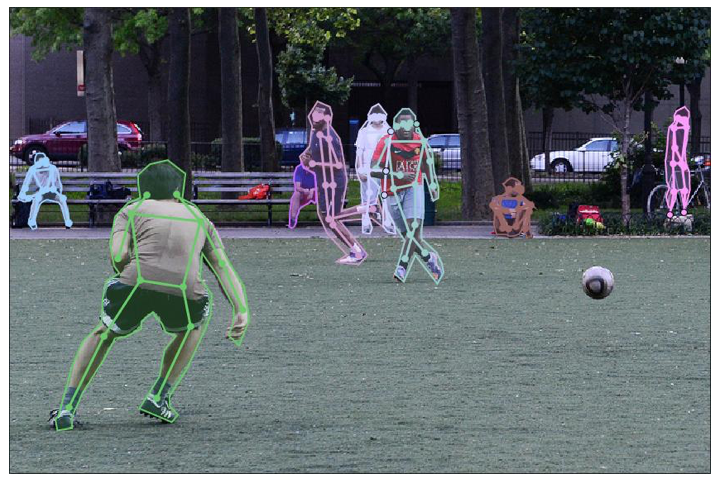}
\caption{Three multi-object annotation tasks show labels from a single annotator only: at the top, text sequences are multiple text span \emph{objects} in a document \emph{item}. On the bottom left, bounding boxes are multiple rectangle \emph{objects} in an image \emph{item}. On the bottom right, keypoints are multiple skeletal structure \emph{objects} in an image \emph{item}.} 
\label{fig:examples-partition}
\end{figure}

Unfortunately, as we move toward larger labeling spaces (e.g., continuous, structured, or high-dimensional annotation tasks), it becomes increasingly rare that any annotators agree on exactly the same label.
Such \emph{complex} annotation tasks are the motivation for this work  
and may involve open-ended answer spaces (e.g., translation, transcription, extraction) or structured responses (e.g., annotating ranked lists, linguistic syntax or co-reference), such as the examples shown in {\bf Figures~\ref{fig:examples-selection}-\ref{fig:examples-merge}}. 
In addition, while simple labeling tasks like classification may require only a single label for each input {\em item} (e.g., a document or image), some complex annotation tasks require annotators to label multiple {\em objects} per item — such as demarcating named-entities in a text \cite{sang2003introduction} or visual entities in an image \cite{branson2017lean} (see {\bf Figure~\ref{fig:examples-partition}}).


This much larger response space for complex annotations has several important consequences. First, we need to be able to compare labels on the basis of partial-credit rather than exact-match. Second, rather than to merely {\em select} the best annotator label available, we may want to {\em merge} annotator labels in some fashion in order to generate a better combined label than any individual annotator produced. Though the classic example of the ``Wisdom of Crowds'' \cite{surowiecki2005wisdom} is merging individual guesses for the weight of an ox, research on label aggregation \cite{dawid1979maximum,sheng2008get,zheng2017truth} has devoted relatively little attention to merging due to a predominant focus on categorical tasks wherein categories cannot be merged and aggregation reduces to selection.


\vspace{1em} 
\noindent{\bfseries Contributions}
\vspace{1em} 

Our broad goal is to provide a unified, general framework for modeling and aggregating annotations across diverse, complex annotation tasks. While our primary focus clearly rests on complex annotations, we also show generalization of our models across simple annotation tasks as well. In this respect, we do not seek to further improve upon the accuracy of prior aggregation models for simple tasks, but rather: 1) to demonstrate conceptual generality of our techniques across both simple and complex annotation tasks; and 2) to offer practitioners with a one-stop solution for both simple and complex annotation tasks, rather than having to develop and maintain different aggregation models for each type of task. We advocate that researchers and practioners first start with our general, open-source model, then if greater accuracy is needed, only then pursue development of task-specific, bespoke models.

In our initial work \cite{braylan2020modeling}, we showed that diverse complex annotations could be modeled via annotation distances using existing evaluation metrics for each task, and we presented several models for identifying the best annotator label for each item. This work tackled ranked lists, syntactic parsing, text sequences and translations. We presented the first aggregation model for ranked list, tree-structure, and free-form text data, which also generalized to text sequences with performance on par with bespoke aggregation models. This work also presented the first semi-supervised aggregation model for any of these complex annotation tasks. Following this, our subsequent work \cite{braylan2021mergeandmatch} further investigated aggregation by merging and aggregation for multi-object annotation tasks (e.g., named-entities and bounding boxes). This work also presented the first aggregation model (we are aware of) for multiple-keypoint data.

In this latest article, we both present unified treatment of our earlier works and investigate the following new {\bf research questions (RQs)}:
\begin{enumerate}
    \item How do complex annotation task and dataset properties impact aggregation accuracy?

	\begin{enumerate}
	    \item Impact of task type? \label{RQ:tasktype}
    	\item Impact of data sparsity?
    	\label{RQ:sparsity}
    	\item Impact of variance and skew in annotator accuracy?
    	\label{RQ:distribution}
	\end{enumerate}

    \item Given a broad arsenal of modeling decisions available, how should a user navigate these choices in order to maximize aggregation accuracy?

	\begin{enumerate}
    	\item When is weighted aggregation better than unweighted aggregation?
    	\label{RQ:wgtvsnot}
    	\item When is it most useful to use semi-supervised learning?
    	\label{RQ:semisup}
	\item When is greater model complexity needed or can simpler approaches suffice?
    	\label{RQ:masvsmadd}
	\item When are bespoke, task-specific models justified over general models?
    	\label{RQ:bespoke}
	\end{enumerate}
	
    \item How can we test and diagnose whether an aggregation model is behaving as expected?
    \label{RQ:modelchecking}

\end{enumerate}
In investigating the research questions above, we make the following new contributions.

{\bf We provide a more tangible definition of annotation complexity} compared to our prior work and use it to characterize various diverse annotation tasks. This definition is detailed in Section \ref{sec:what-is-complex} and helps distinguish between task types for answering RQ\ref{RQ:tasktype}.

{\bf We propose a new model of annotation distances}, a Model of Ability, Difficulty, and Distance (MADD). Introduced in Section~\ref{method:madd}, MADD bridges a conceptual, practical, and empirical gap between models from prior literature on simple annotations and our Multidimensional Annotation Scaling (MAS) model (Section~\ref{method:mas}). While still modeling the same annotation distance data and using similar parameters for annotator ability and item difficulty, MADD makes a simplifying assumption about the relationship between ground truth and the annotation response space. In doing so, MADD helps answer RQ\ref{RQ:masvsmadd} by illuminating what aspects of model design and complexity are most important in effectively modeling complex annotations. We find that MADD occupies a space in the spectrum between simple models and MAS, with overall weaker performance that thereby reveals the benefit of MAS's additional complexity.

{\bf We present a novel semi-supervised learning approach}, much more general than our earlier method \cite{braylan2020modeling}, that can be applied to any annotator-weighted aggregation model, described in Section \ref{method:semisupervised}. We also extend testing of semi-supervised learning to newer complex datasets introduced here in addition to datasets from our earlier work. Addressing RQ\ref{RQ:semisup}, our results show that this semi-supervised learning approach is useful in general, but especially in cases wherein data sparsity and the worker error distribution make it more difficult to effectively model worker ability. Moreover, we show in Section \ref{sec:results-real} that this new approach outperforms the earlier semi-supervised model from \citeA{braylan2020modeling}.

{\bf We contribute a suite of unit tests for aggregation models} to enable researchers and practitioners to verify correctness of deployed models in meeting expected criteria of behavior and performance. These tests ensure desired asymptotic behavior, including correct weighting of workers and correct application of weightings in vote, as well as desired behavior in data scarcity scenarios, such as shrinkage towards majority vote. These address RQ\ref{RQ:modelchecking} and are discussed in Section \ref{sec:unittests}.

{\bf We conduct extensive simulation studies across varying tasks, crowd compositions, and modeling configurations} in order to better understand when and why different modeling decisions matter. We significantly extend simulators from our prior work in order to better control and configure key variations, such as the distribution of annotator ability. These results build on prior theoretical and empirical results on simple annotations to yield new insights specific to complex annotations.
These simulation experiments help answer our questions about the effects of data sparsity and worker error distributions, as well as helping understand when weighted aggregation is more useful than unweighted. The number of labels per item and ratio of high to low quality workers significantly influence findings in relation to the benefit (or lack thereof) from weighted voting.
The results of these simulation experiments (in Section \ref{sec:results-sim}) help answer RQs \ref{RQ:tasktype}, \ref{RQ:sparsity}, \ref{RQ:distribution}, and \ref{RQ:wgtvsnot}.

{\bf We benchmark across a far wider variety of datasets and baseline methods}, including simple categorical and numerical datasets from prior literature. While novel methods garner the bulk of attention in advancing the state-of-the-art, benchmarking also plays an essential yet often under-stated role in a field's progress. In a computational biology keynote, \citeA{tse2012} argued that a field’s progress is often driven not
by new algorithms, but by well-defined challenge problems
and metrics which drive innovation and enable comparative
evaluation. Benchmarking also plays a vital role in promoting conceptual and empirical comparison of prior art that is often left disconnected otherwise \cite{zheng2017truth,Sheshadri13,nguyen2013batc}.
Our benchmarking results (Section \ref{sec:results-real}) help to answer RQ\ref{RQ:tasktype} and RQ\ref{RQ:bespoke}, and also enable us to assess the generality and suitability of our approach as a one-stop shop for supporting diverse crowdsourcing tasks. Surprisingly, we find our general approach to be competitive with or outperforming task-specific, bespoke models in empirical evaluations on real datasets.



{\bf Reproducibility.} To promote reproducibility, follow-on research, and practical adoption, we share our code and data with a permissive BSD-3 license\footnote{\url{https://github.com/Praznat/annotationmodeling}}. In comparison to our prior codebase and data, we add code updating the performance of MAS, the MADD model, the new method for semi-supervision, functions to help with unit testing, new simulators and simulation experiment managers, and data for the new datasets.

\begin{figure}
\centering
\includegraphics[scale=0.5]{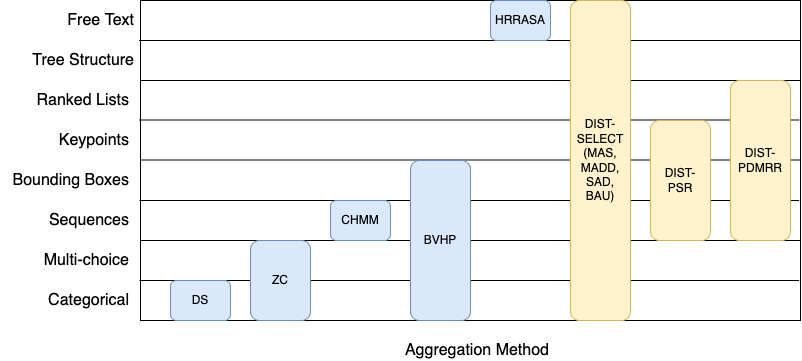}
\caption{How are our methods (yellow) positioned relative to prior work (blue)? The main differentiator is generality across diverse annotation tasks.}
\label{fig:positioning}
\end{figure}

\section{Problem Space and Related Work}
\label{sec:background}

{\bf Label aggregation} is the task of inferring the correct label for a given item from a set of multiple annotations for that item. Typically this task is operationalized as selecting the best label for the item from the set of available annotations, although it can sometimes include label merging as in the averaging of numeric values \cite{zheng2017truth}.
This assumption of there being a best label -- and more generally that some labels are better than others -- is a convenient simplification to leverage the expectation that errors from independent, unbiased, uncorrelated annotators will tend to cancel out.
Given these assumptions and requirements, some soft guidelines for when to aggregate or not are as follows:
\begin{enumerate}
\item Annotators should have minimal systematic bias. In cases where such a bias is present, aggregating labels might actually amplify the bias rather than reduce it \cite{ipeirotis2010quality}.
\item Task should instructions have minimal ambiguity. Task goals and instructions may often be somewhat ambiguous \cite{paun2022statistical,wu2017confusing,gadiraju2017clarity}. Sometimes the ambiguity is intentional, as in the case of more subjective tasks such as those involving humor or political discourse. When ambiguity is unintentional, however, it can lead to higher errors in the sense of labels diverging from what the requester intended as the best label.
\item The task should be more objective than subjective. Having a large space of equally valid responses can be a reason not to use aggregation. Aggregation will discard all but the one label deemed best, so if there is value in maintaining much varied responses, then aggregation could be inappropriate \cite{noble2012minority,tian2012learning,gordon2022jury}.
\end{enumerate}
While these soft guidelines offer a useful starting point, the decision of whether to aggregate can depend on a variety of task-specific considerations that must ultimately be left to the requester's discretion.

{\em Offline, static} aggregation assumes all the annotations have already been collected at the time of aggregation, as opposed to an online approach which dynamically controls the items to annotate during the aggregation process. Research on the offline task has been especially vibrant in the human computation field, with many models proposed and benchmarked \cite{Sheshadri13,hung2013evaluation,zheng2017truth}. Prior work on general-purpose aggregation models has typically assumed simple annotations in which a small label space permits evaluating annotator performance based on exact match vs.\ gold or peer labels. While this paper continues the tradition of research into offline methods for selecting a best available label per item, it expands from prior work away from the assumptions of simple annotations with a small label space.

The benefits of annotation aggregation models are well-known for both traditional and crowd annotation settings \cite{passonneau2014benefits}. In this work, we assume annotation quality is measurable and items to annotate may vary in difficulty.
Factors such as task design, annotator instructions \cite{wu2017confusing}, and spam control mechanisms \cite{parmar2022don} can also influence the quality of collected annotations \cite{welty2019metrology}. 
More complex annotation tasks may be more affected by these factors, revealing greater variability in annotator abilities, and resulting in a wide range of different but satisfactory annotations for the same item. This suggests that complex annotation tasks may stand to benefit more from annotator modeling than is the case with more common, simpler annotation tasks.

\subsection{Aggregation Methods and Models for Categorical Annotation Tasks}
\label{sec:aggregation}

As noted above, a variety of general-purpose and task-independent aggregation models exist for simple annotation tasks. We briefly discuss a few representative approaches. 

\subsubsection{Unsupervised Aggregation Models}

{\bf Majority Voting (MV)} is the simplest aggregation approach and avoids any modeling of workers or task.  When most workers are accurate and have comparable accuracy, it can work quite well, and its being task-agnostic makes it quite versatile across diverse annotation tasks. However, as an unweighted voting method, it can perform poorly when the majority of workers are inaccurate, or when there is high variance in worker accuracy that is not modeled. It also assumes a sufficiently small label space such that some workers will produce the same label, and thus a majority label can be found.

{\bf \citeA{dawid1979maximum} (DS)} proposed the now ``classic'' approach to simultaneously inferring annotator reliability and label quality (i.e. how often labels match ground truth). Their unsupervised method was based on measuring peer agreement between annotators (i.e., a popularity contest for labels) to infer annotator reliability and label quality. DS was one of the earliest uses of the EM algorithm \cite{dempster1977maximum}. Despite its relative age and simplicity, prior benchmarking studies \cite{Sheshadri13,hung2013evaluation,zheng2017truth} found DS to be a top-performer across datasets.

Both DS and later \citeA{snow2008cheap} assume category-based annotation, modeling each annotator by a confusion matrix for their probability of producing a given categorical label given the gold categorical label. Note that this approach cannot be directly applied to non-classification tasks, such as multiple-choice selection in which there is not a fixed set of categories to model across items. Also, for $K$ categories we must estimate a $K\times K$ confusion matrix, which becomes more problematic for space and sparsity as $K$ increases \cite{liu2012truelabel}, such as with complex annotation tasks.

\textbf{ZenCrowd} (ZC) \cite{demartini2012zencrowd} can be interpreted as a simplified variant of DS (though was proposed without reference to DS). Rather than representing each worker by a confusion matrix, each worker is instead modeled by a single reliability parameter. Similar unsupervised estimation is performed via EM. ZC is representative of a larger family of models having a single parameter for each annotator \cite{zheng2017truth}, thus applicable beyond categorical annotation tasks and less prone to sparsity. 

All of the approaches discussed so far assume we can measure peer agreement via exact-match between labels. As $K$ grows, it is less likely annotators will produce the same label for a given item, and so exact-match becomes a harsher 0/1 loss function for testing assessor reliability. Some numerical task label aggregation methods do not require exact match \cite{li2014confidence,li2020gpm}.

\subsubsection{Semi-supervised Aggregation Models}
\label{background:semisup}

The methods described above are all unsupervised in that they can be trained without knowing gold annotations for any items. However, it is often the case that task owners have access to a number of gold annotations, for example when using honeypot questions. In these cases, these trusted annotations can be used to help aggregation models correctly estimate parameters through \emph{semi-supervised learning}. Aggregation methods using even a small number of known-gold annotations for semi-supervised learning can achieve accuracy improvements on simple annotation tasks \cite{wang2011managing,tang2011semi,hovy2013learning}. When label redundancy is low, semi-supervised aggregation approaches are particularly useful \cite{tang2011semi}. Semi-supervised learning is also stipulated to be essential when annotators exhibit overall low reliability or systematic bias in their responses \cite{ipeirotis2010blog}.

As researchers seek to automate ever-more sophisticated tasks, the kinds of annotation needed to train and evaluate AI models becomes increasingly complex. Examples include structured linguistic syntax \cite{marcus1993building}, ranked lists, sequences \cite{rodrigues2014sequence,nguyen2017aggregating,simpson2018bayesian}, open-ended answers to math problems \cite{lin2012crowdsourcing}, or even drawings \cite{dutt2013predicting,ha2017neural}.

\subsection{Defining ``Complexity'' in Annotation}
\label{sec:what-is-complex}

\subsubsection{Annotation Task Complexity}
Human computation tasks span a vast range of complexity, from simple labeling tasks to highly complex work typically involving team coordination and/or highly skilled expertise. With simple tasks, basic task designs and quality assurance methods suffice, whereas more complex work may require very different strategies.
For example, multi-stage \textit{workflow} design strategies aim to intelligently decompose tasks into smaller sub-tasks, sequence the work, and organize human labor accordingly. Such strategies can empower inexpert workers to effectively complete more complex tasks \cite{noronha2011platemate,kittur2011crowdforge,kulkarni2012collaboratively,retelny2017no}. 
One well-known challenge is that each new annotation task often requires designing a new, custom workflow. This challenge has provoked much research to design more general workflows across annotation tasks \cite{kittur2011crowdforge,kulkarni2012collaboratively}. There is a natural correlation between complex annotation tasks and the complexity of the resulting annotations produced (further discussed in the next Section \ref{sec:label-complexity}). 

Our broad goal in this work is to enable aggregation for more complex annotations than has been previously possible. However, we do not claim to be able support all types of arbitrarily complex annotations, nor is our work applicable to all forms of complex human computation tasks. In general, 
task decomposition and workflow design are complementary to aggregation methods, and may include one or more aggregation steps as part of a larger workflow. In some cases, aggregation vs.\ workflow approaches offer alternative paths to producing high quality annotations. In other cases, workflow methods may be essential to achieving success on more difficult human computation tasks.

\subsubsection{Annotation Complexity} \label{sec:label-complexity}
We distinguish the complexity of the annotation task (as just described) from the complexity of the annotations produced by the task.  For example, one could define a highly-complex process to review software program correctness and output a simple binary label (correct or incorrect). On the other hand, a highly complex output (e.g., a software program) might be produced from a one-sentence task description (e.g., ``Write a software program to determine in polynomial time whether two networks are the same.'') 
In the scope of this work, we assume that some set of annotations have been previously collected, and we are agnostic as to the details and complexity of the task design that was used. Instead, our concern rests solely with the complexity of the collected annotations and how their complexity impacts our ability to aggregate them in a general manner. Practically, we consider the extent to which existing general aggregation models do or do not support such complex annotations, and what novel or greater challenges such annotations pose for aggregation that are not met by existing, general models.

Let us consider numeric data and three supported aggregation operators. First (and most simply), a set of numbers can be aggregated by selecting the {\em mode} (the most frequent value, assuming one exists). Second, because numbers support arithmetic operations (addition and division), a set of numbers can be aggregated by computing its {\em mean}. For example, the classic ``Wisdom of Crowds'' example \cite{surowiecki2005wisdom} is averaging individual guesses for the weight of an ox. Note that whereas the mode is selection-based (choose an existing member of set as its aggregate value), the mean is merge-based (combining or fusing set elements into a new value used to represent the set). Finally, a selection-merge hybrid aggregation operator is the {\em median}. Because numbers can be sorted and averaged, the median aggregates a set of numbers with odd cardinality by selecting the middle value, or an even cardinality set by averaging the two middle values. 

Given this foundation of numeric aggregation, let us consider categorical data. Categories are discrete concepts, without any inherent semantics defining relations between categories. Categories do not support sorting or arithmetic operations, and so median and mean operators are not defined. Instead, of the three aggregation operators mentioned above, only the mode is supported for categories.
When the response space (i.e., the number of categories) is small relative to the number of
respondents (e.g., annotators), the same category will tend to get selected multiple times, likely yielding a single,  most-frequent response (the mode). This is the basis for classic majority voting aggregation.
However, as the number of categories grows relative to the number of annotators, it becomes more likely that every annotator selects a unique label, limiting the value of mode-based aggregation.

Moreover, unsupervised weighted voting methods similarly tend to estimate respondent weights based on agreement with the mode response \cite{dawid1979maximum}. 
However, limitations of mode-based aggregation become quickly apparent here as well. For example, with simple binary classification, an even number of annotators may be evenly split, resulting in lack of a mode response. As the number of categories grows, the problem quickly worsens, with every annotator potentially selecting an unique label. In short, weighted voting methods based on the mode work reasonably well on simple annotation tasks (where mode-based aggregation works), but with more complex tasks, use of the mode quickly breaks down. 

A further difference between numeric and categorical data concerns measurement of similarity. 
With numeric responses, one can measure similarity between answers; one might compute the mean and then measure the distance of each annotator's response to it. However, this is not possible with categorical data; there is no inherent notion of similarity between categories, and so no clear way to award partial-credit to annotators whose labels do not exactly match one another's or the mode response. Such reliance on exact-match in estimating annotator reliability for weighted voting represents a key limitation of most prior work on general aggregation models. To date, human computation research on aggregation has largely focused on classification tasks with relatively few categories to choose from \cite{dawid1979maximum,sheng2008get,zheng2017truth}. Thus even with categorical data, prior general models tend to break down as we move beyond small, simple categorization schemes toward fine-grained or {\em extreme classification} \cite{bengio2019extreme}.

This limitation also extends to other data types as well.  If we consider strings (free-form text responses), the response space is typically infinite, challenging mode-based aggregation. There is also no inherent mean operator (effectively requiring text summarization), and while one can define a lexicographic ordering of strings, finding the median response under this ordering is unlikely to provide useful aggregation for most tasks of practical interest. 
%

With ranking data, the space of possible rankings grows rapidly with the number of elements being ranked, again challenging mode-based aggregation. There is no canonical ordering of rankings to identify a median response, nor is there a simple notion of calculating a mean. Classic {\em Borda Count} aggregation \cite{borda1784memoire} first maps each ranked element to its numeric rank position, computes each element's mean rank position across rankings, and finally re-ranks elements by their mean rank. Alternatively, if elements were assigned a value equal to their reciprocal rank position, rather than the rank position itself, this would instead induce {\em Dowdall} aggregation \cite{fraenkel2014borda}. We note that these aggregation operators are applicable only to ranking data, and they operate by effectively mapping rankings to numeric data so that the numeric mean can be applied. 

As we begin to introduce new, arbitrary data structures, such as syntactic parse trees \cite{marcus1993building}, as with categories, strings and rankings, mode-based aggregation remains possible but limited in utility due to the potentially very large response space. In contrast, aggregation by mean and median are unlikely to be applicable since they require defining (meaningful and useful) arithmetic and sorting operations. As we saw with rankings above, one can define custom aggregation operators for each new data type and use case. While customizing the method of aggregation for each distinct annotation task may optimize performance on it, it is cumbersome to have to create and maintain a suite of tailored aggregation methods for every new data type. A great value of a general aggregation model is its ability to be widely applied; while a specialized model formulated for each data type might perform better, a general model that already exists might often be ``good enough'' to save the effort of having to develop a new customized aggregation method for every type.

In general, our exploration of annotation complexity, across diverse data types, reveals a spectrum of complexity rather than a black-and-white, simple vs.\ complex dichotomy. As we have noted, simple classification tasks have few categories that lack any notions of categorical similarity, while more complex classification tasks involve increasingly large and fine-grained categories, with an increasing need to model similarity between fine-grained categories. 
Coreference resolution is such an example task, in which annotators categorize mention-pairs, but some categories are more similar to each other than others \cite{paun2018probabilistic}.
Similarly, because the complexity of rankings datasets increases with the number of ranked elements, pairwise preference labels \cite{simpson2020scalable} that rank two elements sit on the less complex side of this spectrum.
Numeric values can be merged via the mean operator, but while mode-based aggregation would also be possible with a small-interval, likert-scale rating task, the existence of a mode response becomes unlikely with continuous, unbounded numeric tasks. Open-ended, free-form textual response tasks cannot be merged without summarization capabilities and typically have infinite response space, but could be simplified by artificially limiting what responses are allowed.

As we move away from standard data types (e.g., category, number, string) toward more complex and potentially bespoke data structures (e.g., spans/ranges, rankings, or syntactic parse trees), the space of possible labels may expand rapidly while aggregation operators also may become esoteric or non-existent.
Trying to find a general method for aggregating arbitrary data structures appears daunting unless we can somehow translate or decompose these data structures back into standard data types for which general aggregation models can be applied (e.g., as Borda and Dowdell methods do to aggregate rankings).
Limitations of existing approaches helped prompt a 2019 EMNLP workshop calling for modeling of complex annotations \cite{annonlp2019}.

\subsection{Heuristic Quality Controls for Annotations}

\textbf{Attention Checks.} Lacking general annotation and aggregation methods for modeling complex annotations, quality assurance for complex annotations remains a bit eclectic today. For example, one can insert non-task, attention checks (e.g., ``What is the third word on this page?'').
Catching such lapses in attention can help filter out undesired random noise in annotations from what otherwise may be genuine subjective disagreements.
However, such attention checks are easily distinguished from actual task-questions, making it easy for an annotator to pass an attention check while still performing poorly on the actual task of interest~\cite{marshall2013experiences}. 

\citeA{klebanov2008analyzing} implemented a ``validation'' phase, which tries to predict whether disagreements were the result of attention slips or differences of opinion. Unlike the typical concept of an attention check, their method draws conclusions about annotator reliability for a given item rather than globally.

Similarly, free text responses below a certain word count or with bad grammar may be judged as having bad quality \cite{zaidan2011crowdsourcing,li2016tgif}. However, determining how to assess annotation quality by its content requires specifying task-specific checks, which are often relatively ineffective and can significantly reduce the worker pool size \cite{chen2018cicero}. 


\textbf{Gold Questions}. One of the most popular approaches (interchangeably referred to as \emph{honeypot} questions, {\em canaries}, or {\em verifiable checks}), evaluates worker responses against known answers \cite{danula-cscw21-pacm,checco2020adversarial}. This approach can be used to estimate the reliability of workers by how well they score on the honeypot questions. For simple annotation tasks, a small label space permits evaluating responses based on exact match vs.\ gold labels. As we move toward ordinal rating tasks, we might instead assess partial-credit, penalizing ``near misses'' less than other errors. With complex annotation, the space of possible labels may be vast and such partial-credit evaluation becomes essential (e.g., there may be several acceptable ways to translate a sentence, and many other ways of variable quality). However, gold label collection can be expensive as it requires expert annotators, and having a small sample size of gold labels may lead to inaccurate estimates of worker characteristics.

\subsection{Aggregation Models for Complex Annotations} \label{sec:bespoke}
Probabilistic models for annotations \cite{passonneau2014benefits} provide a framework for several useful tools, including parameter inference, semi-supervised learning, and probabilistic task management. The main benefit of such models is their versatility. They can estimate properties of annotators (e.g. reliability) and items (e.g. difficulty) together with inferred truth values. They do not require participation from experts or honeypot items but can benefit from semi-supervised learning. They can be used for decision-theoretic task control and active learning (i.e. how to use estimates to optimize decisions such as when to collect a new label) \cite{dai2010decision,kamar2012combining,nguyen2015combining}. 

For complex annotations of different types, formulating new probabilistic models is certainly doable but non-trivial. Some examples are models based on Hidden Markov Models (HMM) for aggregating crowd-annotated sequences of text within documents \cite{nguyen2017aggregating,simpson2018bayesian} (see top of Figure \ref{fig:examples-partition}) and a Chinese Restaurant Process (CRP) model for short free-response answers \cite{lin2012crowdsourcing}. HMMs assume time-dependent data, and the CRP approach works when there are single discrete correct answers but not when there are continuous spaces of similarly correct ones.
\citeA{branson2017lean} propose a model supporting multi-object bounding box labels that implements an algorithm inspired by the ``facility location problem," using a greedy approach to combine individual bounding boxes into multi-object annotations. Their approach also generalizes to single keypoint labels per item but not multi-object keypoint annotation tasks.

While it may be theoretically possible to use these task-specific probabilistic models for semi-supervised learning, there is no such study yet of semi-supervised learning on complex annotation tasks. There is no simple general formula for adopting semi-supervision in the estimation of any given generative statistical model, replacing some of its unknown parameters with observed data. As an example, for just a standard HMM, enabling semi-supervision in the Stan probabilistic programming language involves implementing the \textit{forward algorithm} to marginalize out the latent states of the chain \cite{stan2018stan}.

Designing probabilistic models for complex tasks requires familiarity with the task domain. More generally, formulating models requires advanced mathematics and statistics knowledge, so it would be helpful to the science community if we could provide reusable models which can be more easily adopted by task owners from diverse backgrounds. A key challenge is formulating the annotation likelihood conditioned on the unknown true value of the item plus any additional parameters that influence the error. If annotator labels $L$ and true value estimators $\hat{L}$ are simple binary, categorical, or one-dimensional continuous variables, it is straightforward to define how they relate to each other according to common conditional probability distributions of the form:
\begin{equation}
P(L|\hat{L}, \theta)
\label{eq:lhat}
\end{equation}
with extra parameters $\theta$ to model effects like annotator reliability. However, when $L$ and $\hat{L}$ are more complex, it becomes less obvious how to treat them in a model. Hidden variables representing complex concepts, such as $\hat{L}$, and their relationship to the observable $L$ can be very difficult to formulate mathematically. An example of a model built specifically for the text sequences task is \citeA{nguyen2017aggregating}'s Crowd Hidden Markov Model, where $\hat{L}$ is the sequence of true latent classes for tokens in a document and $L$ contains the worker-annotated sequences. The mathematical relationship between $\hat{L}$ and $L$ is formulated as an HMM, augmented with additional parameters such as a confusion matrix estimating the error rates by worker and class.  As an example of another complex annotation task -- free text responses -- the task owner would not only have to decide on a latent representation or embedding space for sentences, but also figure out how to translate between that latent space of $\hat{L}$ and the observable space of text $L$, which is challenging. \citeA{li2020crowd} presents such an aggregation method for free-text translations that leverages the field of neural natural language processing to convert strings into an embedding space through pre-trained large language models.

Our goal is to provide more flexible options for complex tasks: general-purpose and task-independent probabilistic models for aggregating complex annotations. We will investigate heuristic estimators of $\hat{L}$, a model that assumes a discrete $\hat{L}$ over the available labels $L$, and a model that projects labels $L$ and estimator $\hat{L}$ into a latent coordinate space. We also investigate label merge methods that yield a $\hat{L}$ value that is not one of the available $L$, and methods for partitioning multi-object annotations before aggregating them.

\subsubsection{Merging via Partition}
\label{sec:prev-decomp}


\citeA{parameswaran2016optimizing} characterize support for open-ended crowdsourcing as the ``next frontier'', noting aggregation as a key challenge. As part of this work, one idea they briefly mention is decomposing complex labels into simpler ones: ``instead of treating [a] bounding box as a whole, we can decompose it down into boolean crowdsourcing answers for individual pixels.'' However, applying this approach as stated to bounding boxes would not necessarily yield a new bounding box because pixel-level aggregation of overlapping bounding boxes typically yields non-rectangular regions. \citeA{lee2018aggregating} side-step this issue by applying this approach to  semantic segmentation, where it is fine if pixel-level aggregation produces arbitrary new segmentation shapes. In \citeA{braylan2021mergeandmatch}, we decompose bounding boxes into upper-left and lower-right numeric vertices rather than pixels, so aggregation  produces two vertices that also define a valid bounding box. Exploring alternative primitive representations to best implement this meta-algorithm for different annotation tasks remains an open area for future work (Section \ref{sec:future-work}). 

\citeA{nguyen2017aggregating} also apply this idea as a baseline for aggregating text sequences. Their Token-wise Majority Vote (TMV) baseline breaks sequence annotations into individual tokens and then performs a token-wise majority vote. However, the authors find that this simple token-level aggregation significantly underperforms compared to their proposed bespoke model. Furthermore, many kinds of complex annotations, such as free text or tree structures simply cannot be aggregated by voting on the individual tokens.



\section{Overview of Methods}

This section presents an initial overview of the set of general methods we propose for aggregating complex annotations. Section \ref{sec:methods} describes distance-based selection methods in detail. Details of decomposition and merging methods appear next in Section \ref{sec:method-decompose}. Finally, we describe distance-based partitioning of multi-object annotations in Section \ref{sec:multi-object}.

An aggregation \emph{method} here is any function that inputs a dataset of items with multiple annotations per item and outputs a dataset of items with a single predicted annotation per item. We also mention several aggregation \emph{models}. These are aggregation methods that involve estimating a generative statistical model -- that is, a formulation and parametrization of the data-generating process -- from the given input data, from which predicted annotations per item can be inferred and output. The benefit of using such a probabilistic model rather than a heuristic was motivated earlier in Section~\ref{sec:bespoke}.
 
Section \ref{sec:methods} presents distance-based selection methods, beginning with two simple (yet often effective) heuristic approaches that span {\em local} aggregation (SAD) vs.\ {\em global} aggregation (BAU) in Sections \ref{method:sad} and \ref{method:bau}, respectively. Following this, we introduce the MAS model in Section  \ref{method:mas} and the MADD model in Section \ref{method:madd}.
Both MAS and MADD can weight labels by estimated annotator ability, thereby combining the local aggregation benefits of SAD with the global aggregation benefits of BAU. MADD simplifies the assumptions of MAS by representing the unknown ground truth as one of the available labels rather than modeling a latent embedding space.
We further describe a semi-supervised learning approach for our aggregation models in Section \ref{method:semisupervised}. 

Next, Sections \ref{sec:method-decompose} and \ref{sec:multi-object} describe approaches for when annotations can be \emph{merged} or contain multiple objects that can be \emph{partitioned}.
Regarding merging, whereas aggregation by selection is constrained to select among the labels produced by annotators, aggregation by merging has the potential to yield a new, better label than was  produced by any annotator.
Regarding partitioning, whenever there are multiple objects to be labeled for a given item, the ability to partition workers' label sets allows a ``mixing and matching'' of labels across different workers, with a similar potential benefit to produce a better combined label set than was produced by any single annotator. 

\label{sec:framework}

\SetKwInOut{Input}{Input}
\SetKwInOut{Output}{Output}
\SetKw{Continue}{continue}
\SetKwRepeat{Struct}{struct \{}{\}}%
\newcommand{\Float}{\KwSty{float}}

\section{Distance-based Selection Methods}
\label{sec:methods}

This section describes our general, distance-based approach for modeling complex annotations without requiring a task-specific aggregation model for every different task. Section \ref{sec:distances} describes how we transform complex annotation datasets into distance datasets, allowing a probabilistic model to operate on simpler, task-agnostic continuous values. Following this, we describe a simple local aggregation method ``Smallest Average Distance'' (SAD) and a simple global aggregation method ``Best Available User'' (BAU) in Sections \ref{method:sad} and \ref{method:bau}. Finally, we introduce the MAS model in Section \ref{method:mas} and the MADD model in Section \ref{method:madd}.


\subsection{From Annotations to Distances}
\label{sec:distances}

Our key idea to obviate the need for task-specific models is to model distances between labels, rather than the labels themselves. The models we propose are agnostic to the distance function; we leave it to the \emph{task owner} (who defines the annotation task) to specify an appropriate distance function for the given task. Typically such distance functions already exist: as long as there is an \emph{evaluation function} to quantify error in comparing predicted labels vs. gold labels, it can used to construct a distance function for our model. In fact, in the literature for measuring \emph{inter-annotator agreement}, \emph{Krippendorff's $\alpha$} \cite{krippendorff2004reliability}'s general formulation allows one to plug in an arbitrary distance function between annotations to quantify annotator agreement \cite{Braylan-web2022}.

Formally, a distance function should satisfy the following: 
\begin{description}
\item[Non-negativity:] $f(x,y) \geq 0$ \item[Symmetry:] $f(x,y) = f(y,x)$ \item[Triangle Inequality:] $f(x,y) \leq f(x,z) + f(z,y)$ for any $z$
\end{description}
In practice, the distance function supplied by the task owner need not meet all three of these requirements because its output can often be transformed to satisfy them. In particular, \emph{non-negativity} can be satisfied by conversion to quantiles or exponentiation, and {\em symmetry} can be satisfied by adding (or averaging) $f(x,y)$ and $f(y,x)$. For example, Jensen-Shannon Divergence \cite{fuglede2004jensen}  
is a symmetrized version of asymmetric Kullback-Liebler Divergence.
It is also unclear how important it is to meet the \emph{triangle inequality} requirement, 
and one may skip this requirement and call it a \emph{dissimilarity} instead of \emph{distance}, as done in \citeA{mathet2015unified}.

Beyond these three requirements, the choice of distance function is left to the task owner. How does the task owner choose a distance function? The aphorism ``all models are wrong, but some are useful'' \cite{box1979all} can be said of distance functions as well -- any distance function is an imperfect summarization of the relation between two complex annotations, with some being more imperfect than others. In \cite{Braylan-web2022}, we propose an annotator agreement metric to help choose between distance or evaluation functions, but the aggregation methods studied here are agnostic to this choice. In the experiments we use whatever distance functions are standard for the given dataset.

After selecting a distance function $f$, we induce a \emph{distance dataset} $D$ from the set of annotations by computing the distance between all pairs of labels for each example.
{\bf Table \ref{Tab:distances}} shows a simple example of input annotations and output distances. This produces a symmetric matrices of distances $D_{i,(u,v)} = f(L_{iu}, L_{iv})$
between annotations by workers (annotators)
$u, v \in U$ for each item $i \in I$. In the extreme case of all workers annotating all items, the total size of this distance dataset would be $\|U\|^2\|I\|$.

A distance-based annotation model can be used to infer true values for each item, and it may also infer further attributes, such as user error and item difficulty. Whereas Equation~\eqref{eq:lhat} ($P(L|\hat{L}, \theta)$) models annotations, we now instead model annotation distances with the conditional likelihood 
$ P(D | \theta)$.
This critical transformation of the data allows distance-based methods to operate entirely on the continuous space of distances, for both observed and inferred variables, rather than modeling the complex objects themselves. In this manner, we avoid the main difficulty in designing probabilistic models directly for the actual complex annotations. In Section \ref{method:sad} through Section \ref{sec:multi-object}, we show how to use distance-based annotation models to produce final aggregated labels.

\begin{table}[t]
\begin{center}
\begin{tabular}{c|p{6cm}}
\toprule
\bf Annotator & 
\begin{center} \vspace{-1em}
{\bf Translation}
\vspace{-1em} \end{center}\\
\hline
1 & Now Hamas and Israel should make peace so that this bloodshed comes to an end.\\
2 & Hamas and Israel should reconcile so that this bloodshed comes to an end. \\
4 & Now that the Hamas and Israel should be made to compromise, so the blood and evil. \\
\midrule
\end{tabular}

\begin{tabular}{ccc}
\\
\toprule
{\bf Annotator-1} & {\bf Annotator-2} & {\bf Annotation Distance}\\
\hline
1 & 2 & 0.4333\\
1 & 4 & 0.8586\\
2 & 4 & 0.8758\\
\bottomrule
\\
\end{tabular}
\end{center}

\caption{Example of input complex annotation dataset (top) converted to annotation distances (bottom). Shown here is a sample of Urdu-to-English translations and the result of using GLEU score as a distance function (discussed in Section \ref{evaluation}).}
\label{Tab:distances}
\end{table}

\subsection{Smallest Average Distance (SAD)}
\label{method:sad}

One simple approach to aggregating complex annotations from their distances, {\em Smallest Average Distance} (SAD), can be interpreted as a generalization of majority voting, operating entirely locally to each individual item. SAD assigns a score $\varepsilon_{iu}$ to each annotation $L_{iu}$ for item $i$ by annotator $u \in U(i)$, equal to that annotation's average distance to all other annotations for the same item $i$.
SAD scores are calculated as follows:
\begin{equation}
\label{eq:sadScore}
\varepsilon_{iu}^\text{SAD} = \frac{1}{\|S_{iu}\|} \sum S_{iu} \ \ \ , \ \ \ 
S_{iu} = \{D_{i,(u,v)} | v \in U, v \neq u\}
\end{equation}
where $S_{iu}$ denotes the set of all annotation distances for item $i$ between annotator $u$ and any other annotator $v \in U(i)$.
SAD predicts this most central annotation, having the smallest average distance to all other annotations for item $i$, to be deemed the best consensus annotation for that item. In other words, the aggregated annotation is:
$$
\hat{L}_i = L_{i u_i^\prime}, \textnormal{\ \ \ }
u_i^\prime = \textrm{argmin}_{u \in U(i)} \varepsilon_{iu}^\text{SAD}
$$

\subsection{Best Available User (BAU)}
\label{method:bau}

Whereas SAD operates entirely {\em local} to each item, {\em Best Available User} (BAU)  passes over the annotation distance dataset to estimate {\em global} annotator error across items, assigning a $\varepsilon_{iu}$ score to each annotation based on estimated annotator error $\varepsilon_{u}$ 
as follows:

\begin{equation}
\label{eq:bauScore}
\varepsilon_{iu}^\text{BAU} = \varepsilon_{u} = \frac{1}{\|S_{u}\|} \sum S_{u} \ \ \ , \ \ \ 
S_{u} = \{D_{i,(u,v)} | i \in I, v \in U, v \neq u\}
\end{equation}

BAU can be interpreted as an unsupervised estimate of annotator reliability based on each annotator's average distance from other annotators' labels across the whole dataset. This means that labels are scored entirely by their annotator's global reliability, regardless of the annotator's label for the particular item. BAU thus predicts the best label for each item to be whichever label came from the best available user (annotator) for that item. SAD and BAU thus present as contrasting extremes in exploiting local vs.\ global information in modeling. Similarly to SAD, the aggregated annotation for BAU is:
$$
\hat{L}_i = L_{i u_i^\prime}, \textnormal{\ \ \ }
u_i^\prime = \textrm{argmin}_{u \in U(i)} \varepsilon_{iu}^\text{BAU}
$$

\subsection{Multidimensional Annotation Scaling (MAS)}
\label{method:mas}

\begin{figure}[p]
\centering
  \vspace{1em} Multidimensional scaling likelihood function: $D_{i,(u,v)} \sim \mathcal{N}(\| \mathbf{x}_{iu} - \mathbf{x}_{iv} \|, \sigma)$ \\
  \includegraphics[scale=.6]{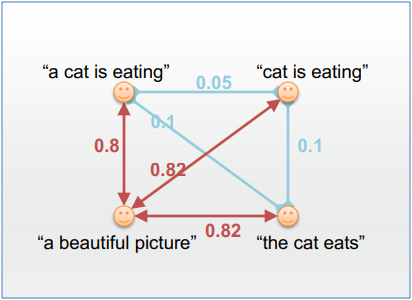}
  \includegraphics[scale=.6]{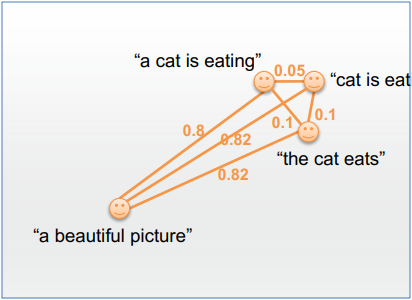} \\
  \vspace{1em} Prior on annotation embeddings: $\mathbf{x}_{iu} \sim \mathcal{N}(\vec{0}, c I)$ \\
  \includegraphics[scale=.6]{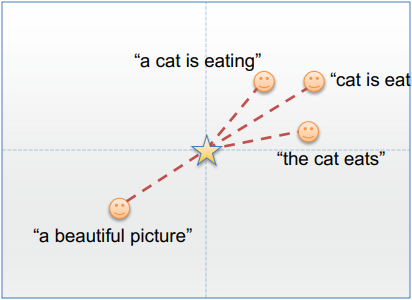}
  \includegraphics[scale=.6]{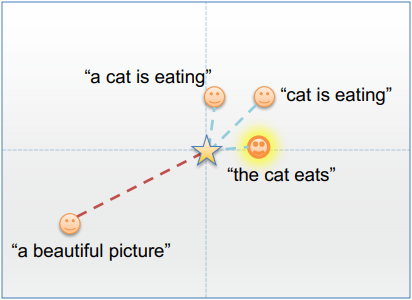} \\
  \vspace{1em} Prior scale varying by annotator reliability and item difficulty: $\mathbf{x}_{iu} \sim \mathcal{N}(\vec{0}, \gamma_{u} \delta_{i}  I)$ \\
  \includegraphics[scale=.6]{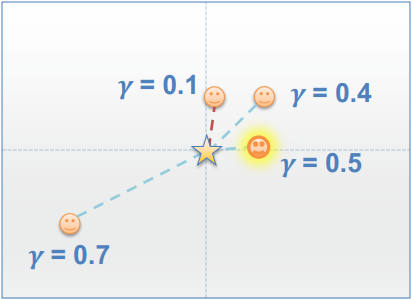}
  \includegraphics[scale=.6]{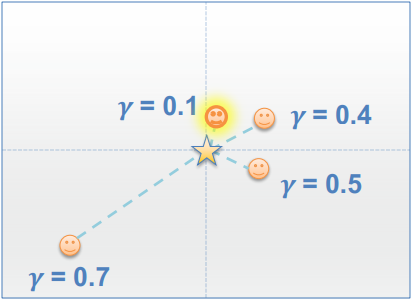}
\caption{Illustrating the MAS model. 
{\bf Top}: the likelihood function uses observed distances between labels to place label embeddings in Euclidean space at approximate relative distances. The red large distances are placed farther, and the blue small distances are placed closer. 
{\bf Middle}: a prior on the magnitude of the embeddings pushes larger clusters of labels (three faces on top right quadrant) closer to the origin point representing ground truth, emulating majority vote and selecting the highlighted face as the aggregate label $\hat{L}$. {\bf Bottom}: varying the scale of the prior by annotator reliability and item difficulty parameters allows global information to affect placement of embeddings, emulating weighted majority vote. Here the annotator with $\gamma=0.1$ representing better global reliability is moved closer to the origin and selected as the aggregate label $\hat{L}$.}
\label{fig:mas-explained}
\end{figure}

The next approach we discuss is to model a $K$-dimensional representation space in which
we designate ground truth as the central point and then embed all annotations as surrounding coordinates. These annotation embeddings are estimated such that their distances to each other match the observed annotation distances, while their norms (distances from the center) are regularized by estimated user error (See Figure \ref{fig:mas-explained}).
Similar in spirit to word and sentence embeddings in NLP \cite{mikolov2013efficient}, the annotation embeddings and other parameters produced by our model will be useful here.

In order to compute annotation embeddings, we devise a probabilistic model based on multidimensional scaling \cite{mead1992review}. Multidimensional scaling estimates coordinates $\mathbf{x}$ of points given only a matrix of distances between those points by minimizing an objective function, generally $\sum(\|\mathbf{x}_i - \mathbf{x}_j\| - D_{ij})^2$. The estimated coordinate vectors carry meaning in their position relative to each other, rather than in their absolute direction or magnitude. Multidimensional scaling is a generalization of kernel PCA when the kernel function is isotropic \cite{williams2001connection} and is popular in dimensionality reduction and data visualization.

Our model, {\em multidimensional annotation scaling} (MAS), is a Bayesian probabilistic model having a multidimensional scaling likelihood function in which the estimated coordinates serve as annotation embeddings. Instead of the data populating a single distance matrix, each item has a separate annotation distance matrix (i.e. distances between annotations for different items are irrelevant). Additionally, because each user may annotate several items, we use the full dataset to compute global parameters representing annotator reliability, which serve in the priors for the local parameters of each item's multidimensional scaling likelihood ($\gamma$ in Equation \ref{eq:mas-likelihood} and Figure \ref{fig:mas-explained}).
%
\label{formula:mas}
We define the MAS model as follows in Equations~\eqref{eq:masSelect}-\eqref{eq:masF}, and we illustrate the basic premise of MAS in Figure~\ref{fig:mas-explained}. 
\begin{equation} \label{eq:masSelect}
\begin{split}
\hat{L}_i = L_{i u_i^\prime}, \textnormal{\ \ \ }
u_i^\prime = \textrm{argmin}_{u \in U(i)} \varepsilon_{iu}^\text{MAS}, \textnormal{\ \ \ }
\end{split}
\end{equation}
\begin{equation} \label{eq:masScore}
\begin{split}
\varepsilon_{iu}^\text{MAS} = \|\mathbf{x}_{iu}\|
\end{split}
\end{equation}
For each item $i$, the model enables selection of the ``best'' annotation as the true value estimator $\hat{L}_i$. Like SAD and BAU, this selected annotation is the one with the smallest inferred error $\varepsilon_{iu}$ out of all annotations made by users $U(i)$ that worked on item $i$. Annotations may also be graded and ranked according to this $\varepsilon_{iu}$, which represents the model's predicted distance from annotation $L_{iu}$ to the best possible annotation. 

In the MAS model, the origin in the space of embeddings $\mathbf{x}$ is taken to represent the true value for an item, so the norm of $\mathbf{x}_{iu}$ is understood as that annotation's distance from the truth, or $\varepsilon_{iu}$. This interpretation differs from standard multidimensional scaling, where the magnitude of the coordinates need not carry any meaning. In order to interpret $\varepsilon_{iu}$ in this way, MAS makes some important assumptions. First, the embedding space is \emph{isotropic}, i.e. direction carries no meaning. This follows from standard Multidimensional Scaling. Second, the space of annotation quality \emph{unimodal}, i.e. there is a single optimal point. This allows MAS to treat the origin point accordingly and interpret the embedding magnitudes as the inverse of their annotation quality. 
\begin{equation} \label{eq:likelihood}
\begin{split}
D_{i,(u,v)} \sim \mathcal{N}(\| \mathbf{x}_{iu} - \mathbf{x}_{iv} \|, \sigma) ~~~~,~~~~
D_{i,(u,v)} \in \mathbb{R}_+
\end{split}
\end{equation}
Equation~\ref{eq:likelihood} is the generalized multidimensional scaling objective function expressed as a probabilistic likelihood. Maximizing the normal likelihood with free scale parameter $\sigma$ minimizes the square error between observed distances in the data and learned distances in the embedding space.

\begin{equation}
\begin{split}
\mathbf{x}_{iu} \sim \mathcal{N}(\vec{0}, \gamma_{u} \delta_{i} I) ~~~~,~~~~
\gamma_u, \delta_i \in \mathbb{R}_+, \mathbf{x}_{iu}, \tilde{\mathbf{x}}_{iu} \in \mathbb{R}^K
\end{split}
\label{eq:mas-likelihood}
\end{equation}


The annotation embeddings $\mathbf{x}$ correspond to draws from a $K$-dimensional multivariate normal distribution centered at zero. The covariance is determined by multiplying the identity matrix $I$ with scale parameters $\gamma$, representing user error, and $\delta$, representing item difficulty. The model prefers to fit larger values of the scale parameters when those users and items are associated with larger distances in the data. When many annotations have small distances between each other, the model favors placing them closer to the origin, thereby rewarding consensus (in contrast with isolated annotations having higher distances from the others). The model also favors placing annotations made by smaller-$\gamma$ users closer to the center, thereby rewarding annotator reliability.

\begin{equation} \label{eq:masF}
\begin{split}
\log{\gamma_u} \sim \mathcal{N}(0, \Phi) ~~~,~~~  
\log{\delta_i} \sim \mathcal{N}(0, \Psi) \\
\end{split}
\end{equation}

The parameters $\gamma$ and $\delta$ are given lognormal priors with configurable scales $\Phi$ and $\Psi$, respectively. \citeA{braylan2020modeling} recommended values of $\Phi=1$ and $\Psi=1$, with $K=8$ embedding dimensions. In contrast, we use $\Phi=0.25$, $\Psi=0.025$, and $K=3$, based on the result of tuning on portions held out of the bounding box and keypoints datasets from \citeA{braylan2021mergeandmatch} (also used during the development of our merging and partitioning methods). Note that setting $\Psi=\frac{1}{10}\Phi$ has the effect of attributing more variability to annotator reliability than item difficulty. Furthermore, whereas \citeA{braylan2020modeling} used parameters $\bar{\gamma}$ and $\bar{\delta}$ as the locations of hierarchical priors for $\gamma$ and $\delta$, respectively, we have swapped those out for the fixed location 0. This not only simplifies the model, but we have found it yields more reliable parameter estimation.


\vspace{1em}
\noindent{\itshape Parameter Estimation}
\vspace{1em}

We estimate MAS by maximizing the joint probability of Equations~\eqref{eq:likelihood}-\eqref{eq:masF}. We specify the model in the {\em Stan} probabilistic programming language \cite{carpenter2017stan}. Stan supports maximum a posteriori (MAP) estimation, variational inference (VI), and Markov chain Monte Carlo (MCMC). Our experiments run the fastest method, MAP, using Stan's default default L-BFGS optimization, until convergence or a maximum of 1500 iterations. 

Free variables $\mathbf{x}$, $\gamma$, and $\delta$ are initialized randomly according to Stan's default settings, except for $\gamma$ parameters, which are set to the average annotation distance of each user (i.e., the BAU score from Section~\ref{method:bau} below). Because L-BFGS performs local optimization, it is helpful to initialize parameters with informed prior estimates when possible. In this case, BAU scores provide easy initial estimates for $\gamma$ parameters. While $\delta$ parameters could be initialized to item-average distance, one depends on the other, so we use default estimates for $\delta$ parameters, as stated above.

\subsection{Model of Ability, Difficulty, and Distances (MADD)}
\label{method:madd}

The {\em Model of Ability, Difficulty, and Distances} (MADD) is inspired by GLAD \cite{whitehill2009whose}, modeling annotator ability, item difficulty, and the probability of each label being the best for each item. GLAD is expressed as:
$$
p(L_{ij}=Z_j \mid \alpha_i, \beta_j) = \frac{1}{1 + e^{- \alpha_i \beta_j}}
$$
This is a log odds model of each label $L_{ij}$ by annotator $i$ for item $j$ being correct, exactly matching gold $Z_j$. Probabilities are influenced by annotator ability $\alpha_i$ and item difficulty $\beta_j$. 
Using an indicator function $I(L_{ij}, Z_j)$ that outputs 1 if label $L_{ij}$ exactly matches gold $Z_j$ (otherwise 0), GLAD can be equivalently formulated as:
$$
I(L_{ij}, Z_j) \sim  \textrm{Bernoulli}(\frac{1}{1 + e^{- \alpha_i \beta_j}})
$$


For complex labels, it is more appropriate to measure partial credit than exact match. We thus replace $I$ with a continuous positive function $d$, such as a distance function. Now, rather than a Bernoulli distribution, a continuous distribution is more appropriate. We use a half-normal distribution, repurposing the expression $\alpha_i \beta_j$ as the scale parameter:
$$
d(L_{ij},Z_j) \sim \mathcal{N}(0, \alpha_i \beta_j), L_{ij} \in [0, \infty)
$$
In other words:
$$
p(d(L_{ij},Z_j) \mid \alpha_i, \beta_j) = \frac{1}{{\alpha_i \beta_j \sqrt {2\pi } }}e^{{{ - d(L_{ij},Z_j)^2 } / {2(\alpha_i \beta_j)^2}}}
$$
This defines the probability of observing a distance between two labels given that one of the labels is the latent ground truth $Z_j$ for the item $j$. The goal is to estimate the probability of each label being that actual ground truth and choose the label with the highest probability.

\vspace{1em}
\noindent{\itshape Parameter Estimation}
\vspace{1em}

To estimate MADD parameters, we maximize the log likelihood over the whole dataset:
$$
- \frac{N}{2} \ln(2\pi) - \frac{1}{2} \sum_{j}\sum_{i} \ln((\alpha_i \beta_j)^2)
- \sum_{j}\sum_{i} \frac{d(L_{ij},Z_j)^2} {2(\alpha_i \beta_j)^2}
$$
$N$ is the total number of labels, and summation is performed only over available i-j (worker-item) combinations.
%
%
The Expectation step consists of marginalizing over possible values of $Z_j$, in this case the set of available labels $L_j$ for item $j$.
$$
\mathbb{E} { \frac{d(L_{ij},Z_j)^2} {2(\alpha_i \beta_j)^2} }
= \frac{1}{\|L_j\|} \sum_{z \in L_j} { \frac{d(L_{ij},z)^2} {2(\alpha_i \beta_j)^2} }
= \frac{1}{\|L_j\|} \sum_{h} { \frac{d(L_{ij}, L_{hj})^2} {2(\alpha_i \beta_j)^2} }
$$

The Maximization step maximizes the log probability with additional terms for priors on $\alpha$ and $\beta$. As there is no analytical solution, we approximate the maximum using the BFGS algorithm with the gradient of log probability along $\alpha$ and $\beta$.

One of the benefits of the MADD model is that it is closely related to existing models from prior literature in label aggregation, which also usually perform EM or Bayesian techniques marginalizing over a discrete latent parameter representing ground truth. While this model was derived from the GLAD model, other models may be adapted in a similar manner from modeling categorical exact match to modeling continuous distance.

However, compared to MAS which projects labels into a continuous embedding space containing latent ground truth, MADD makes a simplifying assumption that the latent ground truth $Z_j$ must be one of the available labels. This difference turns out to be rather important as we find that MAS usually outperforms MADD on real data (Section \ref{sec:masvsmadd}). Therefore we suggest MADD as more of a new baseline, a weighted general complex annotation model that is simpler and perhaps more interpretable, but less powerful, than MAS.

\subsection{Semi-supervised Learning}
\label{method:semisupervised}

Section \ref{background:semisup} discussed the value of utilizing any available gold labels for semi-supervised model estimation. 
Semi-supervised learning is expected to be most useful when peer-agreement falls short in inferring annotator quality, such as when annotators are few and unreliable on average, or when annotators share a consistent bias.  When one annotator disagrees from the rest, peer agreement would assume they are incorrect, whereas use of limited gold can reveal cases when the majority is wrong.


In this section, we discuss how to enable aggregation models for complex annotations to perform semi-supervised learning. Generally speaking, the way semi-supervision is used in models of annotation is that data on how accurately workers performed on known ground truth helps influence the estimation of their reliability parameters. In a probabilistic model of observed annotations given latent parameters, the latent parameter representing unknown ground truth of an item can then be replaced by known observations for the items which have ground truth available. The difficulty with models of complex annotations here is that there is not a direct correspondence between the latent representation of ground truth and the observed one. For example, the MAS model represents latent ground truth as embedding vectors. Instead, we can again work on annotation distances. In this case, performing semi-supervised learning on a model of complex annotations involves replacing the latent estimate of distance between worker annotation and ground truth with an observed distance between worker annotation and ground truth.


We propose a new method here for implementing semi-supervised learning generally with any kind of weighted aggregation model, including MAS and MADD.
Our approach is to measure each annotator's average distance from known ground truth labels, and assign those values to the annotator error parameters.

\begin{equation}
g_{u} = \frac{1}{\|S_{u}\|} \sum S_{u} \ \ \ , \ \ \ 
S_{u} = \{f(L_{iu}, T_{i}) | i \in G\}
\end{equation}
Our method assigns each annotator $u$ an error scale $g$ based on the average output of distance function $f$ between the annotation $L_{iu}$ by $u$ and the ground truth $T_{i}$, over all items $i$ in the set of available ground truth $G$.

\begin{equation}
\begin{split}
\varepsilon_{iu}^\text{SMAS} = \|\mathbf{x}_{iu}\|
~~~,~~~  
\mathbf{x}_{iu} \sim \mathcal{N}(\vec{0}, g_{u} \delta_{i} I)
\end{split}
\label{eq:semisup}
\end{equation}
Here the model for semi-supervised MAS just replaces the learned scale parameter $\gamma$ (Equation \ref{eq:mas-likelihood}) with the scale $g$ as calculated in Equation \ref{eq:semisup} above.

In this setting, annotator error parameters are no longer learned from inter-annotator consensus, but overridden by these ``honeypot'' estimates.
The requirement for this approach is that every annotator must be given these honeypot questions. One benefit of this new approach is that it is much more general and easier to implement than the approach described in \citeA{braylan2020modeling}.
One potential drawback compared to the previous approach is that once these annotator reliability parameters are fixed, they no longer benefit from refinement on the remaining non-gold labels.
Nonetheless, we show in Section \ref{sec:results-real} that, even with only a small fraction of items held out as honeypots, this new approach is more effective than the semi-supervised model from \citeA{braylan2020modeling}.

\section{Decomposition and Merging Methods}
\label{sec:method-decompose}

\begin{figure}
\centering
\includegraphics[scale=.4]{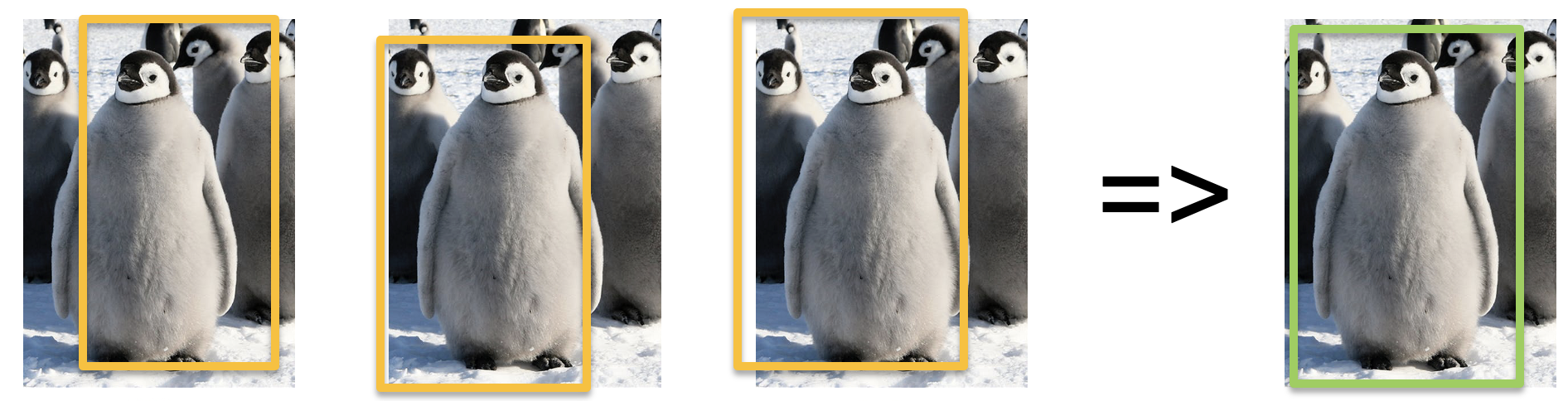}
\caption{The merging of individual bounding boxes by taking the average of the corner coordinates can often be better than any of the original bounding boxes.}
\label{fig:merge-penguins}
\end{figure}

The methods described in Section~\ref{sec:methods} merely {\em select} the best annotator label from the set of all annotator labels. However, further improvement may be possible if we could instead {\em merge} annotator labels, with the goal of inducing a better combined label than any individual annotator actually produced. For example, the classic tale from the ``Wisdom of Crowds'' \cite{surowiecki2005wisdom} is to average all guesses for the weight of an ox, rather than trying to select the best guess. 
%
%
In the case of complex labels, this is exemplified in Figure \ref{fig:merge-penguins}, where bounding boxes are merged by taking the average of their coordinates.

Per Section \ref{sec:prev-decomp}, we pursue decomposing complex annotations into simpler primitives, merging these, and then reconstituting the complex annotation type from the merged primitives.
To accomplish this in a general manner, we exploit standard object serialization techniques \cite{haverlock1998object}. Specifically, we assume any complex annotation is implemented in software as a data structure, which in turn can be automatically decomposed into its constituent primitives (and later reconstituted) using standard serialization operations in a programming language. By decomposing a complex annotation into primitives for which aggregation operators already exist, we can merge complex annotations at the level of their primitives (using aggregation operators defined for such primitives), rather than having to define a new merge function for every new complex annotation type. Once aggregation of the primitives has been performed, serialization operators can be applied in the other direction to reconstitute the complex annotation (data structure) from the aggregated primitives. We assume that primitives are numeric values, and we only consider complex annotation types that can be faithfully decomposed into numeric values without any loss of information. Decomposition of complex annotations into noisy, approximate representations is left for future work.

The merge step can optionally provide weights for each label. In our work, these come from the estimated ``distance from gold'' output by the SAD, BAU, or MAS models, denoted by $\varepsilon_{iu}$ in Equations \ref{eq:sadScore}, \ref{eq:bauScore}, and \ref{eq:masScore}, respectively. We translate these into weights as $w_{iu} = \varepsilon_{iu}^{-1}$.
In the case of the MADD model, instead of distances from gold we get probabilities $p$ of each label being gold, which we could translate into weights as $w = -log(p)^{-1}$.
{\bf Algorithms~\ref{alg:decomposition}-\ref{alg:recomposition}} describe these steps for conducting label decomposition, weighted merging of numeric primitive values, and recomposition into new complex labels.
 
We consider four complex annotation types that can all be decomposed into simple numeric primitives:

\begin{itemize}
 \item \textbf{Text sequences}: merged by aggregating values for (left, right) span boundaries.
 
 \item \textbf{Bounding boxes}: merged by aggregating ($x$,$y$) coordinates of the upper-left and lower-right corners.
 
 \item \textbf{Keypoints}: merged by aggregating ($x$,$y$) coordinates for each vertex of the skeleton (see Section~\ref{sec:datasets} for details on keypoints).

 \item \textbf{Ranked lists}: represented as a mapping from each entry in the list to a numeric value corresponding to that entry's rank position. The set of numeric values associated with each entry across annotators can then be aggregated. Finally, we can sort the entries by value to induce the final aggregated ranking.  Note that this aggregation method induces the popular {\em Borda Count} \cite{borda1784memoire} method for aggregating ranked lists.
 \end{itemize}

{\bf Numeric Aggregation.} The simplest {\em unweighted} method assumes that all labels are equally reliable and applies a standard median, arithmetic mean, or any other relevant statistic, such as a geometric or harmonic mean. This is the classic ``wisdom of crowds'' setup: assuming all guesses for the weight of an ox are equally plausible \cite{surowiecki2005wisdom}. However, if we have positive-valued reliability estimates for each label, we might instead compute a weighted mean (or {\em expectation}) for more informed weighted voting.
Rather than the mean, the use of the median can increase aggregation robustness against outliers (e.g., very low quality annotations) \cite{zheng2017truth}.
In \citeA{braylan2021mergeandmatch}, we compared unweighted and weighted median merging using estimates of label reliability from various compared models and found MAS to give the best weights overall. Following numeric aggregation, the original complex annotation type can then be reconstituted from the aggregated numeric primitives. 



\label{sec:decomposition}



\section{Methods for Distance-based Partitioning of Multi-object Annotations}
\label{sec:multi-object}

\begin{figure}
\centering
\includegraphics[scale=.4]{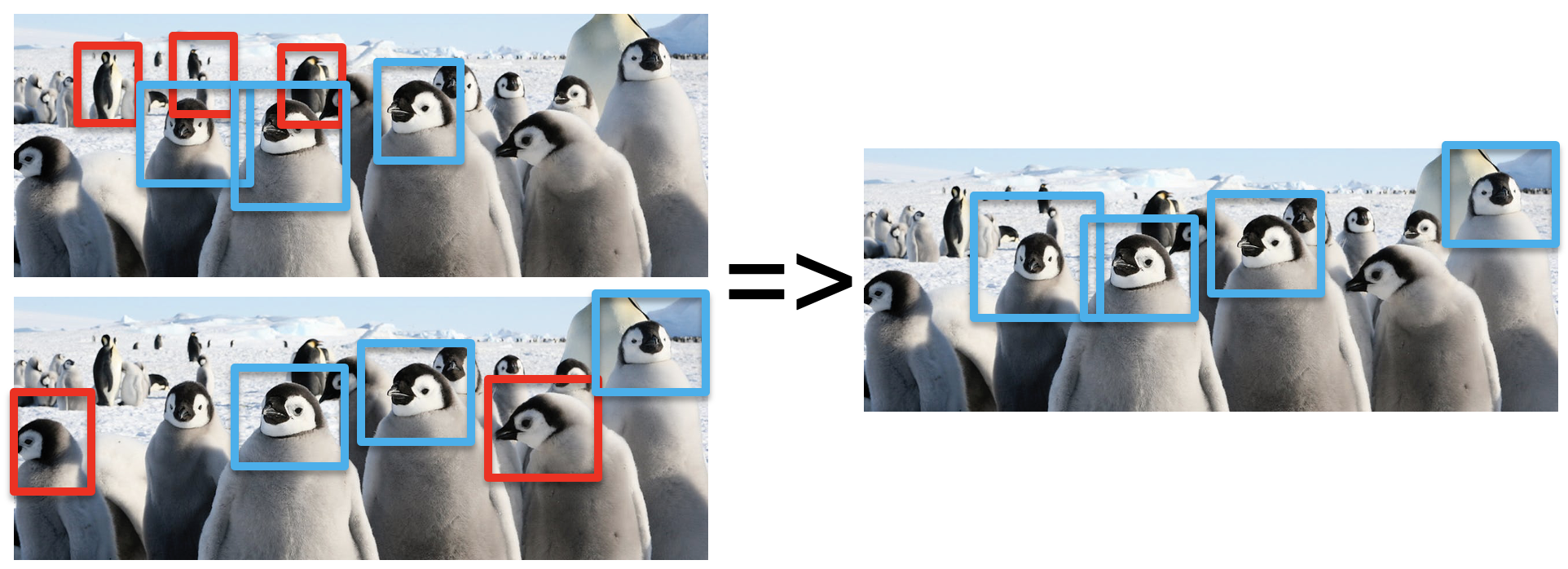}
\caption{Choosing different subsets of multi-object annotations to aggregate may result in a better final overall multi-object annotation than any one annotator's original set. For example, imagine if the instructions were ``highlight all forward-facing penguins in the foreground'' and one annotator overlooked the ``forward-facing'' instruction while the other overlooked the ``foreground'' instruction.}
\label{fig:mixnmatch-penguins}
\end{figure}

\begin{figure}
\centering
\includegraphics[scale=.8]{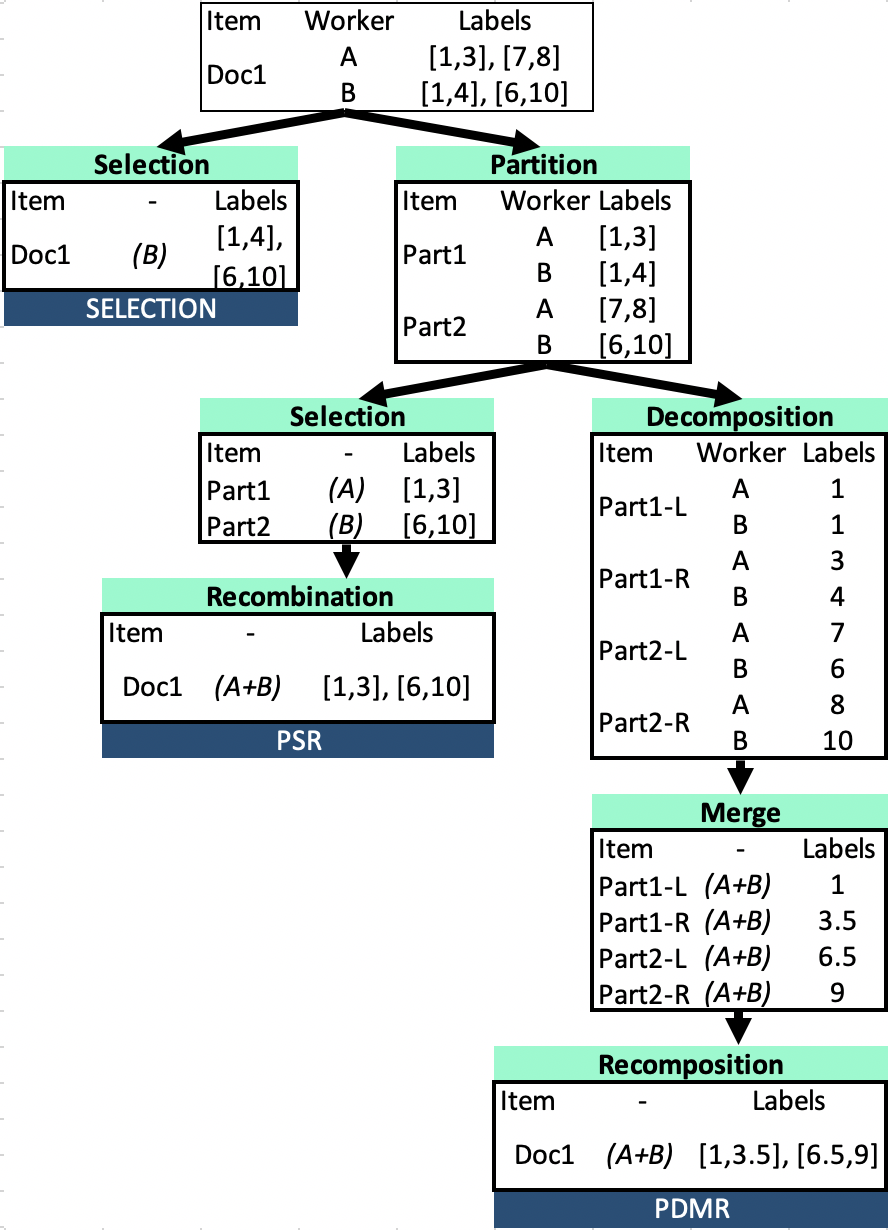}
\caption{Alternative aggregation paths for multi-object annotation. We show an example text sequence annotation task, with two workers each marking two differing spans in a document. Spans are defined by a {\em start} and {\em end} position.}
\label{fig:partitiondiagram}
\end{figure}


So far we have considered annotation tasks that require only a single annotation per item. However, for {\em multi-object annotation} tasks, such as those in Figure~\ref{fig:examples-partition}, each item is actually annotated by a set of labels, each corresponding to a distinct {\em object} found in the item. In such tasks, items may contain any number of objects, and the annotator must search the item to find all of the objects to be labeled. It is important to understand that the methods we have described thus far can be used for such multi-object annotation tasks, by selecting one annotator's entire label set, all-or-nothing, for the entire item. However, there is potential for more accurate aggregation by ``mixing-and-matching'' objects found by different annotators. 
This is accomplished via a {\em partitioning} step, exemplified in Figure \ref{fig:mixnmatch-penguins}, where the mix-and-match aggregate annotation is seen to be superior to the more rudimentary all-or-nothing approach.

To conceptually map out the range of options available with decomposition and partitioning, we describe below a set of alternative paths that can be pursued to aggregate complex annotations, depending on whether the annotations can be merged or partitioned (see Table \ref{table:typesandmethods}). We first summarize the different options available (immediately below), then describe each option in the context of the alternatives (below that).

For single-object annotation tasks (e.g., aggregating a set of rankings), options include:
\begin{itemize}
\item Direct Selection
\item Decomposition, Merge, and Recomposition (DMR)
\end{itemize}

With multi-object annotation tasks (e.g., identifying all of the named-entity spans in a text), the choices further expand to:
\begin{itemize}
\item Direct Selection
\item Partition, Selection, and Recombination (PSR)
\item Partition, Decomposition, Merge, Recomposition, and Recombination (PDMRR)
\end{itemize}

Figure \ref{fig:partitiondiagram} illustrates our framework via an example. Assume we receive a set of annotator labels from each of two annotators $A$ and $B$ to aggregate for a multi-object text sequence annotation task on document $Doc1$. The shortest path of {\bf Direct Selection} (branching to the left) is to simply select one of the annotator's labels, all-or-nothing, for the entire document. Another option (branching to the right) is to partition the annotator labels corresponding to each object, then aggregate labels independently within each partition. If we follow this later path, we reach another fork in the road. Within each partition, the {\bf PSR} path from this fork will select one of the annotator's labels (branching to the left) and then recombine the selected labels across partitions (arrow down). Alternatively, within each partition the {\bf PDMRR} path will aggregate complex labels by decomposition (branching to the right from Partition), merge primitives (arrow down), recompose the complex labels (final arrow down), and then recombine the merged labels across partitions (this recombination is not shown in Figure \ref{fig:partitiondiagram}). In this article, we do not consider selecting labels in some partitions while merging in others; future work might explore intelligently deciding when to select vs.\ merge on a case-by-case basis.

\begin{figure}
\centering
\includegraphics[scale=.75]{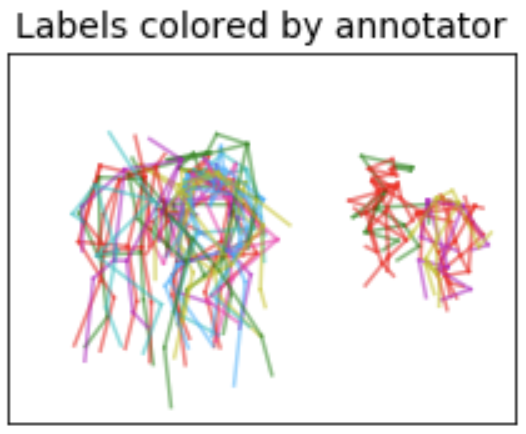}
\includegraphics[scale=.75]{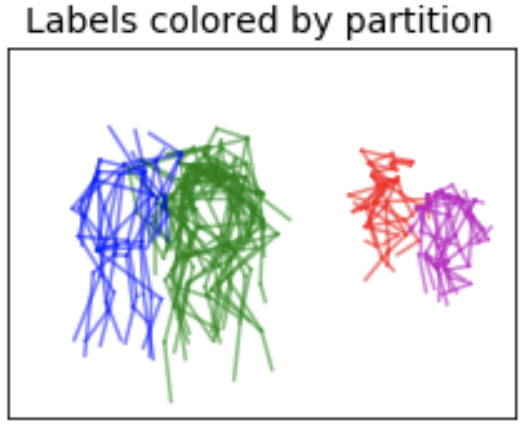}
\caption{An illustration of the matching problem with multi-object annotation tasks. Here, keypoints should be marked for each of four people shown. The left figure shows keypoint labels colored by annotator (each color represents an annotator), while the right figure shows labels corresponding to the same underlying keypoint object assigned the same color after clustering (each color represents an estimated object in the image).} 
\label{fig:labels}
\end{figure}

With multi-object annotation tasks, we expect annotator labels to cluster around each object being annotated,  forming natural clusters we can induce from the labels. Figure \ref{fig:labels} illustrates this for keypoints.
We adopt simple 
\emph{agglomerative
clustering}, a form of hierarchical clustering \cite{ward1963hierarchical}
wherein each observation is initialized in its own cluster before being merged into gradually larger clusters according to which pairs of clusters minimize a given linkage criterion (average distance in our case). The input to this clustering algorithm is the distance matrix between labels for each object. As before, we assume that the user supplies an appropriate distance function for each annotation type (Section \ref{sec:distance}). For each item, we set the number of clusters based on the maximum number of objects found by any annotator. Clusters of size one are removed as outliers. Given scarcity of development data, we leave experimentation of alternative clustering methods for future work.


For error analysis only, we also consider an upper-bound \emph{partition oracle} that uses known gold labels as cluster centers, assigning each label to the cluster associated with the nearest gold label.
In Figure \ref{fig:labels}, for example, the oracle would use the known gold skeletons for the four figures to assign each of the annotated skeletons to the closest gold one, as opposed to using unsupervised clustering.
Comparing accuracy achieved using actual clustering vs.\ this upper-bound oracle enables us to assess the degree to which clustering errors impact overall accuracy. Once clusters are induced, we adopt a {\em divide-and-conquer} approach to aggregate multi-object annotations: 1) partition labels into clusters; 2) perform aggregation on each cluster independently; and 3) output the aggregated label for each cluster.

\begin{algorithm}[]
    \SetKwInOut{Input}{Input}
    \SetKwInOut{Output}{Output}
    
    \Input{
        Single-object labels $L = l_1, l_2, ...$ \\
        Label deconstructor $\Delta(l) \rightarrow$ key-value map
    }
    \Output{
        Key identifiers $k_1, k_2, ..., k_Q$ \\
        Numeric values $x_1, x_2, ..., x_Q$ \\
    }
    $i = 0$ \\
    \ForAll{$l \in L$} {
        \ForAll{$(k \rightarrow x) \in \Delta(l)$} {
            $k_i = k$ \\
            $x_i = x$ \\
            $i++$ \Comment{flatten keys and values over labels} \\
        }
    }
    \caption{Decomposition}
    \label{alg:decomposition}
\vspace{-0.5em}
\end{algorithm}

\begin{algorithm}[]
    \SetKwInOut{Input}{Input}
    \SetKwInOut{Output}{Output}
    
    \Input{
        Key identifiers $k_1, k_2, ..., k_Q$ \\
        Numeric values $x_1, x_2, ..., x_Q$ \\
        Optional weights $w_1, w_2, ..., w_Q$ \\
        Merge function $M(X, W)$  \Comment{weighted median}
    }
    \Output{
        Unique keys $K = k_1, k_2, ..., k_P$ \\
        Merged values $V = v_1, v_2, ..., v_P$ \\
    }
    $K$ = Set($k_1, k_2, ..., k_Q$) \Comment{unique keys, e.g. "upper-left"} \\
    \ForAll{$\{k_h \in K\}$} {
        $I_h = \{i \mid k_i = k_h\}$ \Comment{all indices for key h} \\
        $v_h = M(\{x_i \mid i \in I_h\}, \{w_i \mid i \in I_h\})$ \Comment{merged value for h}
    }
    \caption{Merge}
    \label{alg:merge}
\vspace{-0.5em}
\end{algorithm}

\begin{algorithm}[]
    \SetKwInOut{Input}{Input}
    \SetKwInOut{Output}{Output}
    \Input{
        Unique keys $K = k_1, k_2, ..., k_P$ \\
        Merged values $V = v_1, v_2, ..., v_P$ \\
        Label constructor $\Gamma()$ \Comment{input key-value map}
    }
    \Output{
        Merged single-object label $\mathbf{\hat{l}}$
    }
    $\mathbf{\hat{l}} = 
        \Gamma(\{(k_i \rightarrow v_i) \mid i \in 1, 2, ..., P\})
    $
    \Comment{constructed label} \\
    \caption{Recomposition}
    \label{alg:recomposition}
\end{algorithm}

\begin{algorithm}[]
    \SetKwInOut{Input}{Input}
    \SetKwInOut{Output}{Output}
    
    \Input{
        Multi-object annotations $A_1, A_2, ..., A_N$ \\
        \Comment{each $A$ is a set of labels $l$} \\
        Ground truth $G$ (Oracle ONLY)
    }
    \Output{
        Single-object labels $L = l_1, l_2, ...$ \\
        Map $M(l)$ of labels to item partitions \\
    }
    $L = \bigcup \{ A_1, A_2, ..., A_N \}$
    \Comment{set of all labels} \\ 
    $\hat{C} = max(\|A_1\|, \|A_2\|, ..., \|A_N\|)$ \Comment{max clusters} \\
    \If{ $G$ }
    {$M$ = PartitionOracle($L, G$) \Comment{$|G|$ partitions}  }
    \Else
    {$M$ = AgglomerativeClustering($L, \hat{C}$)  \Comment{Section~\ref{sec:multi-object}}
    }
    \caption{Partition}
    \label{alg:partition}
\vspace{-0.5em}
\end{algorithm}

\begin{algorithm}[]
    \Input {
        Annotator single-object labels $L = l_1, l_2, ...$ \\
        Ground truth single-object labels $G = g_1, g_2, ...$ \\
        Distance Function $D$ 
    }
    \Output {
        Map $M(l)$ of annotator labels to $|G|$ partitions\\ \\
    }
    \ForAll{$l \in L$} {
        $p = \textrm{argmin}(\{D(l, g) \mid g \in G \})$ \Comment{assign $l$ to nearest $g$}  \\
        Add assignment $(l \rightarrow p)$ to $M$ 
    }
    \caption{PartitionOracle}
    \label{alg:oraclepartition}
\vspace{-0.5em}
\end{algorithm}

\begin{algorithm}[]
    \SetKwInOut{Input}{Input}
    \SetKwInOut{Output}{Output}
    
    \Input{
        Selected/merged label for each partition $\hat{l}_1, ..., \hat{l}_M$
    }
    \Output{
        Combined multi-object annotation $\mathbf{\hat{A}}$
    }
    $\mathbf{\hat{A}} = \bigcup \{\hat{l}_1, ..., \hat{l}_M\}$ 
    \caption{Recombination}
    \label{alg:recombination}
\end{algorithm}

\vspace{1em}

\subsection{Partition and Recombination: PSR and PDMRR} 
\label{sec:partition}
\label{sec:recombination}
For selection-based aggregation, we apply {\em Partition, Selection, and Recombination} (PSR), as illustrated in Figure \ref{fig:partitiondiagram}. 
For a given item, the partition step splits all annotator labels into label subsets, where each subset corresponds to a distinct (predicted) object. We then select the (predicted) best label within each partition using selection-based aggregation.  The final aggregated output is then produced by taking the union of aggregated labels across all partitions: one label per partition and its (predicted) object.

With merge-based aggregation, we instead apply {\em Partition, Decomposition, Merge, Recomposition, and Recombination} (PDMRR). This operates the same as PSR except the middle step of aggregation based on selection (S) is replaced by merge-based aggregation using decompose-merge-recompose (DMR) (see Section~\ref{sec:method-decompose}). 


{\bf Algorithm~\ref{alg:partition}} describes our practical partition algorithm using clustering. {\bf Algorithm~\ref{alg:oraclepartition}} describes how we compute the upper-bound, oracle partition using ground truth gold labels. {\bf Algorithm~\ref{alg:recombination}} describes recombination, the union of aggregated labels over partitions.

\subsection{Measuring Distance Between Multi-object Annotations}
\label{sec:distance}

Clustering labels requires a notion of distance between labels. Fortunately, the same distance function used to map complex annotations to annotation distances (Section~\ref{sec:distances}) can be reused here.  
%
However, one subtlety of note relates to multi-object annotation tasks, such as marking text sequences or bounding boxes. Evaluation metrics for such tasks already operate on multi-object annotations, matching a set of model-predicted labels to the set of gold labels corresponding to objects.  The distance function supplied by the task owner should therefore be understood to directly operate on multi-object annotations (i.e., measuring the distance between two sets of bounding boxes). However, such a set-based evaluation function naturally also applies equally to unit sets that contain only a single element. This means that we can also use this same function to measure distance between individual labels as well (e.g., between two individual bounding boxes).

\section{Unit Testing: Is the Model Doing What it is Supposed to?}
\label{sec:unittests}


In this section, we propose several unit tests to assess how well a given unsupervised weighted aggregation model is satisfying expected behavior and minimum requirements for this assessment. Unit testing has a long history in software engineering for assessing whether or not system components are operating as expected, and this idea of unit testing is now being increasingly translated to testing of AI systems, such as via the notion of checklists 
\cite{ribeiro2020beyond,rottger2021hatecheck}. We are not familiar with any prior work investigating such unit testing specifically for aggregation models.

Related to unit testing is the practice of statistical model checking to assess the fit of a model to data. Model checking can involve several approaches:
\begin{enumerate}
    \item Simulation-based calibration
    \item Out-of-sample predictive evaluation
    \item Prior and posterior predictive checks (for Bayesian models)
\end{enumerate}
\citeA{gelman2013bayesian} provide a thorough overview of these kinds of fitness tests for probabilistic models. Furthermore, because we implement our probabilistic models in the Stan language heavily based on their work, we automatically get convergence diagnostics that can be early warnings for model mis-specification issues.
The checks for simulation-based calibration and out-of-sample prediction are conducted in our synthetic data studies, while further out-of-sample prediction is evaluated on our real data experiments.
These checks can generally catch mis-specification problems with probabilistic models, but they do not necessarily catch all the ways an aggregation model might not be capturing what it is supposed to.

Concretely, we propose a suite of unit tests inspired by the above for checking correctness of aggregation models.
The first two of these tests only clearly show desired results asymptotically, that is, for large enough data. Tests can be run on real or synthetic data, where simulated settings provide greater control and knowledge of these configurations. Some of these tests are only relevant for distance-based models, making them inapplicable to many of the classical models, such as
\citeauthor{dawid1979maximum} \citeyear{dawid1979maximum}.
Some of the tests are only relevant for Bayesian models as they involve checking that priors are applied correctly. Some models like MAS may have all six tests applied. It is important that as many of these tests be conducted as are applicable to a model. Failure in any one of them can be a serious sign of model mis-specification or low data quality.

Unit tests we propose are summarized in Table \ref{table:unittests} and the sections below.
Many of the tests are based on the Pearson correlation coefficient $\rho$:
$$ \rho(X,Y) = \frac{N\sum{xy}-(\sum{x}\sum{y})}{\sqrt{ [N \sum{x^2}-(\sum{x})^2 ][N \sum{y^2}-(\sum{y})^2 }]} $$
Two of the tests measure instead a standard deviation $\varsigma$:
$$
\varsigma(X) = \sqrt{\frac{1}{N} \sum (x - \frac{1}{N}\sum x)^2}
$$

\begin{table}[]
\centering
\begin{tabular}{llll}
Test & Model Type & Asymptotic & Measure \\ \hline
\ref{sec:unittestBAU} Worker error well-estimated 

& ALL
& YES & $\rho$\\
\ref{sec:unittestWgtVote} Model worker error affects aggregation

& ALL
& YES & $\rho$\\
\ref{sec:unittestDist} Model fits distances

& Distance
& NO & $\rho$\\
\ref{sec:unittestGold} Model produces healthy predictions

& Distance
& NO & $\varsigma$\\
\ref{sec:unittestScarce} Parameter distributions follow scarcity

& Bayesian
& NO & $\varsigma$\\
\ref{sec:unittestSAD} Weight confidence is controllable

& Bayesian
& NO & $\rho$\\
\end{tabular}
\caption{Unit tests assessing how well a candidate model is satisfying the minimum necessary requirements for an unsupervised weighted aggregation model. Model Type indicates whether the test is only necessary for specific kinds of models: \emph{Distance} models attempt to fit observed distances between observed labels and unobserved gold, and \emph{Bayesian} models attempt to allow uncertainty from data sparsity affect their parameters and predictions. Note that these are not mutually exclusive, models can be both distance-based and Bayesian, as is the case for MAS. Asymptotic denotes whether the test satisfaction threshold grows tighter with the size of data. Asymptotic tests may require simulators to produce data in enough abundance to meet the test conditions.}
\label{table:unittests}
\end{table}

\subsection{ Is Worker Error Well-estimated? }
\label{sec:unittestBAU}

\begin{table}[h!]
\centering
\bgroup
\def\arraystretch{1.5}
\begin{tabular}{r|l}
$\textrm{INPUT }$ & 
$\varepsilon^{BAU}, \gamma^{MAS} $\\
\hline
$\textrm{PASSING CRITERIA } $ & 
$ \rho(\{\varepsilon^{BAU}_u, \gamma^{MAS}_u : u \in U \}) \geq 0.5 $\\
\end{tabular}
\egroup
\caption{Unit Test \ref{sec:unittestBAU} Worker error well-estimated (non-simulated data).}
\label{table:unit-test-worker-error-1}
\end{table}

Weighted aggregation models rely on being able to use good estimates of annotator reliability as label weights. We propose two tests for this, one using BAU and one using simulated worker error.
BAU is meant to be a reasonable single-iteration estimate of worker error. An aggregation model's estimates of worker error will likely differ from BAU, because they are estimated jointly with other model parameters. Still, having a zero or negative correlation between the model's predicted worker error and BAU should raise suspicion. If it happens repeatedly on different datasets, it is likely the model is either mis-specified or the data is very challenging (see ``difficult skew'' in Section \ref{sec:error-skew}).
The correlation expected for this test is shown in Table \ref{table:unit-test-worker-error-1}.
Figures \ref{fig:model-checks-ranking}-\ref{fig:model-checks-binary} (upper left) show the results of this test against BAU for an example dataset.

 To check if the cause is model mis-specification, the model can be run on robust simulated data to confirm whether the model achieves close to the expected amount of correlation between predicted and actual worker error.
 This would be a kind of simple posterior predictive check \cite{gelman2013bayesian} based on simulation-based calibration.
 We describe in Section \ref{sec:error-corr} various simulator experiments and the relationship between different configurations and the resulting correlations between predicted and actual worker error. If simulating with worker error $\sigma_u$ this way, the correlation to check is shown in Table \ref{table:unit-test-worker-error-2}.

\begin{table}[h!]
\centering
\bgroup
\def\arraystretch{1.5}
\begin{tabular}{r|l}
$\textrm{INPUT }$ & 
$\sigma, \gamma^{MAS} $\\
\hline
$\textrm{PASSING CRITERIA } $ & 
$ \rho(\{\sigma_u, \gamma^{MAS}_u : u \in U \}) \geq 0.5 $\\
\end{tabular}
\egroup
\caption{Unit Test \ref{sec:unittestBAU} Worker error estimated well (simulated data).}
\label{table:unit-test-worker-error-2}
\end{table}

As for what values for this correlation are tolerable, we suggest around 0.5 or higher as being ``healthy'', although between 0 and 0.5 can also be normal for sparser datasets. These targets are based on extensive simulation studies (see Section \ref{sec:results-sim} and Table \ref{Tab:resultssim}).

\subsection{Model Worker Error Affects Aggregation}
\label{sec:unittestWgtVote}
A difference in functionality between weighted and unweighted aggregation models is that the weighted models should give additional preference to labels from workers that are estimated to be more reliable. 
Both weighted and unweighted models can produce estimates $\varepsilon$ of label quality. For example, the unweighted SAD score for a label is its average distance to each other label for that item, with lower average distances indicating higher quality. Likewise, a weighted model such as MAS can produce a predicted distance to gold, with lower distances indicating higher quality. The differences in the quality between weighted and unweighted models, for each given label, should correlate with the estimated annotator reliability for that label. As opposed to the prior test (Section \ref{sec:unittestBAU}) of whether the model is accurately estimating  weights, this test is for whether the model is applying those weights correctly.
This test calculates correlation as shown in Table \ref{table:unit-test-wgt-vote}.

\begin{table}[h!]
\centering
\bgroup
\def\arraystretch{1.5}
\begin{tabular}{r|l}
$\textrm{INPUT }$ & 
$\gamma^{MAS}, \varepsilon^{MAS}, \varepsilon^{SAD} $\\
\hline
$\textrm{PASSING CRITERIA } $ & 
$ \rho(\{\gamma^{MAS}_u, (\varepsilon^{MAS}_{iu} - \varepsilon^{SAD}_{iu}) : u \in U, i \in I \}) \geq 0.2 $\\
\end{tabular}
\egroup
\caption{Unit Test \ref{sec:unittestWgtVote} Model worker error affects aggregation.}
\label{table:unit-test-wgt-vote}
\end{table}

The resulting correlation should be significant, but not necessarily very high, generally above 0.2 according to our simulation studies. It may however be lower in cases of data scarcity (see Section \ref{sec:unittestScarce}) or lower weight confidence (see Section \ref{sec:unittestSAD}).
Figures \ref{fig:model-checks-ranking}-\ref{fig:model-checks-binary} (upper right) show the results of this unit test.

\subsection{Model Fits Distances (For Distance-based Models)}
\label{sec:unittestDist}
Distance-based aggregation models train on observed distances between labels for each item. Such models may produce an estimate for each of these observed distances, as for example MAS does ($\| \mathbf{x}_{iu} - \mathbf{x}_{iv} \|$ in Equation~\ref{eq:likelihood}).

A distance-based aggregation model should explain observed distances between labels for a given item, as that is the primary data used to train the model.
For example, this test on MAS calculates correlation based on embeddings $x$ as shown in Table \ref{table:unit-test-distance}.

\begin{table}[h!]
\centering
\bgroup
\def\arraystretch{1.5}
\begin{tabular}{r|l}
$\textrm{INPUT }$ & 
$\mathbf{x}, D $\\
\hline
$\textrm{PASSING CRITERIA } $ & 
$ \rho(\{ \| \mathbf{x}_{iu} - \mathbf{x}_{iv} \|, D_{i,(u,v)} : u \in U, v \in U, u \neq v, i \in I \}) \geq 0.8 $\\
\end{tabular}
\egroup
\caption{Unit Test \ref{sec:unittestDist} Model fits distances.}
\label{table:unit-test-distance}
\end{table}

If the model fits well, we should expect a high correlation (at least 0.8), since this is the objective function of the model. For MAS, increasing the dimensionality configuration should improve this correlation if it is too low.
If using a Bayesian model, this test can be converted into a posterior predictive check \cite{gelman2013bayesian} by sampling the distribution of $\| \mathbf{x}_{iu} - \mathbf{x}_{iv} \|$.
If testing on a model developed differently from MAS, for example one that does not use embeddings, the term $\| \mathbf{x}_{iu} - \mathbf{x}_{iv} \|$ can be replaced with whatever the model estimates to be the distance between annotations by $u$ and $v$.
Figures \ref{fig:model-checks-ranking}-\ref{fig:model-checks-binary} show in the lower left the results of this unit test.

\subsection{Model Produces Healthy Predictions}
\label{sec:unittestGold}

\begin{table}[h!]
\centering
\bgroup
\def\arraystretch{1.5}
\begin{tabular}{r|l}
$\textrm{INPUT }$ & 
$ \varepsilon^{MAS}$\\
\hline
$\textrm{PASSING CRITERIA } $ & 
$ \varsigma(\{ \varepsilon^{MAS}_u : u \in U \}) 
> 0.05 $\\
\end{tabular}
\egroup
\caption{Unit Test \ref{sec:unittestGold} Model produces healthy predictions.}
\label{table:unit-test-gold}
\end{table}

\begin{figure}[t]
\centering
  \includegraphics[scale=.5]{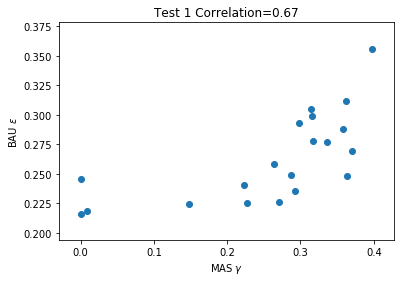}
  \includegraphics[scale=.5]{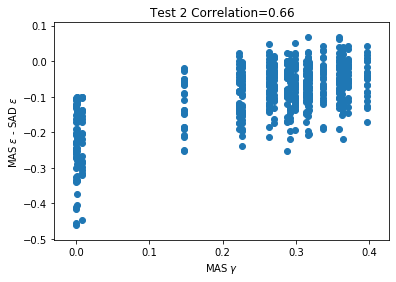}
  \includegraphics[scale=.5]{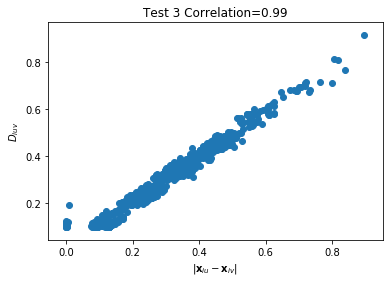}
  \includegraphics[scale=.5]{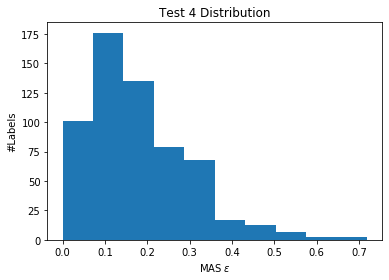}
\caption{Results of unit tests on an example ranked list annotation simulator run. We see some near-zero $\gamma$s in Unit Test \ref{sec:unittestBAU} which might be investigated, though overall correlation 
is significantly positive. Some near-zero $\gamma$s in Unit Test \ref{sec:unittestWgtVote} contribute to correlation; MAS will put much more weight on those high-reliability annotators. High correlation in Unit Test \ref{sec:unittestDist} and non-degenerate distribution in Unit Test \ref{sec:unittestGold} passes visual inspection. }
\label{fig:model-checks-ranking}
\end{figure}

\begin{figure}[t]
\centering
  \includegraphics[scale=.5]{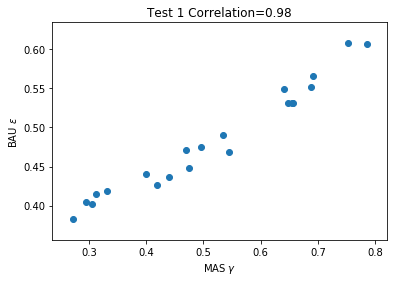}
  \includegraphics[scale=.5]{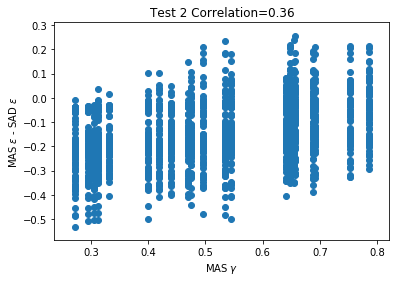}
  \includegraphics[scale=.5]{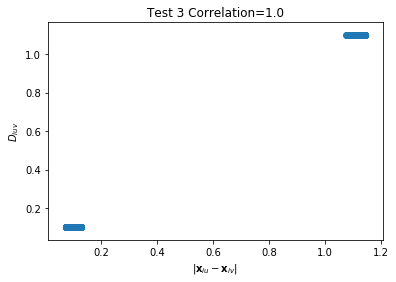}
  \includegraphics[scale=.5]{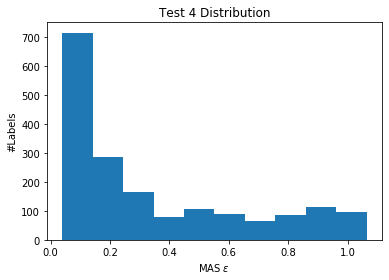}
\caption{Results of unit tests on an example binary annotation simulator run. We observe very tight correlation in Unit Test \ref{sec:unittestBAU}, and correlation appears strong in Unit Test \ref{sec:unittestWgtVote}. Correlation passes Unit Test \ref{sec:unittestDist} (there are only two clusters because it is binary data). Distribution in Unit Test \ref{sec:unittestGold} is non-degenerate.}
\label{fig:model-checks-binary}
\end{figure}

An aggregation model should produce estimates of label quality. For a distance-based model this could be the predicted distance $\varepsilon$ from gold. The distribution of predicted distances can indicate trouble. For example, if there is no variation, with the predictions all clustering around zero, this could indicate that the model is mis-specified or the algorithm failed. The test described in Table \ref{table:unit-test-gold} ensures that the predictions vary by more than a negligible standard deviation (0.05). Other strange distributions that are dubious characterizations of label quality may also be warnings that something is off, so it is recommended that prediction histograms also be checked visually. Figures \ref{fig:model-checks-ranking}-\ref{fig:model-checks-binary} show in the lower right the results of this unit test on example data.

\begin{figure}[t]
\centering
  \includegraphics[scale=.5]{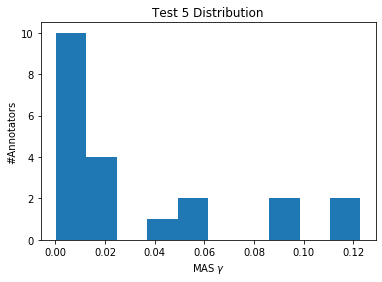}
  \includegraphics[scale=.5]{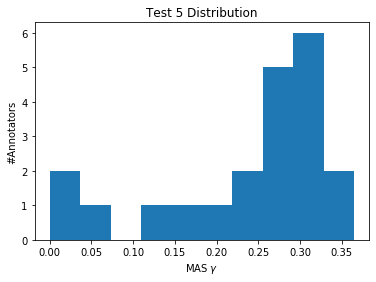}
  \includegraphics[scale=.5]{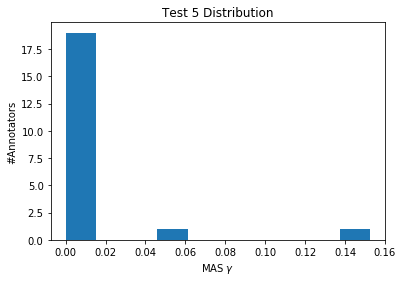}
  \includegraphics[scale=.5]{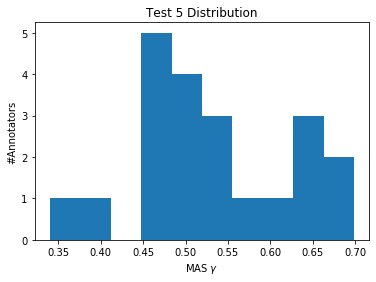}
\caption{Results of Unit Test \ref{sec:unittestScarce} on ranked list (top) and binary (bottom) annotations. Histograms on the left show the results of dataset with 2 annotations per item, whereas those on the right show the results of having 6 annotations per item. As expected, with fewer annotations per item, the Bayesian MAS model has a much narrower distribution in the annotator reliability parameter $\gamma$, with data scarcity causing shrinkage toward the prior. With a healthier number of annotations per item, the model should find more varying estimates of the reliability parameter, as on the right.}
\label{fig:model-checks-5}
\end{figure}

\begin{figure}[t]
\centering
  \includegraphics[scale=.5]{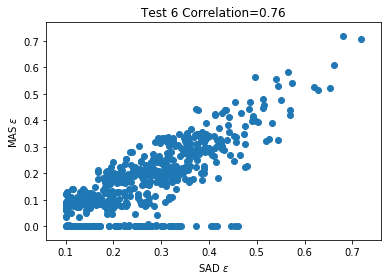}
  \includegraphics[scale=.5]{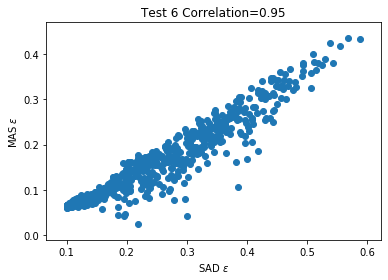}
  \includegraphics[scale=.5]{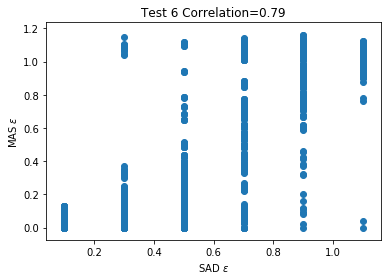}
  \includegraphics[scale=.5]{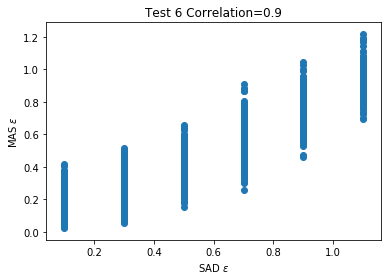}
\caption{Results of Unit Test \ref{sec:unittestSAD} on ranked list (top) and binary (bottom) annotations. Scatterplots on the left show the results of setting MAS $\Phi$ to 0.25. Those on the right show the results of setting MAS $\Phi$ to 0.01, effectively tightening the prior on annotator reliability and forcing it to act more like unweighted SAD.}
\label{fig:model-checks-6}
\end{figure}

\begin{table}[h!]
\centering
\bgroup
\def\arraystretch{1.5}
\begin{tabular}{r|l}
$\textrm{INPUT }$ & 
$ \gamma^{MAS}(\textrm{Sim1}), \gamma^{MAS} (\textrm{Sim2})$\\
\hline
$\textrm{PASSING CRITERIA } $ & 
$ \varsigma(\{ \gamma^{MAS}_u(\textrm{Sim1}) : u \in U \}) 
<  \varsigma(\{ \gamma^{MAS}_u(\textrm{Sim2}) : u \in U \}) $\\
\end{tabular}
\egroup
\caption{Unit Test \ref{sec:unittestScarce} Parameter distributions follow scarcity. This test requires comparing two simulated datasets, Sim1 and Sim2, with fewer/more workers per item, respectively.}
\label{table:unit-test-shrinkage}
\end{table}

\begin{table}[h!]
\centering
\bgroup
\def\arraystretch{1.5}
\begin{tabular}{r|l}
$\textrm{INPUT }$ & 
$\varepsilon^{SAD}, \varepsilon^{MAS(\Phi=0.25)}, \varepsilon^{MAS(\Phi=0.01)} $\\
\hline
$\textrm{PASSING CRITERIA } $ & 
$ \rho(\{ \varepsilon^{SAD}_u, \varepsilon^{MAS(\Phi=0.25)}_u : u \in U \}) 
< \rho(\{ \varepsilon^{SAD}_u, \varepsilon^{MAS(\Phi=0.01)}_u : u \in U \}) $\\
\end{tabular}
\egroup
\caption{Unit Test \ref{sec:unittestSAD} Weight confidence is controllable.}
\label{table:unit-test-phi}
\end{table}

\subsection{Parameter Distributions Follow Scarcity (Hierarchical Bayesian Aggregation)}
\label{sec:unittestScarce}

As data becomes scarce, the signal-to-noise ratio decreases and parameter estimates for annotator reliability should shrink toward a prior \cite{kruschke2014doing}. Doing so allows the model to behave more like majority vote when confidence in worker reliability is low. Otherwise, high-confidence but low-accuracy parameter estimates for the label weights will cause the aggregation to behave more like it is choosing a random worker (see RU in Section~\ref{sec:baselines}), which will likely be worse than other aggregation methods such as majority vote.
Figure \ref{fig:model-checks-5} shows results of this unit test on simulated data, varying the number of annotations per item. Generally, variation in the annotator reliability parameter should increase noticeably as the annotations per item goes from very scarce (two in this example) to a healthier number (six in this example). This unit test is described in Table \ref{table:unit-test-shrinkage}.

\subsection{Weight Confidence is Controllable (Hierarchical Bayesian Aggregation)}
\label{sec:unittestSAD}
MAS has a configurable parameter $\Phi$ for the prior on annotator reliability. Setting this to be extremely narrow forces annotators to effectively share a reliability parameter. Under this setting, MAS should behave almost identically to SAD. In general, hierarchical Bayesian models for label aggregation place a prior on the worker-level parameters that can be completely pooled, partially pooled, or unpooled \cite{carpenter2008multilevel}, \cite{simpson2013dynamic}, \cite{venanzi2014community}, \cite{moreno2015bayesian}, \cite{paun2018comparing}, \cite{paun2022statistical}. Forcing the scale of this prior to be zero effectively causes the worker-level parameter to be fully pooled across workers, disabling the ``weighted'' aspect of the aggregation. When doing so, if the model does not behave very similarly to SAD, it is a sign that the model is mis-specified.
Figure \ref{fig:model-checks-6} shows the results of this unit test across two simulated datasets.
Generally, this correlation between SAD and MAS should increase as $\Phi$ goes to zero, and vice-versa, since the $\Phi$ scale is what allows MAS to vary the annotation weights. Shown in Table \ref{table:unit-test-phi} is a version of this test that compares a $\Phi$ of 0.25  to a $\Phi$ of 0.01.


\section{Tasks, Datasets, and Evaluation Metrics}
\label{sec:datasets}

In this section, we describe the simple and complex annotation tasks, datasets, and metrics that we report on in our evaluation.  Following this, 
Section~\ref{sec:methods_compared} describes the methods we compare, with results presented in Section \ref{sec:results-sim} (synthetic data) and Section \ref{sec:results-real} (real data).

\subsection{Simple Annotations}
\label{data:simple}

\subsubsection{Tasks}
Below we describe four task types we consider in benchmarking label aggregation models. 

    \textbf{Categorical} tasks (binary or multi-class). All items are assigned one category from a fixed set of  options, permitting modeling of categorical confusion matrices.
    
    
    \textbf{Ordinal} (rating) tasks also give annotators a fixed set of categories from which to choose, for each item. However, unlike categorical tasks, ordinal categories relate to one another via consecutive ordering of level or degree (e.g., likert scale subjective judgments or rating movies or products). Though ordinal tasks can be modeled as categorical, doing so treats all errors as equal and ignores relations between categories. 

    \textbf{Multi-choice} tasks resemble categorical tasks in that an annotator selects an answer from a set of choices. However, unlike categories, the set of answer choices are unrelated across items, meaning one cannot model confusion matrices across answers. Such multiple choice ``tests'' are relatively rare and most commonly employed when evaluating annotators based on known answers. A practical use case is when a model automatically predicts the top 4 or 5 answers to a question, randomizes these to avoid ranking bias, and then uses human computation to select the best answer \cite{yan2010crowdsearch,rodriguez2011crowdsight,savenkov2016crqa}.
    
    \textbf{Numerical} tasks ask annotators to answer with a (typically real-valued continuous) number, rather than select from a discrete set of choices.


\subsubsection{Real Data}
Below we list eleven simple annotation task datasets we report on in our benchmarking:

\begin{itemize}
	\item\textbf{Movie Review Sentiment} (binary). Participants rate a movie review as positive or negative \cite{venanzi2015sentiment}. 
	\item\textbf{Temporal Ordering} (binary). Annotators decide if an event happened before or after another. This dataset is based on a subset of the TimeBank corpus \cite{pustejovsky2003timebank}, which provided gold labels aggregated by five linguistics experts and reviewed by undergraduate computer science students \cite{snow2008cheap}. 
	\item\textbf{Recognizing Textual Entailment (RTE)} (binary). Annotators are shown two sentences and must decide whether one sentence (hypothesis) can be inferred from the other (premise) \cite{snow2008cheap}. 
	\item\textbf{Dog Species} (multi-class). Participants select the appropriate dog breed for dog pictures. Gold labels were based on a consensus among nine experts \cite{zhou2012learning}.
	\item\textbf{Face Sentiment} (multi-class). Face photos are classified as neutral, happy, sad, or angry. The dataset's authors annotated gold labels by hand \cite{mozafari2014scaling,mozafari2012active}.
	\item\textbf{Face Sentiment Scrambled (FaceMC)} (multi-choice). To prevent modeling of categorical confusion matrices, we transform the original multi-class categorical task into a multi-choice selection task, prefixing each category with an item-specific identifier. Annotators still answer the same items correctly or not, but by remapping the category values, we break the categorical relationship across items. 
	\item\textbf{Word Sense Disambiguation (WSD)} (multi-choice). Given a word used in a passage of text, the participant must choose the correct sense of the word from three other passages of text. This task is multi-choice because the answer choices are different for each item. As \citeA{snow2008cheap} noted, most annotations are correct (skewed distribution). We further note that the first answer is correct for most questions. Gold labels were provided by linguistics students, and annotators examined and resolved disagreements \cite{pradhan2007semeval}.
	\item\textbf{Weather Sentiment} (Ordinal). Participants rate tweets about the weather as negative, neutral, positive, `not related,' or `I can't tell.' Only the first three labels have a natural ordering, so we follow the same practice as \citeA{lakshminarayanan2013inferring} and remove items with nominal values \cite{crowdflower2013sentiment}.
 
	\item\textbf{Adult Content} (Ordinal). The explicitness of content is rated on a four-point ordinal scale (G, P, R, or X). \cite{ipeirotis2010quality}. 
	\item\textbf{Emotion Score} (Numerical). Participants rate the intensity of emotion on a face on a scale from -100 to 100 \cite{snow2008cheap}.
	\item\textbf{Word Similarity} (Numerical). Similarity between  words is given a continuous score from 0-10 \cite{snow2008cheap}.
	\item\textbf{Population} (Numerical). This dataset was collected by aggregating Wikipedia edit histories for the populations of cities. As noted by \citeA{li2014confidence}, most annotators only answer a few items and most items are only answered by a handful of annotators. Gold labels were taken from U.S. Census data. After removing extreme outliers, values for the annotations range from 0 - 2048446 \cite{pasternack2010knowing}. 
\end{itemize}

\subsubsection{Synthetic Data: Binary classification}
\label{data:binary}

We report a binary label data simulator that randomizes gold labels, worker abilities, and worker labels based on the gold values and worker abilities.

We generate $N$ items with binary gold labels $g_i$ assigned uniformly at random (though Bernoulli probability $p$ is configurable in the simulator). We also generate $J$ workers and assign $r J$ workers to each item, with $r$ representing the fraction of workers drawn for each item (with replacement such that some workers annotate more items than others). Each worker $u$ is assigned an error rate $\sigma_u$ drawn from a configurable Beta($\alpha$, $\beta$) distribution with support between 0 and 1, as illustrated in Figure \ref{fig:beta-dists-reliability}. Finally, each worker generates a binary label for each of their assigned items with probability = $\sigma_u$ that it is incorrect (and probability $1 - \sigma_u$ that it is correct). In other words, the workers' error rates represent their probability of flipping the label from the item's ground truth.

\subsubsection{Evaluation Metrics} \label{sec:evaluation_metrics_simple}

We report a single metric for each task. Although any arbitrary evaluation metric could be used in practice, most of the metrics we adopt are defined in the range $[0,1]$ to measure annotation quality (larger is better). Exceptions are noted below.

\textbf{Categorical.} Both the distance function and evaluation metrics compare values by exact match (0/1 loss). We score binary tasks by accuracy and traditional F1, while  multi-class categorical tasks are scored with accuracy and a weighted F1.
	
\textbf{Ordinal (rating).} Prior work on annotation modeling does not set a clear precedent for how to evaluate ordinal labels. Some prior work  \cite{hovy2013learning,dawid1979maximum,zheng2017truth,raghu2019direct} treats ordinal data as categorical, ignoring ordering relations between categories. In other ordinal classification work, Mean Squared Error (MSE) and Mean Absolute Error (MAE) are widely used \cite{cardoso2011measuring,baccianella2009evaluation,lakshminarayanan2013inferring} and preferred over accuracy and error rate metrics, which penalize all errors equally \cite{gaudette2009evaluation}. 
We report MAE for two reasons: it translates to simple distance and evaluation functions and it penalizes the number of errors more than the magnitude of the errors. MAE is lower-bounded at zero with lower values being better.
	
\textbf{Multi-choice.} We treat multi-choice selection tasks similarly to categorical tasks; since the answer choices do not have a natural ordering, we do not assign partial credit. For multi-choice selection problems, we only report accuracy, since F1 assumes all categories are the same between items.
	
\textbf{Numerical.} Regarding evaluation metrics, RMSE, MSE, and MAE are all widely used; we report MAE for consistency with ordinal tasks.

\subsection{Complex Annotation Tasks: Synthetic Data}

\subsubsection{Ranking Elements}
\label{data:rankings}

This task involves ranking a set of top elements for each question (item). For example, users might be asked to name and sort the ten largest countries by population, best-selling fiction books by sales volume, or richest people in the world.

We assume $N$ such items, each having 50 elements, and respondents needing to identify and rank the top 10 elements for each item. Our simulator generates a ``true score'' $g_e$ for each element $e$ in an item from a standard normal distribution, and a gold ranking over elements for each item is induced from these scores.

Top-10 rankings for each of $r J$ workers assigned to each item are simulated by sorting the top ten elements of that item by the worker's ``perceived score''.
The perceived score is drawn from a normal distribution with location = $g_e$ and scale = $\sigma_u \sigma_i$. These $\sigma$ parameters simulate worker skill and item difficulty. To simulate variation over users and items, each $\sigma_u$ is drawn from a configurable Beta($\alpha$, $\beta$) distribution.




\subsubsection{Marking Keypoints}
\label{data:keypoints}

Keypoints 
are an ordered sequence of vertices, where each corresponds to a particular node in the \emph{skeleton} defined for the keypoint class \cite{lin2014microsoft}. For example, a given task might specify that annotators mark coordinates for the "head", "neck", "shoulders", and so on of every human body in each image.
Each vertex may alternatively be marked as not visible and not given coordinates. Since there may be few or many entities (e.g., people) in the image to be detected, there may be one to many keypoint annotations produced per annotator for each image.

We simulate noisy keypoint annotations from COCO \cite{lin2014microsoft} gold labels. 
We randomly select $N$ items having $>=4$ gold objects.
$J$ simulated annotators are each assigned a ''randomness'' parameter $\sigma_u$ drawn from a Beta($\alpha$, $\beta$) distribution. Each item is then assigned $r J$ annotators. Annotator labels are simulated by corrupting gold keypoints via the randomness parameter in several ways: 1) translating vertices horizontally and vertically; 2) rotating the  keypoint; 3) re-scaling the keypoint; and 4) complete omission. 
 

\subsubsection{Constituency Parsing (i.e., Syntactic Tree Induction)}
\label{data:parses}

Syntactic parsing represents a challenging annotation task which has traditionally required trained linguists. The task has attracted great attention in the NLP community, and syntax trees clearly represent complex annotations. Such a difficult annotation task could reveal varying abilities with even trusted annotators which might be usefully modeled. Finally, given aggregation modeling support, we can envision ambitious crowdsourcing task designers pushing the envelope to engage the crowd in more complex tasks like this.

We focus specifically on constituency parsing, as embodied in the Penn Treebank (PTB) \cite{marcus1993building}. 
We randomly sample sentences of length 10 or more from PTB's Brown corpus \cite{francis1979brown}. We employ a diverse set of automatic parsing models included in NLTK \cite{loper2002nltk}: the Charniak parser \cite{mcclosky2006effective}, MaltParser \cite{nivre2007maltparser}, and the Stanford Parser \cite{manning2014stanford}. The purpose of having these different parsers is simply to introduce diversity in the kinds of noise affecting the simulated data. From each parser we generate a $k$-best set of candidate parses per sentence.  
Next, we evaluate the quality of each candidate parse vs.\ PTB's gold parse by the \emph{EVALB} metric \cite{sekine1997evalb}. Finally, we merge all model output parses into a single ranking, ordered by decreasing EVALB score.

We simulate varying annotator accuracy by assigning each annotator $u\in U$ an overall accuracy parameter $a_u$ from a beta distribution. We define two configurations for setting this parameter. In the ``basic'' setting, most annotators will be fairly accurate, sampling $\forall_u \,a_u \sim \text{Beta}(4, 1)$.  In contrast, the ``noisy'' configuration assumes low accuracy workers are more prevalent, sampling  $\forall_u \,a_u \sim \text{Beta}(3, 2)$.

We randomly assign 20\% of the sentences to each annotator. For each sentence $i$ and annotator $u$, we generate an item-level error parameter $e_{iu}$ from the distribution $e_{iu} \sim \text{Normal}(0,0.1)$. The annotator's adjusted accuracy for the given sentence will then be $(a_u + e_{iu})$. Finally, the simulator ``generates'' annotator $u$'s  parse for sentence $i$ by selecting the output model parse having EVALB score $s$ best matching the adjusted accuracy, i.e., minimizing $|s - (a_u + e_{iu}) |$.

\subsection{Complex Annotation Tasks: Real Data}

\subsubsection{Translation} \label{data:translations}

\citeA{li2020crowd} produces three sets of crowdsourced Japanese-to-English translations, together comprising the CrowdWSA2019 dataset \cite{li2019dataset}. They collect an average of about ten answers per question from the crowd, over a total of 450 questions.
The crowd consists mostly of workers who are native speakers of Japanese speakers and non-native speakers of English. They encourage even beginner English speakers to participate and thus collect a dataset of more diverse quality than usually used to train machine translation models. This dataset includes translations and translators of varying ability. Since the dataset was based on pairs of sentences bilingual texts, the existing translation was used for evaluation.


\subsubsection{Sequence Tagging (PICO)} \label{data:sequences}

\citeA{nye2018corpus} and \citeA{nguyen2017aggregating} share an information extraction dataset of 4,800 medical paper abstracts, each annotated by $\sim$6 Mechanical Turk workers on average. Workers were asked to highlight text sequences that identify the population enrolled in the medical study described in each abstract. Medical students annotated a subset of 200 abstracts to create the gold dataset.
Out of the 200 abstracts with available gold, \citeA{nguyen2017aggregating} share outputs from their Crowd-HMM model for 191 of the abstracts. Our final dataset contains 188 abstracts (three removed due to single annotator), with 1152 text sequence annotations by 91 unique workers. Each \emph{sequence} contains multiple \emph{spans} defined by a start and end positional index, and each span captures different properties of the study population being identified.

\subsubsection{Sequence Tagging: Named Entity Recognition (NER)} \label{data:ner}
\citeA{nguyen2017aggregating} also study the CoNLL2003 English Named Entity Recognition dataset \cite{sang2003introduction}, testing on 200 annotations collected by \citeA{rodrigues2014sequence}.  This task is about recognizing and marking named entities in news articles. Compared to the PICO data described above, the items annotated in this dataset are entire news articles rather than relatively shorter abstracts. Therefore, we see many times the number of highlighted sequences (Table \ref{Tab:data}) and fewer annotators per item. Furthermore, in addition to the start and end position of each span, annotators are required to tag the class of each span -- person, location, organization, or miscellaneous. Gold labels were constructed by hand by researchers at the University of Antwerp.

\subsubsection{Drawing Bounding Boxes} \label{data:boundingboxes}

We share a dataset of Mechanical Turk bounding box annotations containing many latent objects per item. 
A total of 400 images were annotated, with half reserved for final testing. Each bounding box is defined by an upper-left and lower-right vertex.  An image may contain one to many entities, leading to one to many bounding boxes per image per annotator. 

\subsubsection{Multidimensional Rating} \label{data:affect}

\citeA{snow2008cheap} provide a dataset in which participants were asked to score short text headlines according to six emotions (dimensions) on a [0-100] integer scale.
While each value within this six-dimensional vector label might be considered a simple numerical label, we categorize vectors as complex because of the combinatorial explosion of possible labels and the added ambiguity involved in extending existing numerical aggregation techniques such as CATD \cite{li2014confidence} and GPM \cite{li2020gpm} discussed in Section \ref{sec:simple-baselines}.

\subsection{Evaluation Metrics (Complex Annotation Tasks)} \label{sec:evaluationmetrics}
\label{evaluation}

As with simple annotation tasks, we report a single metric for each task, and most of the metrics we adopt are defined in the range $[0,1]$ (larger is better). Exceptions are noted.

{\bf Translation.} For evaluating worker translations against gold translations we use the GLEU score \cite{wu2016google}, which is a variant of the BLEU \cite{papineni2002bleu} score but specialized for comparisons between individual sentences. 
We use the NLTK \cite{loper2002nltk}'s implementation.

{\bf Sequence Annotation.} We adopt the same F1 evaluation metric reported by \citeA{nguyen2017aggregating}. They first define textual span-based precision and recall metrics as follows:
\begin{eqnarray}
\text{Precision}~P = \frac{\text{\# true positive tokens}}{\text{\# tokens in labeled span}} \\
\text{Recall}~R = \frac{\text{\# true positive tokens}}{\text{\# tokens in gold span}}
\end{eqnarray}
They then average these span-based $P$ and $R$ metrics over all spans, and report F1 as the harmonic mean of these $P$ and $R$ averages.

\textbf{Bounding Boxes} use \emph{Intersect Over Union} (IoU), a common measure for bounding box tasks \cite{lin2014microsoft}. To evaluate a multi-object annotation $L$ (set of bounding boxes) against a gold set $G$, we take the mean of the per-bounding-box best IoU scores $P(L,G)$ and $R(L,G)$, where:
$$
P(L,G) = \{ \text{max}(\{\text{IoU}(a, g) \mid g \in G \}) \mid a \in L \} \\
$$
$$
R(L,G) = \{ \text{max}(\{\text{IoU}(a, g) \mid a \in L \}) \mid g \in G \}
$$

where IoU compares a single bounding box against another:
$$
IoU(a,g) = \frac{\| \text{area of } a \bigcap  \text{area of } g \|}{\| \text{area of } a \bigcup \text{area of } g \|}
$$

\textbf{Numerical Vectors.} We use RMSE as the distance function and its reciprocal as the evaluation function, which is lower-bounded at zero with lower values being better.

{\bf Parsing.} We use EVALB \cite{sekine1997evalb} for evaluating annotations vs.\ gold.

{\bf Ranking.} We report \citeA{kendall1938new}'s $\tau$ correlation to evaluate the position of elements in annotated ranked lists against their positions in the gold ranked lists.
$\tau$ is defined as
$$\dfrac{C-D}{C+D} = \dfrac{C-D}{{n \choose 2}}$$
where $C$ is the number of concordant pairs, $D$ is the number of discordant pairs, and $n$ is the number of items.
While the gold list contains an exhaustive ranking over all elements, the annotation task is only to rank the top 10 elements. In evaluating annotator rankings, we assume that any element not in the top 10 of an annotator list is assigned the maximum position and thus``tied for last place''.

\textbf{Keypoints} use an \emph{Object Keypoint Similarity} (OKS) score defined in COCO \cite{lin2014microsoft}. Comparing two single keypoints is defined as:
$$
\text{OKS}(a, g) = \frac{\sum_{i} \exp \left(-D(a_i,g_i)^{2} / 2 s^{2} k_{i}^{2}\right) \delta\left(V(g_i)>0\right)}{\sum_{i} \delta\left(V(g_i)>0\right)}
$$
Here, $D(a_i,g_i)$ is the Euclidean distance between node $i$ in the label and node $i$ in ground truth, node $i$ representing for example, the head or shoulder in the defined skeleton structure of the keypoint class. $V(g_i)$ is the visibility of node $i$ in the ground truth, $s$ is the object scale, and $k_i$ is a constant 1 for our simulated keypoints.

For sets of keypoints, the evaluation function compares multi-object annotation $L$ against gold $G$ by the mean of $P(L,G)$ and $R(L,G)$:
$$
P(L,G) = \{ \text{max}(\{\text{OKS}(a, g) \mid g \in G \}) \mid a \in L \} \\
$$
$$
R(L,G) = \{ \text{max}(\{\text{OKS}(a, g) \mid a \in L \}) \mid g \in G \}
$$

\subsection{From Metrics to Distances for Complex Annotations}

As discussed earlier, the models we propose for aggregating complex annotations are agnostic to the distance function used. In general, such distance functions already exist: as long as we can quantify error in comparing predicted labels vs. gold labels (i.e., an evaluation metric), the same error measure can be used as a convenient distance function for our model. We adopt this approach in our study and leave for future work investigation of how alternative distance functions interact with choice of evaluation metric.

All evaluation metrics we report quantify the annotation quality $q$. Since all are  conveniently defined over $[0,1]$, we can induce an error measure $e$ by simply taking $e=1-q$. In general, non-bounded metrics would require empirically identifying the maximum score quality in order to convert from quality to error.

{\bf Translation} is the only task we consider in which the the evaluation metric (GLEU) violates the symmetry requirement for distance functions (Section \ref{sec:distances}). As described there, we apply the general approach of symmetrizing the metric by computing in both directions and averaging, then proceeding as with other metrics.

\begin{equation}
f(x, y) = 1 - 0.5 (\mathbb{GLEU}(x,y) + \mathbb{GLEU}(y, x))
\end{equation}

{\bf Sequence Annotation.} We use $f(x, y) = 1 - \mathrm{F1}(x, y)$. 

{\bf Parsing.} The distance function is $f(x, y) = 1 - \mathrm{EVALB}(x, y)$.

{\bf Ranking.} The distance function is $f(x, y) = 1 - \tau(x, y)$. 

\section{Methods Compared} \label{sec:methods_compared}

\subsection{General Baselines} \label{sec:baselines}

\textbf{Random User (RU)}.
The simplest baseline is to choose a random user's annotation for each item. This represents a scenario in which only a single label were collected per item. We report the RU performance as an average over five trials, sampling with replacement from available labels.

\subsection{Simple Annotation Baselines}
\label{sec:simple-baselines}

Three other trivial baselines predict labels through direct calculations: \textbf{mean}, \textbf{median}, and majority vote (\textbf{MV}). The following baselines optimize via EM \cite{dempster1977maximum} or other approaches. 

Since Dawid and Skene (\textbf{DS}) models a categorical confusion matrix for each annotator, direct application to multi-choice tasks would incorrectly assume consistent answer categories across items. Similarly, modeling arbitrary numerical tasks does not make sense with confusion matrices \cite{lin2012crowdsourcing}. DS can be used with ordinal data but at the cost of ignoring ordering relations between categories.

Multi-Annotator Competence Estimation (\textbf{MACE}) \cite{hovy2013learning} uses EM to learn the probability that each annotator is trustworthy, as well as how each annotator behaves when spamming. MACE supports categorical and ordinal tasks (when ordinal is modeled as a categorical task), but MACE does not support continuous data. 
The model assumes categories are consistent across items, so it is not suited to multi-choice tasks.

	
The Generic Probabilistic Model (\textbf{GPM}) \cite{li2020gpm} models ground truth with the categorical distribution, rather than a single value, and supports all simple labeling tasks.

	
Confidence Aware Truth Discovery (\textbf{CATD}) \cite{li2014confidence} learns annotators `weights' via the probability that each annotator answers an item correctly and the number of items each annotator answers. Like GPM, CATD is versatile enough to handle all types of simple annotation tasks and does not assume categories are consistent across items.

The methods discussed above represent a diverse sampling of prior work, but our list is by no means comprehensive. For example, though ZenCrowd \cite{demartini2012zencrowd} 
and GLAD \cite{whitehill2009whose} are widely used, past benchmarking \cite{zheng2017truth,li2020gpm} found that neither consistently outperforms DS on simple annotation tasks.

\subsection{Task-specific Baselines} 

\textbf{Sequences: Token-wise Majority Vote (TMV)}.
In addition to comparisons against prior work, \citeA{nguyen2017aggregating} report a simple baseline which breaks sequence annotations into individual tokens and then performs a token-wise majority vote. 

\textbf{Sequences: Crowd-HMM (CHMM)}. \citeA{nguyen2017aggregating} proposed a novel bespoke Crowd-HMM probabilistic model for sequence labeling. Using the same dataset that we also adopt in this study, they show empirical improvement of Crowd-HMM over  prior work \cite{rodrigues2014sequence,huang2015}.

Note: \citeA{nguyen2017aggregating} use 4,800 abstracts without gold for unsupervised training, whereas 
%
we limit ourselves to the 191 abstracts shared by the authors with Crowd-HMM model outputs. Our empirical results thus potentially underestimate the relative performance of our distance-based models in comparison to Crowd-HMM. 

\textbf{Bounding Boxes (BVHP)}.
\citeA{branson2017lean} present a joint model of labels, annotator skill, and item difficulty, used for multiple bounding boxes, single keypoints, or binary labels.

\textbf{Translations (HRRASA)}.
\citeA{li2020crowd} provide a model for translations. Theirs is also a distance-based model, but using a wide assortment of natural language distance functions, discussed in more detail in Section \ref{sec:vs-baselines}.

\subsection{Our Methods} 

Our models of annotation distance that select the best annotation for each item consist of the following:

\begin{enumerate}
    \item \textbf{Smallest Average Distance (SAD)}, described in Section \ref{method:sad}, estimates label quality in inverse proportion to average distance from each label to all others within an item, generalizing majority vote.
    \item \textbf{Best Available User (BAU)}, described in Section \ref{method:bau}, estimates label quality in inverse proportion to average distance recorded by each annotator over the entire dataset.
    \item \textbf{Model of Ability, Difficulty, and Distances (MADD)}, described in Section \ref{method:madd}, models distances between annotations without the need for a coordinate space. It models label distances from parameters estimating annotator abilities, item difficulties, and probabilities that each label is closest to ground truth.
    \item \textbf{Multidimensional Annotation Scaling (MAS)}, described in Section \ref{method:mas}, estimates coordinates for each label in a Euclidean space, such that their quality, represented by proximity to the origin, is regularized by estimated annotator error and item difficulty parameters.
    \item \textbf{Semi-supervised Learning, \citeA{braylan2020modeling} (SSL-BL)} is our prior semi-supervised learning approach (called ``S-MAS'' in that paper) that is only usable by the MAS model. It includes known ground truth annotations in the modeled distance dataset but places a tight prior on their associated embedding magnitudes.
    \item \textbf{Semi-supervised Multidimensional Annotation Scaling (SMAS)} uses our new semi-supervised method described in Section \ref{method:semisupervised}. While it is generally applicable to any model weighted by worker reliability, in these experiments we only combine it with the MAS model. To test this approach empirically, we reserve 10\% of items as known-gold and remove them from testing data. Note that for all experiments and all methods tested, these same known-gold items were removed from the test set to ensure fair comparison of unsupervised vs. semi-supervised settings for model estimation.
    \item \textbf{Partition-Select-Recombine using MAS (PSR-MAS)} is one of our methods specialized for multi-object annotations (PICO, NER, and BB). This method uses clustering to partition item annotations into subsets subsequently aggregated by MAS, as described in Section \ref{sec:multi-object}.
    \item \textbf{Partition-Decompose-Merge-Recompose-Recombine using MAS (PDMRR-MAS)} is the other of our methods specialized for multi-object annotations (PICO and BB), which merges elements in the partitioned annotations rather than selecting the best one. Merging is discussed in Section \ref{sec:method-decompose}.
\end{enumerate}

The applicability of the above methods to each dataset is specified in Table \ref{table:typesandmethods}.

\subsection{Oracles}

We define several distinct oracle methods (abbreviated ORC). First, Select-ORC selects the best single worker's labels to keep for each item based on gold labels. PSR-ORC and PDMRR-ORC both use the {\em partition oracle} to partition the set of all annotator labels for each item (i.e., each gold item defines a partition, and each label is assigned to the partition of its nearest gold label). Given these oracle partitions, PSR and PDMRR proceed otherwise as usual, predicting the best annotator's label to select (PSR) or merging all labels (PDMRR). Note that unlike Select-ORC, PSR-ORC and PDMRR-ORC are not given access to the best label, just the best partition, so Select-ORC is expected to be the most powerful.

\begin{table}[]
\centering
\begin{tabular}{r|rrr}
Data type	&	\multicolumn{3}{r}{Available aggregation methods} \\ \hline
Categorical	&	Selection & & \\
Multi-choice	&	Selection & & \\
Numerical &		Selection & Merge & \\
Vector	&		Selection & Merge & \\
Rankings &		Selection & Merge & \\
Text sequences &		Selection & Merge & Partition \\
Bounding boxes	&	Selection & Merge & Partition \\
Keypoints &	Selection & Merge & Partition \\
Parse trees	 &	Selection & & \\
Translations &		Selection & & \\
\end{tabular}
\caption{Types of labels and proposed aggregation methods available.}
\label{table:typesandmethods}
\end{table}

\section{Results on Synthetic Datasets}
\label{sec:results-sim}

Some research questions are difficult to answer using the limited real data available:
\begin{itemize}
    \item How do the many different possible aspects of a dataset affect performance of aggregation models? (RQs\ref{RQ:tasktype},\ref{RQ:sparsity},\ref{RQ:distribution})
    \item When is it useful to use a weighted aggregation model like MAS versus an unweighted approach like SAD? (RQ\ref{RQ:wgtvsnot})
    \item When is it most useful to use semi-supervised learning? (RQ\ref{RQ:semisup})
\end{itemize}

To answer these research questions, we need better controls to identify the variables that matter. To this end we develop simulators to produce synthetic data according to our best understanding of the many variations in the way annotation data might be generated, as explained in Sections \ref{data:binary}, \ref{data:rankings}, \ref{data:keypoints}, and \ref{data:parses}. We hypothesize that the most important variations for answering our research questions relate to the number of items, annotators, and annotators per item, as well as the distribution of worker reliability. Furthermore, simulators give us the ability to check the models' inference of controllable variables like worker reliability that cannot be directly observed in real data, only estimated. Together with the unit testing described in Section \ref{sec:unittests}, simulation studies allow us to verify the correct behavior of new and complicated models by seeing how well they recover the true values of their assumed parameters
 \cite{talts2018validating}.

To test our hypotheses, we run experiments with the following hierarchy of variation in configurations:

\begin{figure}
\centering
\includegraphics[scale=0.6]{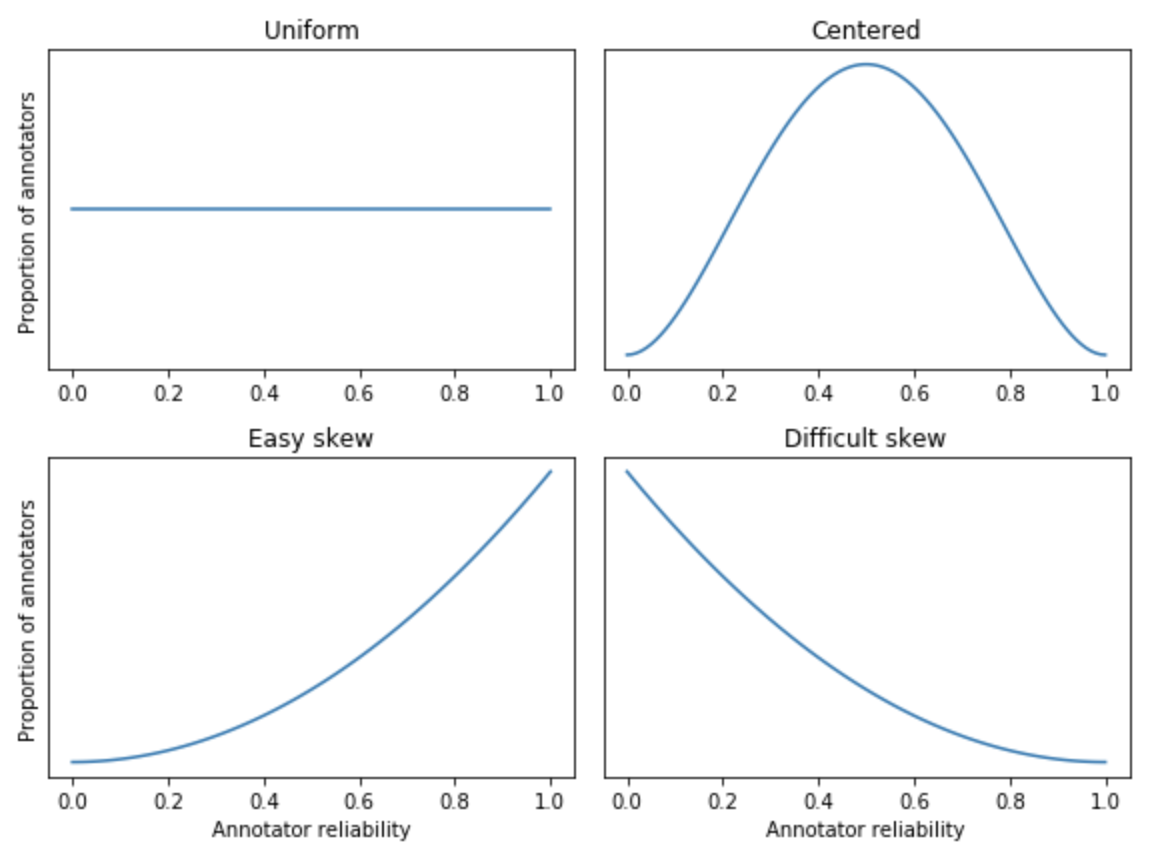}
\caption{Different simulator settings for distributions of annotator reliability. Uniform: Beta(1,1); Centered: Beta(3,3); Easy skew: Beta(1,3); Difficult skew: Beta(3,1).}
\label{fig:beta-dists-reliability}
\end{figure}

\begin{itemize}
    \item Data type: {Ranked list, keypoints, categorical}
    \item Data configuration
        \begin{itemize}
            \item \#Items $N$: {300, 400}
            \item \#Workers $J$: {10, 14, 21, 28}
            \item \%Workers/Item $r$: {0.3, 0.4, 0.5}
            \item Worker reliability distribution: {Uniform, concentrated, hard-skewed, easy-skewed}
        \end{itemize}
\end{itemize}

In total, 288 datasets are generated, and for each we record the following:
\begin{itemize}
\item Average SAD score (evaluation of SAD aggregate label vs.\ gold, averaged over items)
\item Average MAS score (evaluation of MAS aggregate label vs.\ gold, averaged over items)
\item Average DS score, only for the Binary classification experiments (evaluation of Dawid-Skene aggregate label against gold, averaged over items)
\item Average number of workers per item
\item Correlation between MAS-estimated worker error $\gamma_u$ and ``true'' simulated worker error $\sigma_u$
\item Krippendorff \citeyear{krippendorff2004reliability}'s $\alpha$ measure of inter-annotator reliability, using the distance function defined for the task in Section \ref{evaluation}.
\end{itemize}

The results of our simulator experiments help us answer various questions about the behavior of aggregation models on both simple and complex annotations.

\begin{figure}
\centering
  \fbox{\includegraphics[scale=.175]{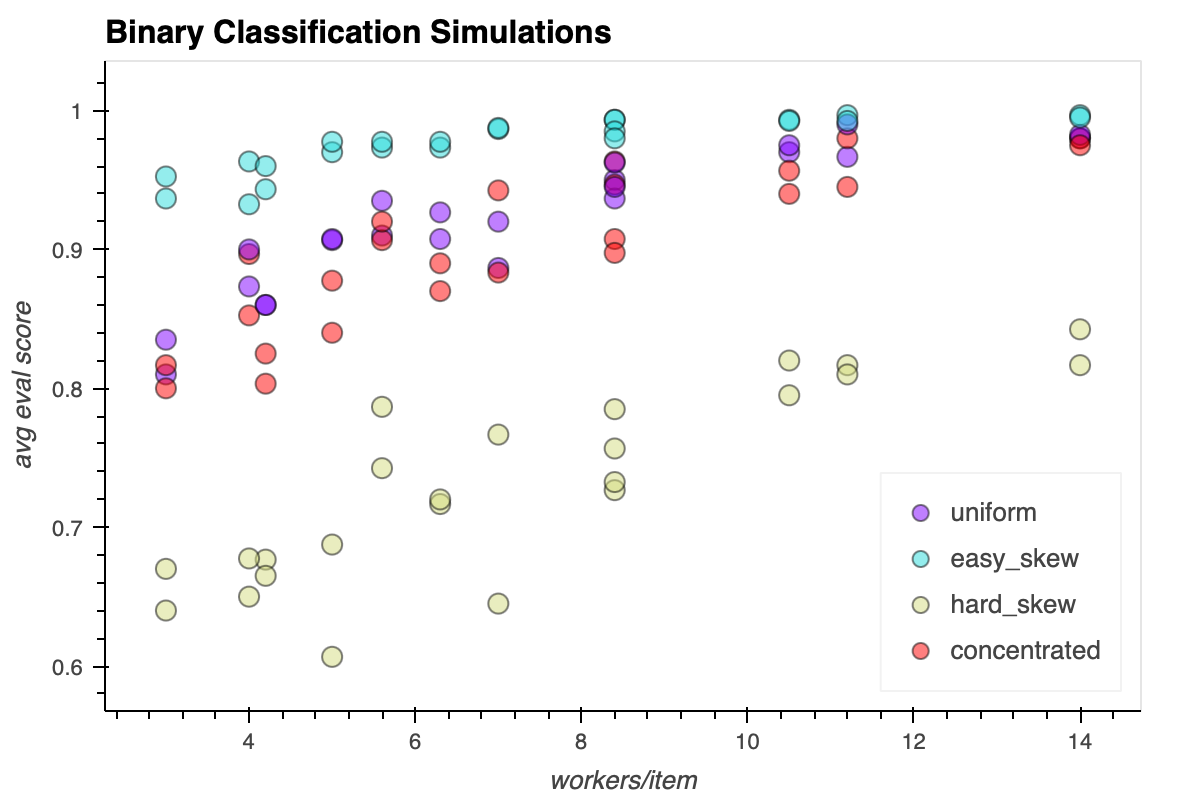}
  \includegraphics[scale=.175]{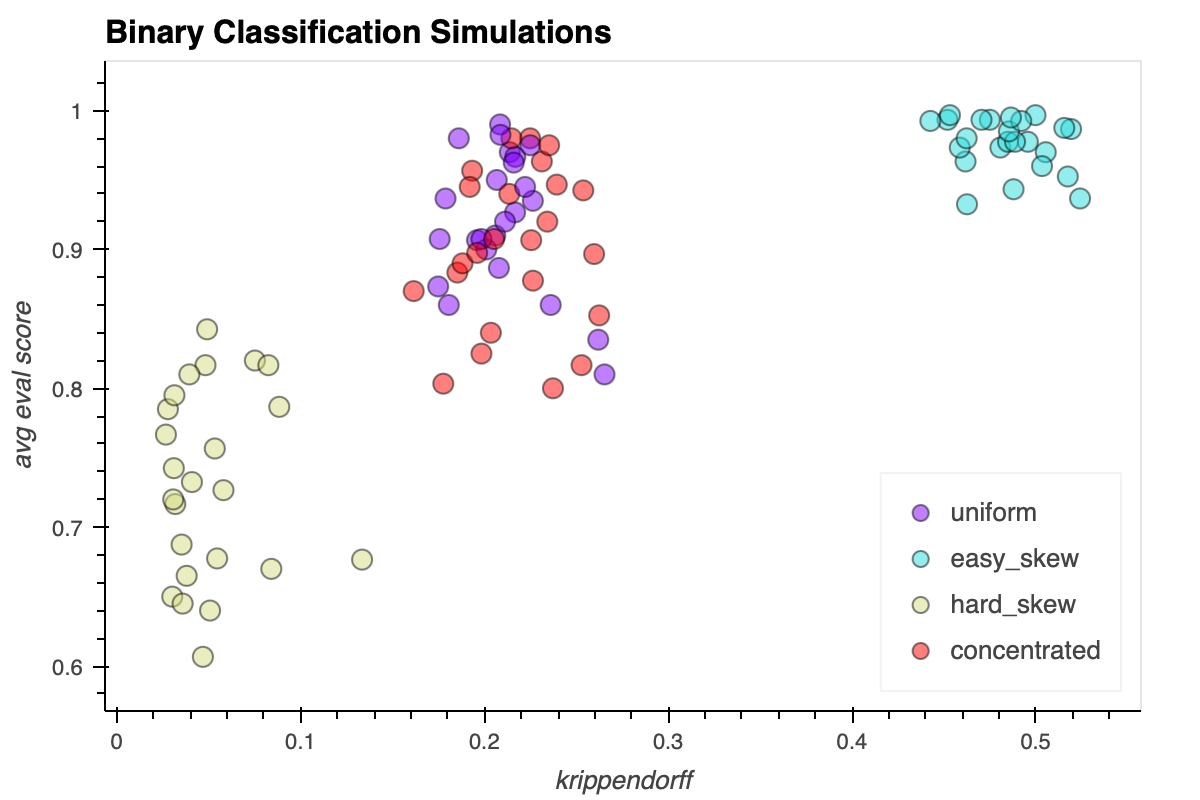}}
  
  \fbox{\includegraphics[scale=.175]{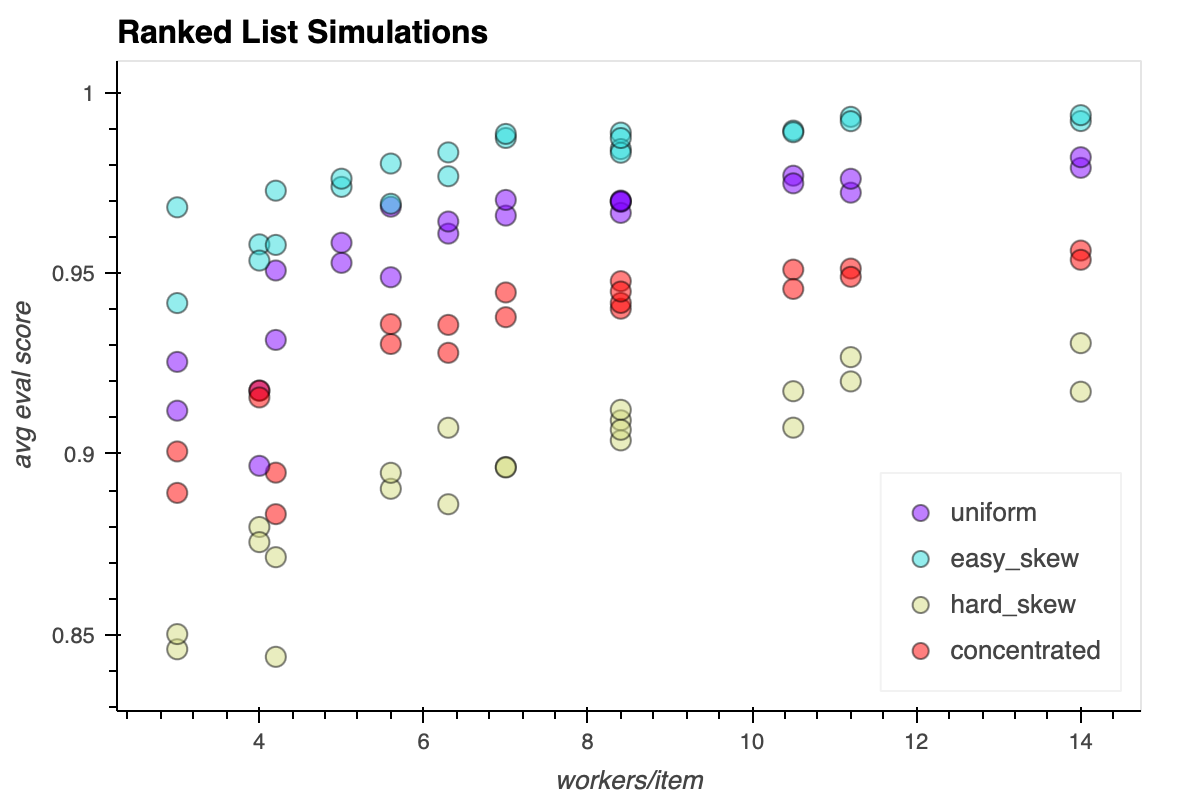}
  \includegraphics[scale=.175]{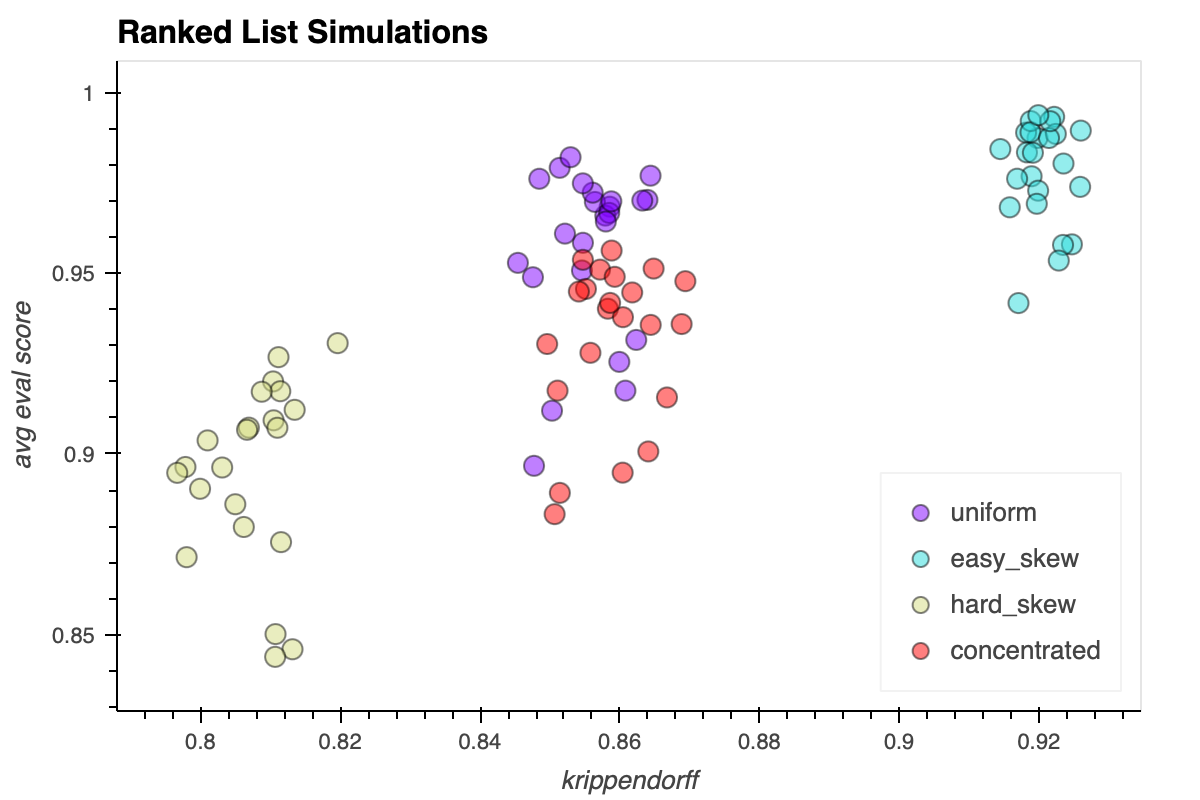}}
  
  \fbox{\includegraphics[scale=.175]{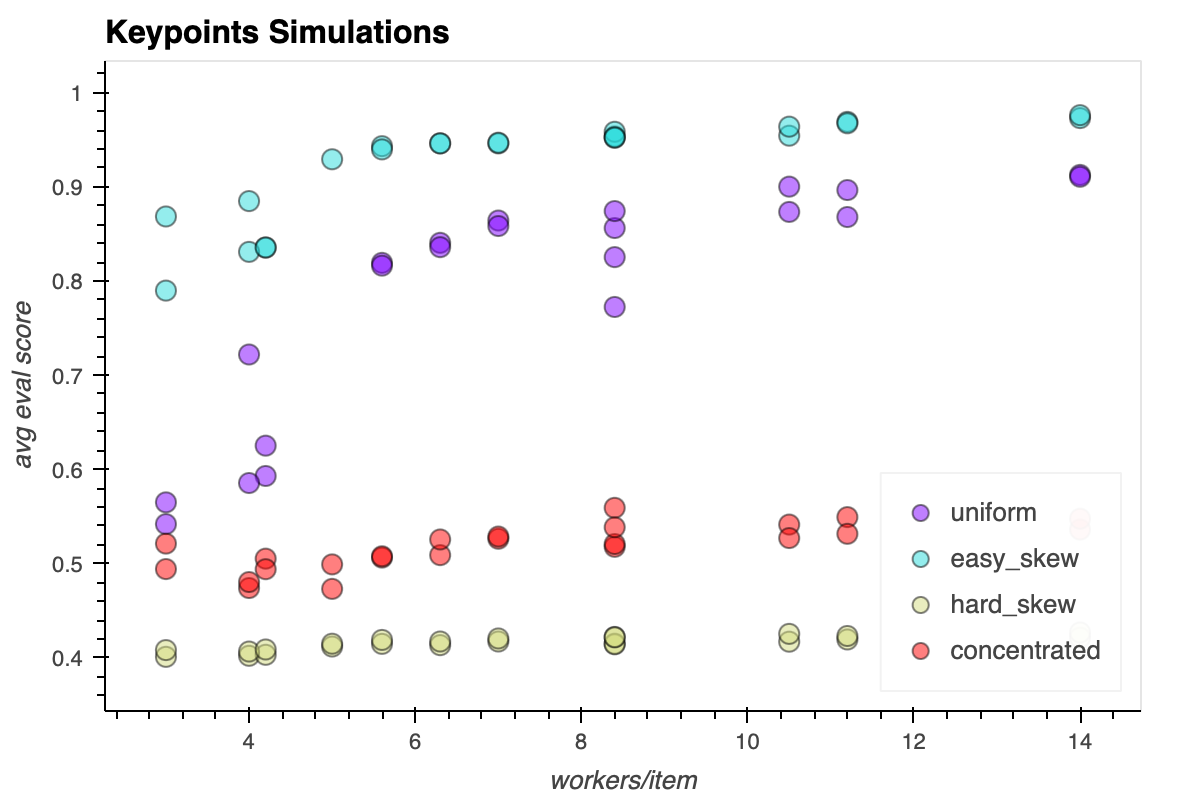}
  \includegraphics[scale=.175]{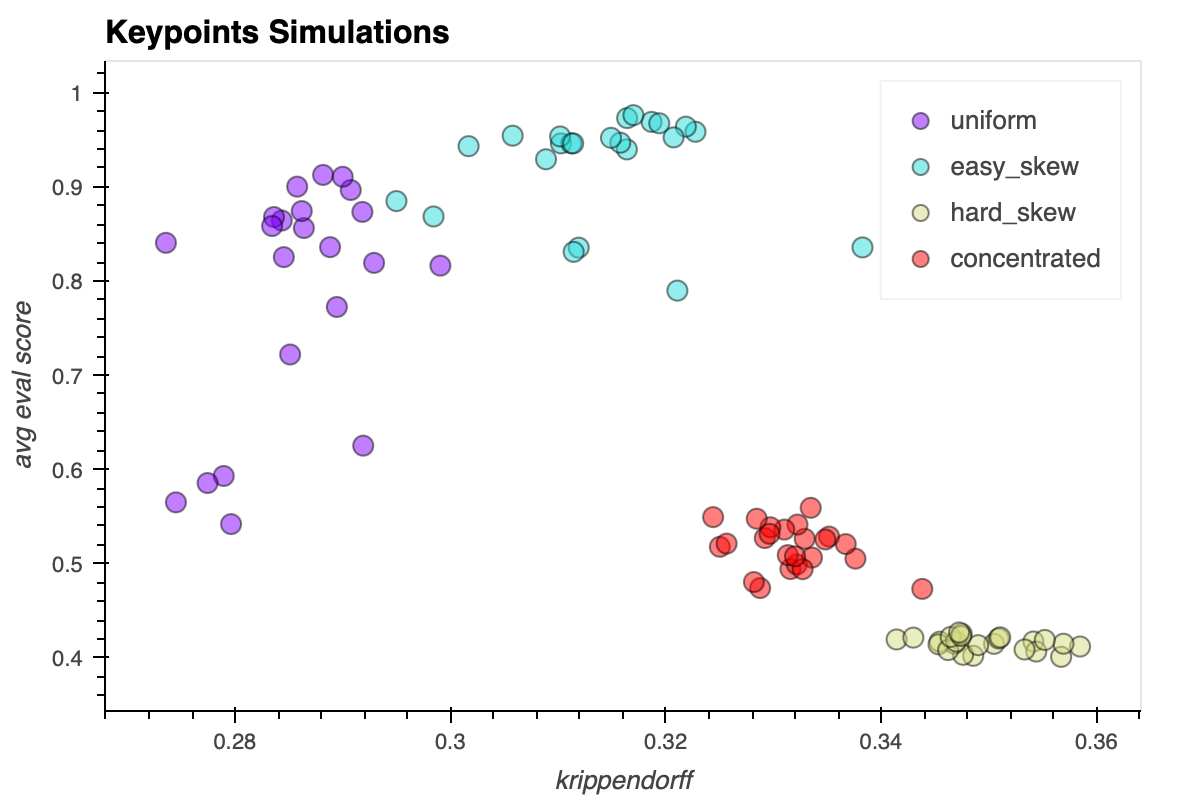}}
\caption{Simulation results explained by using worker count per item and worker reliability ($\gamma$) distribution. Left: effect on overall aggregation score. Right: effect on Krippendorff's $\alpha$ measure of inter-annotator agreement.}
\label{fig:sim-results-1}
\end{figure}

\begin{figure}
\centering
  \fbox{\includegraphics[scale=.175]{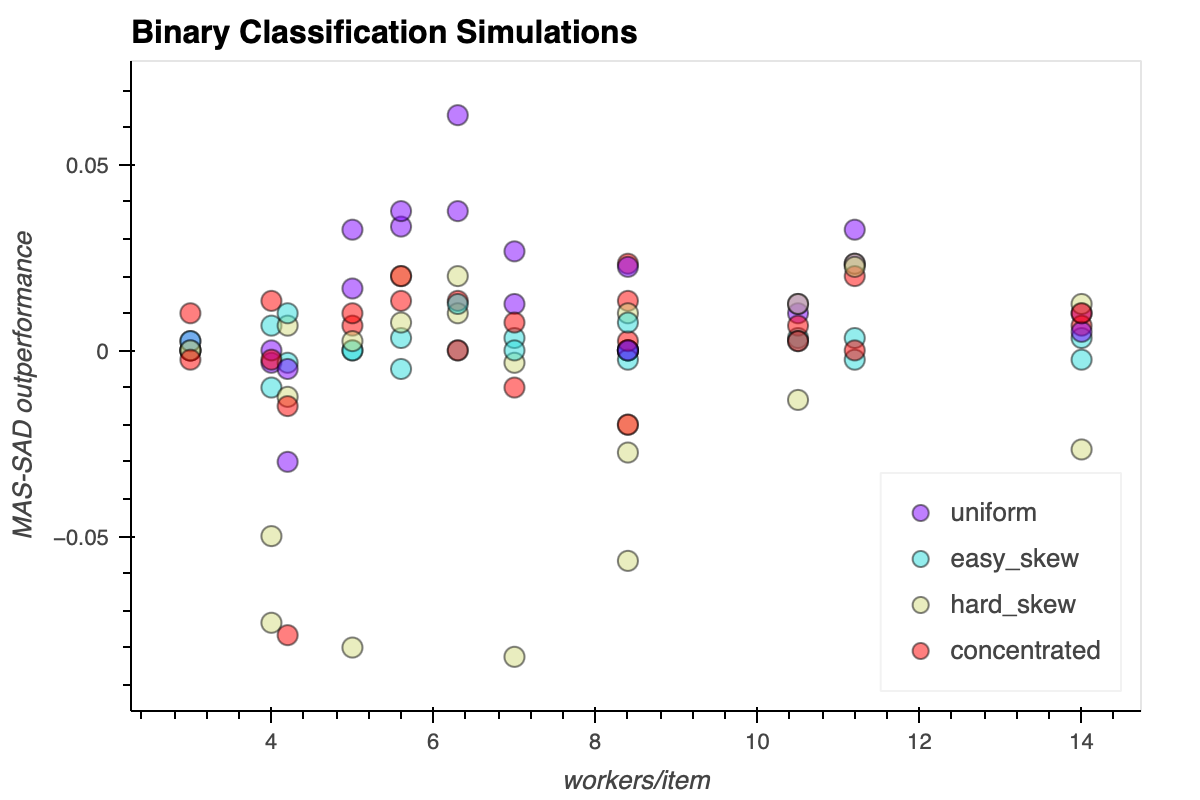}
  \includegraphics[scale=.175]{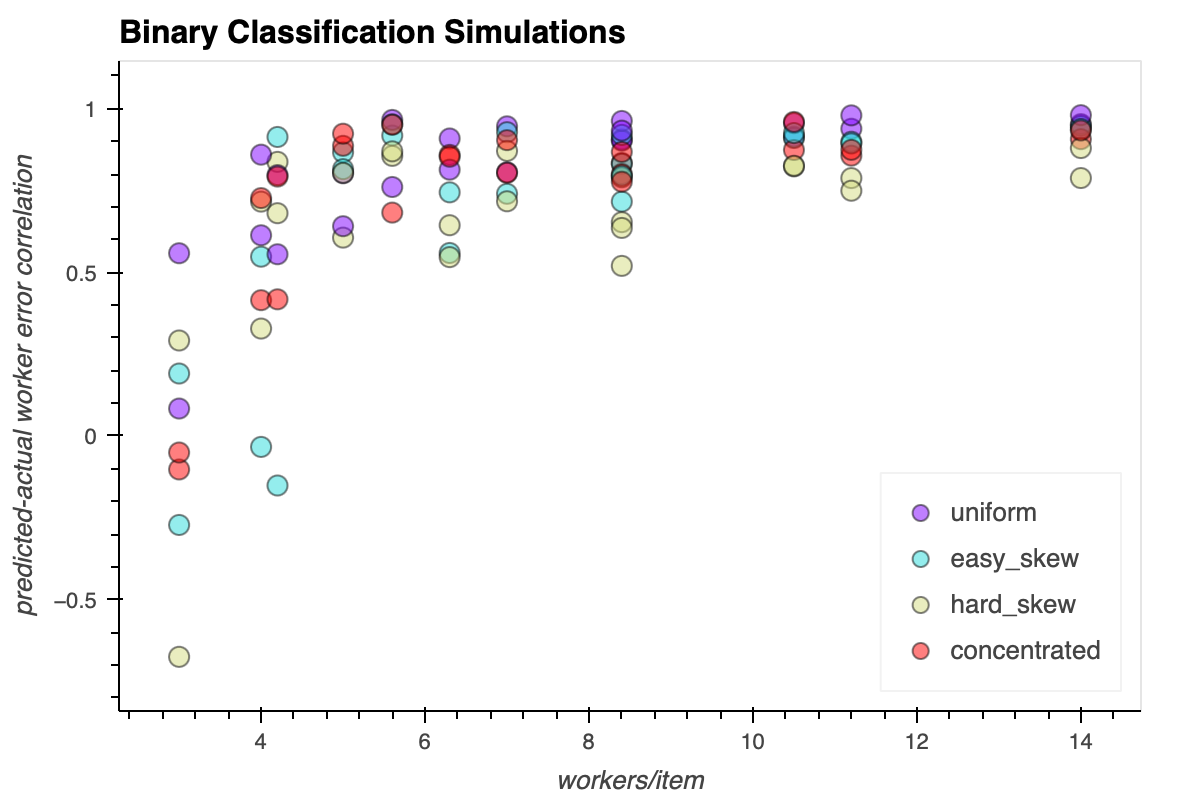}}
  
  \fbox{\includegraphics[scale=.175]{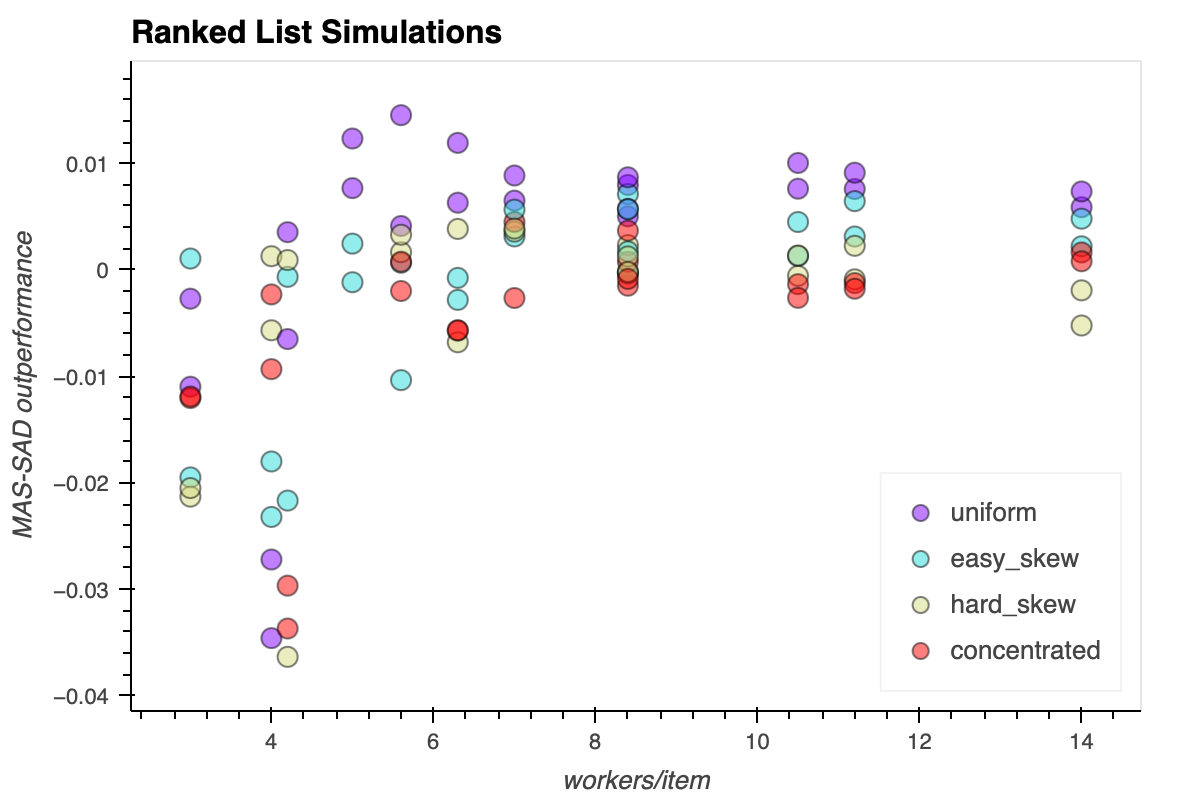}
  \includegraphics[scale=.175]{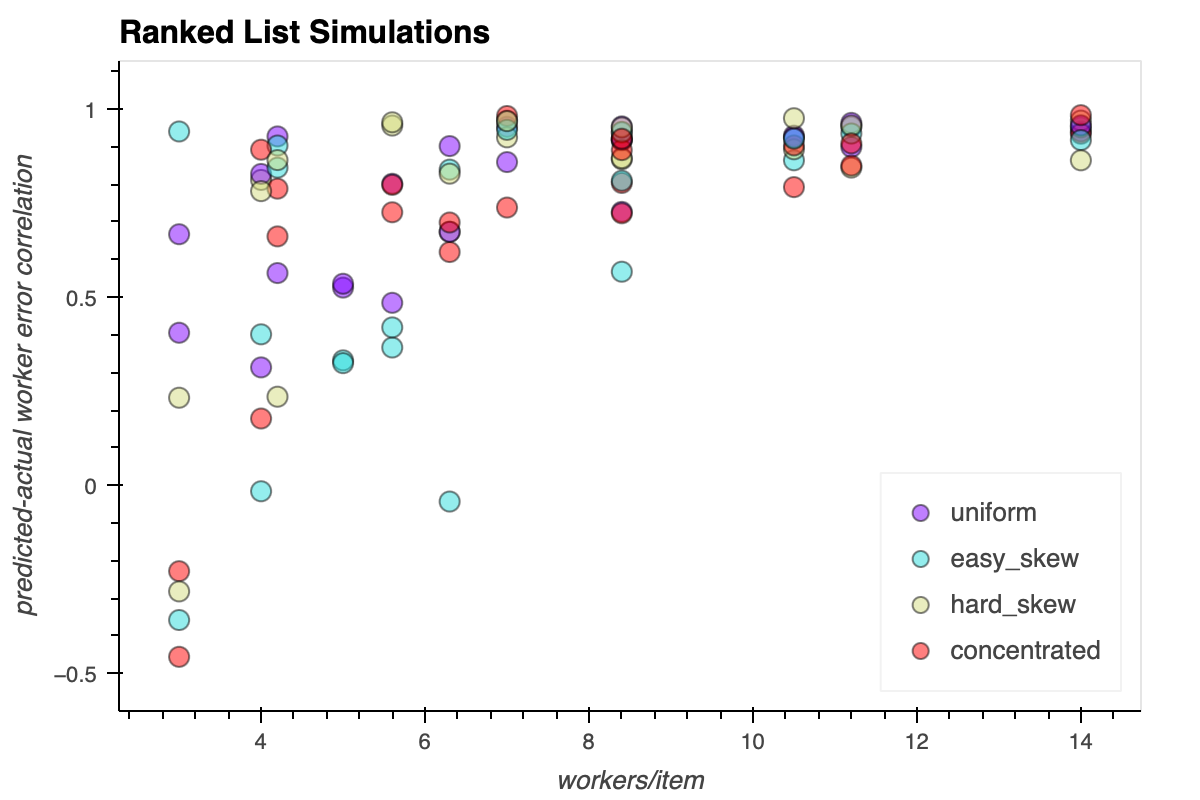}}
  
  \fbox{\includegraphics[scale=.175]{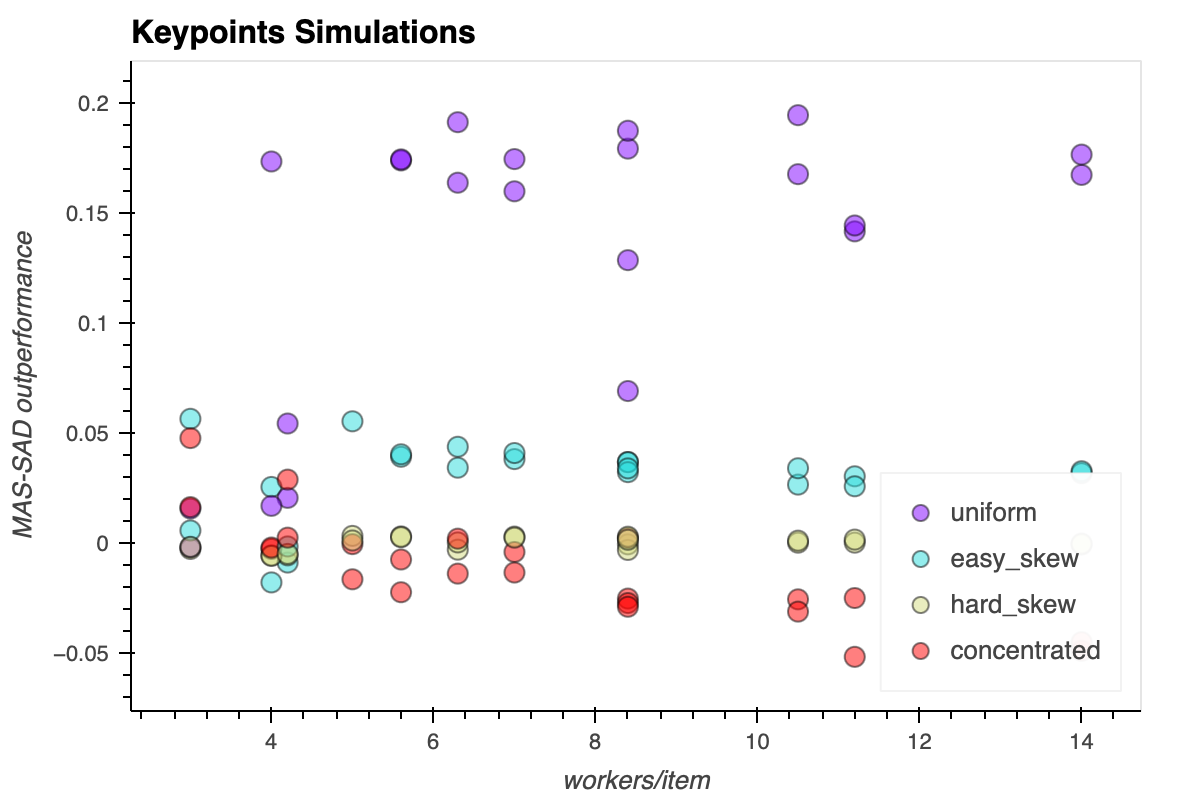}
  \includegraphics[scale=.175]{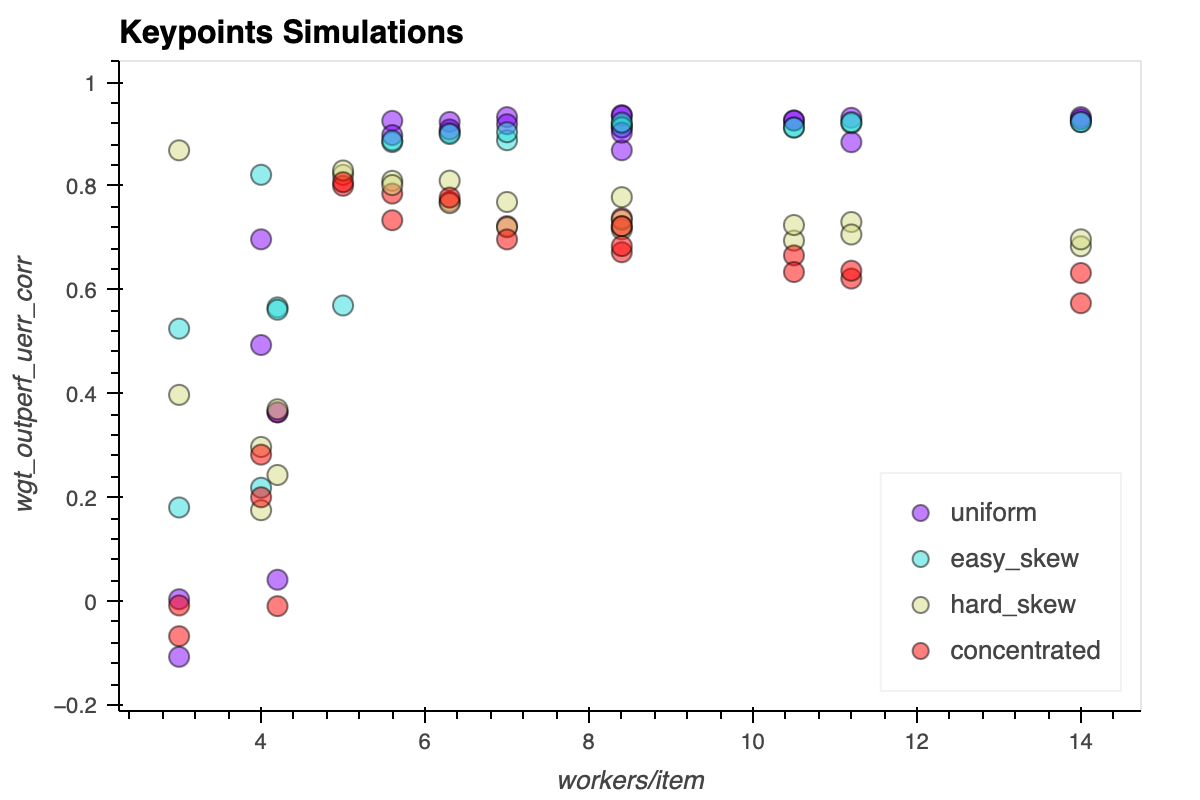}}
\caption{Simulation results explained by using worker count per item and worker reliability ($\gamma$) distribution. Left: effect on difference between weighted aggregation and unweighted aggregation performance. Right: effect on $(\gamma_u,\sigma_u)$ correlation (true worker reliability and model-estimated reliability parameters).}
\label{fig:sim-results-2}
\end{figure}

\begin{figure}
\centering
  \includegraphics[scale=.2]{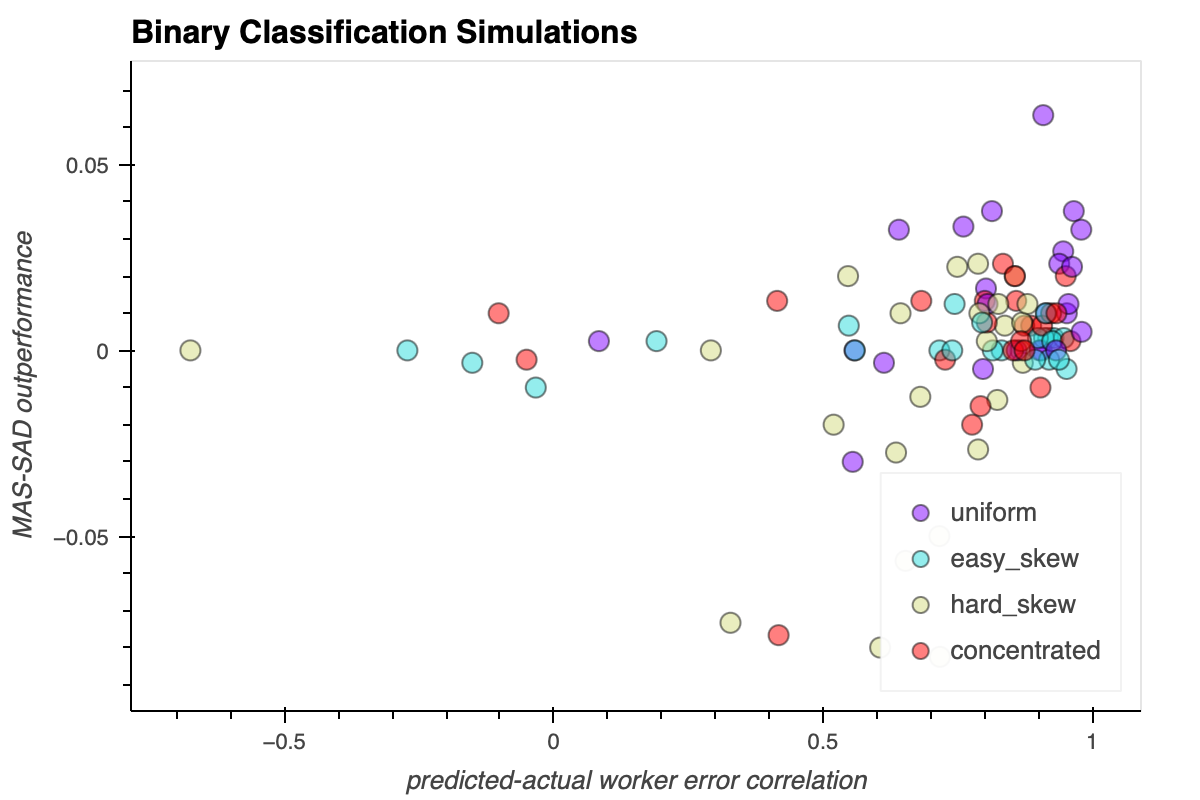}
  \includegraphics[scale=.2]{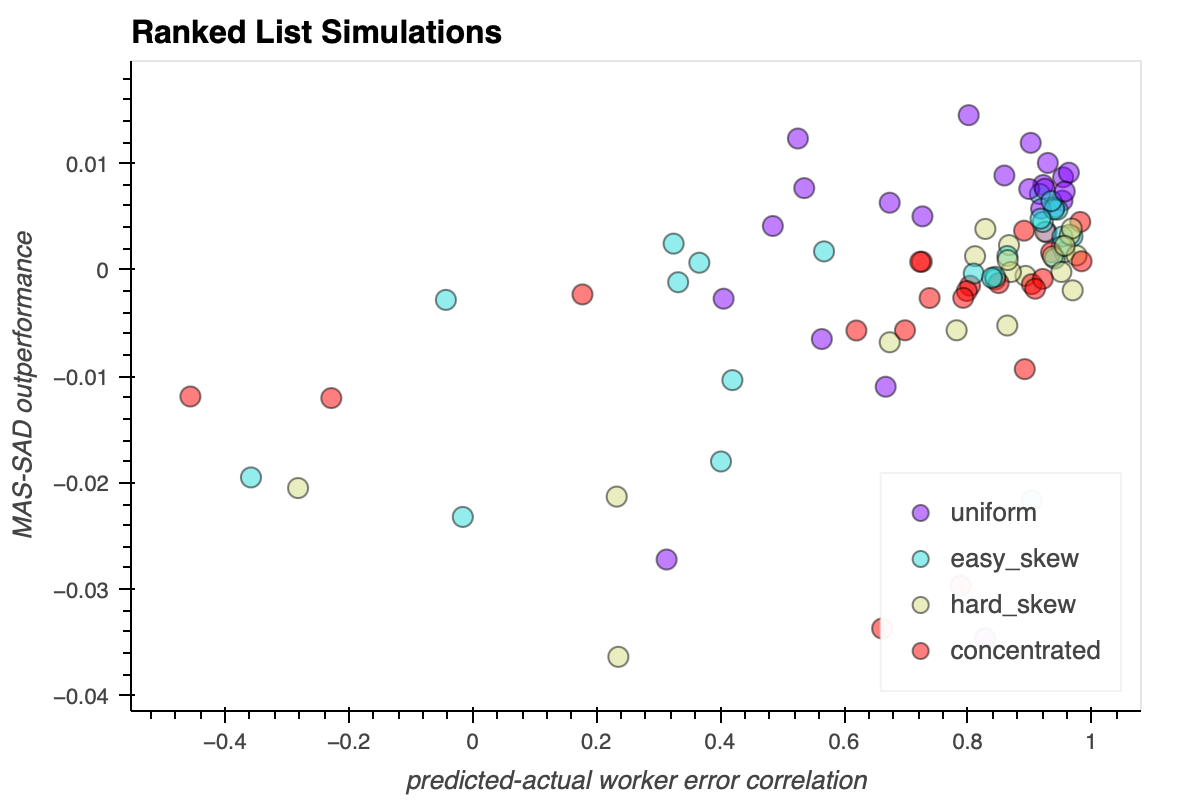}  \includegraphics[scale=.2]{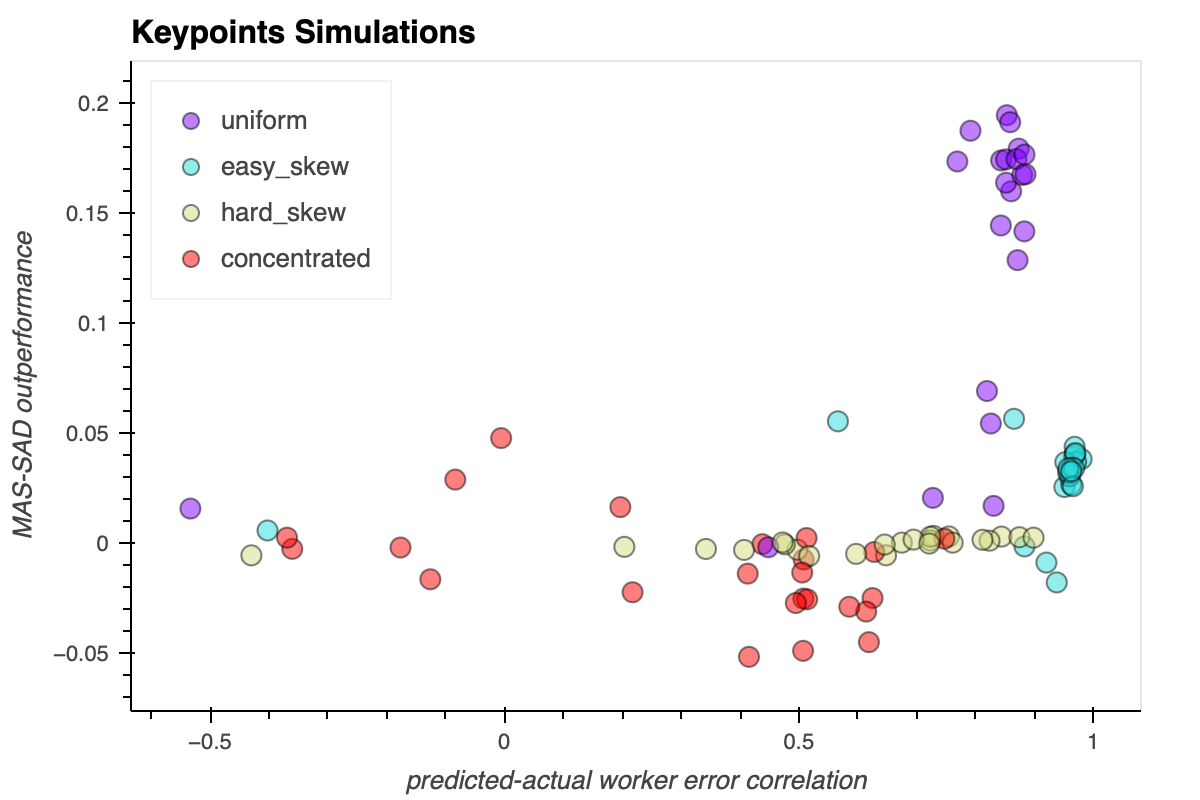}
  
\caption{Simulation results showing two measures of model quality against each other. In general the weighted model must accurately recover worker error estimates (indicated by correlation between $\gamma_u$ and $\sigma_u$ on x-axis) in order to outperform unweighted aggregation (indicated on y-axis).}
\label{fig:sim-results-3}
\end{figure}

\begin{figure}
\centering
\includegraphics[scale=0.8]{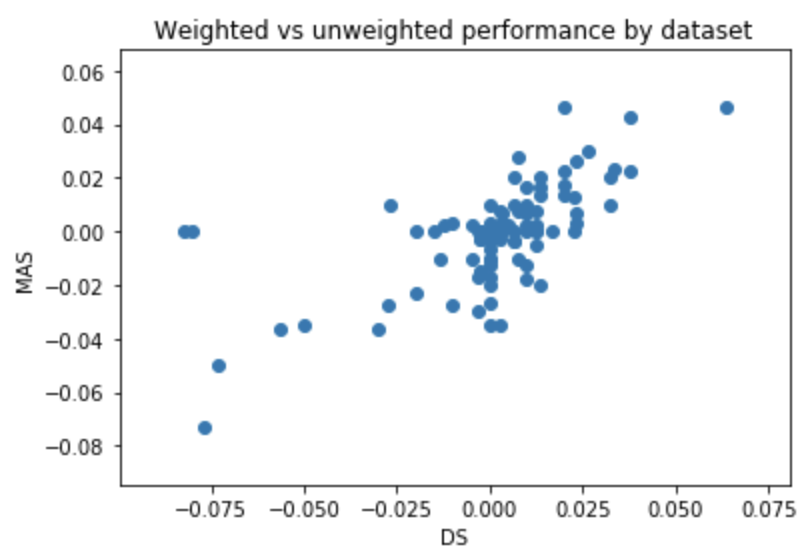}
\caption{Performance of Dawid-Skene (x-axis) and MAS (y-axis) relative to Majority Vote (SAD). Each point represents one simulated binary classification dataset. High correlation between DS and MAS shows that our weighted aggregation model for complex annotations exhibits similar behavior to the flagship weighted aggregation model for simple annotations.}
\label{fig:mas-vs-ds}
\end{figure}

\subsection{Weighted Model Performance Under Annotator Scarcity}
\label{sec:scarcity}
According to \citeA{ratner2017snorkel}, even under oracle annotator weightings, with enough scarcity in annotators per item, weighted aggregation is not expected to substantially outperform unweighted aggregation because disagreements are infrequent.
However, this may not be the case with complex annotations, where disagreements materialize on a continuous scale rather than all-or-nothing.
Meanwhile, we hypothesize that when annotator weightings are imprecise due to how noise and data scarcity affect model accuracy, weighted aggregation can actually underperform as it tends to favor random workers rather than the best workers (even for simple categorical datasets).
Once there are enough workers per item, however, weighted aggregation is then expected to begin performing better. 

Our simulation results confirm these hypotheses. 
The charts on the left side of Figure \ref{fig:sim-results-2} show how for each simulated task, MAS improvement over SAD varies according to the number of workers per item. On the low end of that range (fewer than five workers), there is little or no improvement, or even significantly worse performance in the case of Ranked Lists.

\subsection{Weighted Model Performance Under Annotator Abundance}
\label{sec:abundance}
According to \citeA{li2013error}, as the number of annotators per items grows large enough (and annotators tend to be reliable on average), majority vote should then converge towards the optimal solution. Therefore, we expect to see a weakening of improvement by the weighted model over majority vote as the annotators increase.
In our simulation results, we see hints of a slight downward trend in MAS-SAD improvement as the workers per item grows larger, but only by 14 workers per item does the improvement disappear in the Binary Classification and Ranked List simulations, while actually remaining positive for Keypoints. So even for numbers of workers per item that are much larger than the typical crowdsourced dataset, we see value in weighted aggregation over unweighted aggregation.

\subsection{Correlation Between Actual and Estimated Worker Reliability}

\label{sec:error-corr}
Using a model of annotation has benefits beyond simply producing a better aggregate label. One of those benefits is the ability to estimate annotator reliability or error. 
In experiments with low correlation, weighted models tend to underperform. The charts on the right side of Figure \ref{fig:sim-results-2} show that this correlation is very dependent on the number of workers per item. As expected, the MAS model is better at estimating worker-level parameters when the worker data is not too sparse.

\subsection{Distance-based Weighted Models Compared to Dawid-Skene on Categorical Data}
We find high correlation between MAS and DS in terms of performance over simulated datasets. MAS tends to outperform SAD in the same simulated datasets that DS outperforms SAD, and by similar amounts, as shown in Figure \ref{fig:mas-vs-ds}.
This result provides some assurance that our weighted aggregation model MAS is exhibiting similar behavior to the classical aggregation model, even though the mechanics are quite different, further supporting the positioning of MAS as the ``Dawid-Skene for complex annotations''.


\begin{table}[]
\begin{tabular}{ccl|cc|c|c}
\toprule
& workers &  worker $\sigma_u$  &    \multicolumn{2}{c}{Aggregation model score} &    $(\gamma_u,\sigma_u)$    &   Krippendorff's     \\
task & per item & distribution     &   Unweighted         &  Weighted         &     correlation      &   $\alpha$        \\
\midrule
\parbox[t]{2mm}{\multirow{8}{*}{\rotatebox[origin=c]{90}{Binary}}}
&
\parbox[t]{2mm}{\multirow{4}{*}{$< 5$}}

 & uniform      &     0.8629 &   0.8576 &    0.5780 &       0.2218 \\
& & concentrated &     0.8429 &   0.8314 &    0.3840 &       0.2315 \\
& & easy\_skew    &     0.9471 &   0.9481 &    0.2217 &       0.4930 \\
& & diff\_skew    &     0.6857 &   0.6643 &    0.3456 &       0.0642 \\
\cline{2-7} & \parbox[t]{2mm}{\multirow{4}{*}{$\geq 5$}}

  & uniform      &     0.9219 &   0.9429 &    0.8926 &       0.2072 \\
& & concentrated &     0.9167 &   0.9232 &    0.8662 &       0.2129 \\
& & easy\_skew    &     0.9841 &   0.9856 &    0.8510 &       0.4818 \\
& & diff\_skew    &     0.7621 &   0.7537 &    0.7446 &       0.0447 \\
\midrule
\parbox[t]{2mm}{\multirow{8}{*}{\rotatebox[origin=c]{90}{Ranked}}}
&
\parbox[t]{2mm}{\multirow{4}{*}{$< 5$}}
 & uniform      &     0.9350 &   0.9216 &    0.6146 &       0.8560 \\
& & concentrated &     0.9164 &   0.8996 &    0.3148 &       0.8572 \\
& & easy\_skew    &     0.9723 &   0.9595 &    0.4727 &       0.9204 \\
& & diff\_skew    &     0.8747 &   0.8619 &    0.4436 &       0.8081 \\
\cline{2-7} & \parbox[t]{2mm}{\multirow{4}{*}{$\geq 5$}}

 & uniform      &     0.9600 &   0.9681 &    0.8234 &       0.8556 \\
& & concentrated &     0.9440 &   0.9431 &    0.8312 &       0.8593 \\
& & easy\_skew    &     0.9829 &   0.9848 &    0.7235 &       0.9202 \\
& & diff\_skew    &     0.9072 &   0.9078 &    0.9068 &       0.8069 \\

\midrule
\parbox[t]{2mm}{\multirow{8}{*}{\rotatebox[origin=c]{90}{Keypoints}}}
&
\parbox[t]{2mm}{\multirow{4}{*}{$< 5$}}
 & uniform      &     0.5595 &   0.6023 &    0.5382 &       0.2815 \\
& & concentrated &     0.4804 &   0.4953 &   -0.1405 &       0.3305 \\
& & easy\_skew    &     0.8301 &   0.8376 &    0.6610 &       0.3142 \\
& & diff\_skew    &     0.4101 &   0.4054 &    0.3309 &       0.3511 \\
\cline{2-7} & \parbox[t]{2mm}{\multirow{4}{*}{$\geq 5$}}

 & uniform      &     0.6954 &   0.8572 &    0.8561 &       0.2878 \\
& & concentrated &     0.5467 &   0.5244 &    0.4801 &       0.3322 \\
& & easy\_skew    &     0.9178 &   0.9536 &    0.9378 &       0.3146 \\
& & diff\_skew    &     0.4184 &   0.4190 &    0.7037 &       0.3492 \\
\bottomrule
\end{tabular}
\caption{Simulation experiment results. The main objective is to compare weighted aggregation modeling (MAS) against unweighted (SAD), under various simulation conditions. On the left we describe the simulation settings: the task (Sections \ref{data:binary}-\ref{data:keypoints}), the workers per item (Sections \ref{sec:scarcity}-\ref{sec:abundance}), and the worker $\sigma_u$ distribution (Sections \ref{sec:error-skew}-\ref{sec:error-scale}). The next two columns contain the average evaluation score of unweighted models and weighted models for the given simulation. Next we show the correlation between the true worker error scale $\gamma_u$ and the estimated model parameter for worker error $\sigma_u$. Finally we note Krippendorff's $\alpha$ for the simulated data. These results show that particularly when there are at least five annotators and the worker ability is widely distributed or more concentrated in higher-ability workers, weighted models tend to estimate worker ability better and thus out-perform unweighted models.}
\label{Tab:resultssim}
\end{table}


\subsection{Model Performance by Skew of Annotator Error Distribution}
\label{sec:error-skew}
We hypothesize that the skew of worker error (see Figure \ref{fig:beta-dists-reliability}) should affect the performance of weighted aggregation models. That is, weighted aggregation models should perform well on an \emph{easy skew} dataset in which the proportion of reliable workers is higher than the proportion of unreliable workers. This is because we might expect a weighted aggregation model (that relies on consensus to estimate worker parameters) to be ``confused'' on a \emph{difficult skew} dataset in which the proportion of unreliable workers is much higher.
On the other hand, we might also speculate that with a larger response space typical of complex datasets, unreliable workers will not form clusters of responses that the model will take as consensus opinion, but rather be more scattered apart randomly in the response space.

The results of the simulation experiments show that worker error is more accurately estimated by MAS under the easy skew setting than under the difficult skew setting. Consequently MAS-SAD improvement is also generally better under the easy skew setting. This result shows that confusion from high proportion of inaccurate workers is harmful to annotator weight inference even in the larger response spaces of complex annotation tasks.

\subsection{Model Performance by Scale of Annotator Error Distribution}
\label{sec:error-scale}
The scale of worker error also matters. The uniform and concentrated distributions (see Figure \ref{fig:beta-dists-reliability}) both have the same skew, but different scales. The uniform scale has a higher proportion of high-reliability and low-reliability workers, whereas the concentrated scale has fewer workers in the extremes and a much higher proportion of similarly-skilled workers. 

The higher variation in worker skill produced by the uniform setting makes weighted voting more consequential, with the potential for both larger gains as well as greater losses, depending on how well the model estimates worker skills. The results here clearly show MAS outperforming SAD under the uniform setting compared to the concentrated setting. Correlation between actual and predicted worker error is also higher under the uniform setting than under the concentrated setting, and this gap is especially stark when considering fewer than five workers per item.

\subsection{Unexplained Differences between Simulated Tasks}
Although we noted several common patterns across the experiments, there are several observations about the differences between binary classification, ranked list, and keypoints simulations that remain unexplained.

For binary classification and ranked lists, average MAS score improves with workers/items, regardless of the worker error distribution. However, with keypoints the two ``worse'' distributions of concentrated and hard skew barely show any improvement at all when aggregating over larger numbers of workers (Figure \ref{fig:sim-results-1}).

Another anomaly shown by the keypoints experiments is that the overall Krippendorff's $\alpha$ measure of annotator agreement is actually higher for the concentrated and hard skew worker error distributions, which is counter-intuitive and different from the binary classification and ranked list experiments which show a reasonable order.

On the other hand, the keypoints experiments show the largest and most consistent improvement of weighted MAS aggregation over unweighted SAD. Even for concentrated and hard skew distributions, the underperformance of MAS typically seen in these distributions is minimal, although it strangely deteriorates as the number of workers increases (Figure \ref{fig:sim-results-2}).

Another surprise is how badly weighted aggregation underperforms in ranked lists when the number of workers is sparse. Even for the ``good'' uniform and easy skew distributions of worker error, the underperformance is quite bad when the average number of workers per item is about four. Although this underperformance under sparse worker conditions is true for the binary classification experiments, it is not quite so dramatic.

These unexplained differences show that there are factors that may be unique to the specific task or perhaps the distance functions for that task, affecting the performance of aggregation models. While we can identify some common factors affecting performance across tasks, these task-specific factors are much more difficult to characterize. One potential unexplored factor of influence may be the choice of distance function. Future work should investigate other factors that might further explain variation in performance.

\subsection{Semi-supervised Learning}

The number of workers per item is known and controlled by the data collector, who can therefore use that information to either collect more data or choose between weighted and unweighted aggregation.
On the other hand, the distribution of true worker error is not actually known in real life the way it is in these simulated experiments. Since a distribution with a difficult skew can lead to the worst overall performance (see charts on left of Figure \ref{fig:sim-results-1}), this is a serious concern for annotation tasks suspected to be particularly challenging for workers.

\begin{figure}[] 

\begin{subfigure}{0.44\textwidth}
\includegraphics[width=\linewidth]{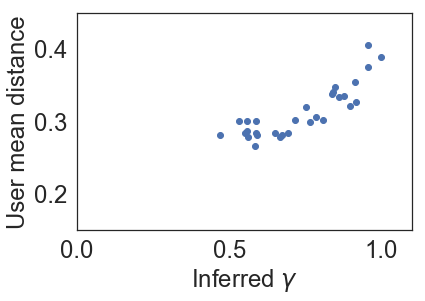}
\caption{\tabular[t]{@{}l@{}}Parses (easy skew) \\ unsupervised\endtabular} \label{fig:peasyus}
\end{subfigure}\hspace*{\fill}
\begin{subfigure}{0.44\textwidth}
\includegraphics[width=\linewidth]{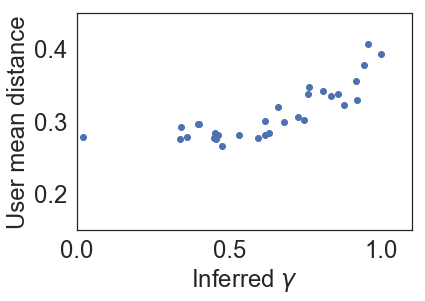}
\caption{\tabular[t]{@{}l@{}}Parses (easy skew) \\ semi-supervised\endtabular} \label{fig:peasyss}
\end{subfigure}
\medskip

\begin{subfigure}{0.44\textwidth}
\includegraphics[width=\linewidth]{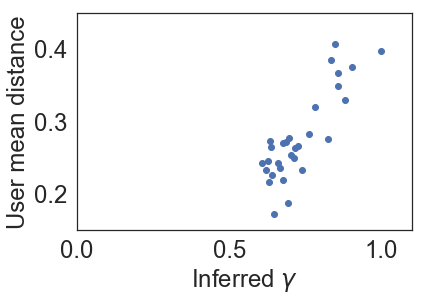}
\caption{\tabular[t]{@{}l@{}}Parses (difficult skew) \\ unsupervised\endtabular} \label{fig:phardus}
\end{subfigure}\hspace*{\fill}
\begin{subfigure}{0.44\textwidth}
\includegraphics[width=\linewidth]{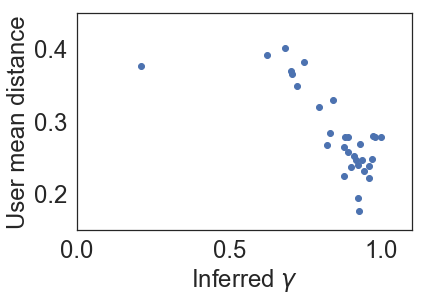}
\caption{\tabular[t]{@{}l@{}}Parses (difficult skew) \\ semi-supervised\endtabular} \label{fig:phardss}
\end{subfigure}

\caption{These scatterplots show the MAS inferred $\gamma$ against the BAU scores per annotator (``User mean distance''). In the semi-supervised cases, inference of the $\gamma$ parameters by MAS is improved by learning directly from a subset of items that have known gold annotations. In the difficult skew simulator configuration, this reordering against consensus is what allows semi-supervised MAS (SMAS) to outperform.}\label{fig:parsergamma}
\end{figure}

Experiments on simulated Syntactic Parsing annotations in 
\citeA{braylan2020modeling} show how semi-supervised MAS can overcome this challenge.
SMAS uses semi-supervised learning to succeed in the difficult skew simulation by learning more accurate worker error parameters as depicted in {\bf Figure \ref{fig:parsergamma}}. Unsupervised MAS learns annotator error $\gamma$ parameters that correlate closely with BAU's calculated annotator average distances, as seen in Figure \ref{fig:phardus}. In this unsupervised case both MAS and BAU suffer because their most trusted annotators are actually the most erroneous. When using semi-supervised learning, however, SMAS is able to rearrange its inferences for annotator reliability in the reverse and correct direction, as seen in Figure \ref{fig:phardss}.
This way SMAS is able to outperform other methods even when the distribution of worker error is of difficult skew.
Syntactic Parsing annotations used in this experiment were excluded from the larger simulation study because the parsers are extremely slow for generating large amounts of data.

\section{Results on Real Datasets}
\label{sec:results-real}

The experiments on real datasets help answer some of our overarching research questions:
\begin{itemize}
    \item How much better is it to aggregate labels by modeling distances between them, compared to just taking a single label?
    \item Are distance-based aggregation models competitive against task-specific baselines? (RQ\ref{RQ:bespoke})
    \item What benefit is there to using a weighted model compared to an unweighted method? (RQ\ref{RQ:wgtvsnot})
    \item How much is performance improved by using our semi-supervised learning method? (RQ\ref{RQ:semisup})
    \item Is it better to select or merge labels, when applicable?
    \item Is it better to partition multi-object annotations or select the best label set from a single annotator?
    \item What is the benefit of using the more complicated MAS over MADD? (RQ\ref{RQ:masvsmadd})
\end{itemize}

 For six real datasets, the results of evaluating each aggregation method's aggregate label were averaged across all items in the dataset, shown in Table \ref{Tab:resultscomplex}.
 Values are bolded if the difference in performance between it and the highest score (excluding oracles) is not statistically significant. In other words, among usable methods, non-bolded results are statistically significantly worse than the best-performing method. We used a t-test to assess statistical significance for the ordinal, numerical, and complex annotation tasks.
 
These results help us answer our research questions:

\begin{table*}[]
\begin{center}
\resizebox{\linewidth}{!}{
\begin{tabular}{lrrrrrrr}
\\
\toprule

{\bf Dataset} & {\bf Annotators} & {\bf Items} & {\bf Annotations} & {\bf Annotators/Item} & {\bf Objects/Item} & {\bf Section}\\
\hline

Movie             & 143   & 500   & 10000  & 20 & - & \ref{data:simple}     \\
Temporal          & 76    & 462   & 4620   & 10 & - & \ref{data:simple}    \\
RTE               & 164   & 800   & 8000   & 10 & - & \ref{data:simple}     \\
Dog               & 109   & 807   & 8070   & 10 & - & \ref{data:simple}     \\
Face              & 27    & 584   & 5242   & 9 & - & \ref{data:simple}   \\
FaceMC           & 27    & 584   & 5242   & 9 & - & \ref{data:simple}      \\
Disambiguation    & 34    & 177   & 1770   & 10 & - & \ref{data:simple}      \\
Weather           & 745   & 664   & 3180   & 4.8 & - & \ref{data:simple}      \\
Adult            & 268   & 318   & 3309   & 10.4 & - & \ref{data:simple}    \\
Emotion                  & 38    & 700   & 7000   & 10 & - & \ref{data:simple}     \\
Similarity                & 10    & 30    & 300    & 10 & - & \ref{data:simple}     \\
Population               & 603   & 296   & 1011   & 3.4 & - & \ref{data:simple}      \\
\midrule
Translation (J1)  & 70 & 250 & 2490 &  9.96$\pm$0.4 &  - & \ref{data:translations}\\
Translation (T1)  & 42 & 100 & 1000 &  10 &  - & \ref{data:translations}\\
Translation (T2)  & 42 & 100 & 1000 &  10 &  - & \ref{data:translations}\\
Sequence (PICO)  & 91 & 191 & 1165 &  6.1$\pm$1.2 & 3.87 & \ref{data:sequences}\\
Sequence (NER)  & 47 & 199 & 981 &  4.93$\pm$2.24 &  25.79 & \ref{data:ner}\\
Bounding box (BB)  & 196 & 200 & 1723 &  8.62$\pm$1.88 &  6.54 & \ref{data:boundingboxes}\\
Affect vector  & 38 & 100 & 1000 &  10 &  - & \ref{data:affect}\\
\bottomrule
\\
\end{tabular}
}
\caption{Real datasets used and summary statistics, for simple annotation tasks and complex annotation tasks both real and synthetic. The number of annotators, number of items, and number of annotations vary by dataset. The number of annotations per item suggests how much information the model can learn from the data.}
\label{Tab:data}

\end{center}
\end{table*}

\begin{table*}
\centering
\resizebox{\linewidth}{!}{
\begin{threeparttable}
\begin{tabular}{r|cc|cc|c|cc|cc|c|c|c|c|c|c|r}
        \toprule
        \multirow{3}{*}{Method} & %
            \multicolumn{7}{c|}{\textbf{Categorical (binary)}} & %
            \multicolumn{4}{c|}{\textbf{Categorical (multi-class)}} & %
            \multicolumn{2}{c|}{\textbf{Multi-choice}} & %
            \multicolumn{2}{c|}{\textbf{Ordinal}} & %
            \multicolumn{1}{c}{\textbf{Numerical}} \\%
        \cline{2-17} & \multicolumn{2}{c|}{Movie} & %
            \multicolumn{2}{c|}{Temporal} & %
            \multicolumn{1}{c|}{RTE} & %
            \multicolumn{2}{c|}{Dog} & %
            \multicolumn{2}{c|}{Face} & %
            \!\!FaceMC\!\! & %
            WSD & %
            \!\!Weather\!\! & %
            Adult & %
            \!\!Emotion\!\! & %
            \!\!Similarity\!\! & %
            Population \\%
        \cline{2-17} & $Acc$ & $F_1$ & $Acc$ & $F_1$ & $Acc$ & $Acc$ & $F_1$ & $Acc$ & $F_1$ & $Acc$ & $Acc$ & $MAE$ & $MAE$ & $MAE$ & $MAE$ & \multicolumn{1}{c}{$MAE$} \\%
        \midrule

     BAU &  \underline{\textbf{0.956}} &  \underline{\textbf{0.957}}
 & \textbf{0.924} & \textbf{0.932}
& 0.8962
 & 0.798 & 0.797 & 0.568 & 0.548 & 0.568 &  \underline{\textbf{0.994}}
 & 0.122 & \textbf{0.296} & 16.050 & \textbf{3.715} &  \underline{\textbf{2684.868}} \\
 SAD & \textbf{0.940} & \textbf{0.941}
 & \textbf{0.929} & \textbf{0.934}
 & 0.8775
 & \textbf{0.822} & \textbf{0.821} & \textbf{0.646} & \textbf{0.629} &  \underline{\textbf{0.646}} &  \underline{\textbf{0.994}}
 & \textbf{0.087} & \textbf{0.299} & 13.954 & \textbf{3.592} & \textbf{4610.044} \\
 MAS & \textbf{0.944} & \textbf{0.945} & \textbf{0.931} & \textbf{0.938}
   & \textbf{0.9225}
 & \textbf{0.828} & \textbf{0.826} & \textbf{0.618} & \textbf{0.596} & \textbf{0.618} &  \underline{\textbf{0.994}}
 & 0.130 & \textbf{0.289} & \textbf{13.253} & \textbf{3.838} & \textbf{5045.454} \\
 MADD & \textbf{0.944} & \textbf{0.945} & 0.894 & 0.899
   & 0.8888
 & \textbf{0.819} & \textbf{0.819} & \textbf{0.637} & \textbf{0.616} & \textbf{0.637} &  \underline{\textbf{0.994}}
 & \textbf{0.084} &  \underline{\textbf{0.274}} & \textbf{12.576} & \textbf{3.622} & \textbf{2841.927} \\
 DS & \textbf{0.944} & \textbf{0.945} &  \underline{\textbf{0.944}} &  \underline{\textbf{0.949}}
  & \underline{\textbf{0.9275}}
 &  \underline{\textbf{0.845}} &  \underline{\textbf{0.844}} &  \underline{\textbf{0.647}} &  \underline{\textbf{0.635}} & \textbf{0.632} &  \underline{\textbf{0.994}}
 & 0.117 &  \underline{\textbf{0.274}} & - & - & - \\
 MACE & \textbf{0.946} & \textbf{0.947} & \textbf{0.942} & \textbf{0.934}
   & \textbf{0.9263}
 & \textbf{0.830} & \textbf{0.829} & \textbf{0.644} & \textbf{0.625} & \textbf{0.627} &  \underline{\textbf{0.994}}
 & \textbf{0.092} & \textbf{0.292} & - & - & - \\
 MV & \textbf{0.944} & \textbf{0.945} & \textbf{0.929} & \textbf{0.935}
   & 0.8775
 & \textbf{0.822} & \textbf{0.821} & \textbf{0.634} & \textbf{0.613} & \textbf{0.635} &  \underline{\textbf{0.994}}
 & \textbf{0.086} & \textbf{0.305} & - & - & - \\
 median & - & - & - & - & - & - & - & - & - & - & - & \textbf{0.087}
 & \textbf{0.300} & 13.529 & \textbf{3.576} & \textbf{3336.726} \\
 random & 0.880 & 0.883 & 0.758 & 0.767 & 0.695 & 0.695 & 0.555 & 0.536 & 0.568 & \textbf{0.983}
& 0.6088
 & 0.164 & 0.418 & 19.377 & \textbf{3.375} & 20719.010 \\
 \midrule
 CATD & \textbf{0.946} & \textbf{0.947} & \textbf{0.931} & \textbf{0.937}
   & \textbf{0.9233}
 & \textbf{0.824} & \textbf{0.823} & \textbf{0.618} & \textbf{0.596} & \textbf{0.616} &  \underline{\textbf{0.994}}
 & \textbf{0.095} & \textbf{0.308} & 16.359 & \textbf{3.692} & 1023\tnote{*} \\
 GPM & \textbf{0.944} & \textbf{0.945} & \textbf{0.939} & \textbf{0.946}
   & \underline{\textbf{0.9288}}
 & 0.818 & 0.817 & \textbf{0.637} & \textbf{0.618} & \textbf{0.637} &  \underline{\textbf{0.994}}
 &  \underline{\textbf{0.072}} & \textbf{0.299} & \textbf{12.022} & \textbf{3.368} & -\tnote{**} \\
 BAU-M & - & - & - & - & - & - & - & - & - & -
  & -
 & 0.150 & \textbf{0.339} &  \underline{\textbf{11.934}} &  \underline{\textbf{3.363}} & 12544.370 \\
 SAD-M & - & - & - & - & - & - & - & - & - & -
  & -
 & 0.132 & \textbf{0.319} & \textbf{11.964} & \textbf{3.410} & 15330.807 \\
 MAS-M & - & - & - & - & - & - & - & - & - & -
  & -
 & 0.138 & \textbf{0.296} & \textbf{12.128} & \textbf{3.552} & 6290.098 \\
 MADD-M & - & - & - & - & - & - & - & - & - & -
  & -
 & 0.119 & \textbf{0.324} & \textbf{12.165} & \textbf{3.440} & 11936.113 \\
 uniform-M & - & - & - & - & - & - & - & - & - & -
  & -
 & 0.164 & 0.353 & \textbf{12.022} & \textbf{3.368} & 21618.924 \\
 mean & - & - & - & - & - & - & - & - & - & -
  & -
 & 0.164 & 0.353 & \textbf{12.022} & \textbf{3.368} & 21618.924 \\

    \end{tabular}
    \begin{tablenotes}
        \item[*] Value from experiments run by \citeA{li2014confidence}. We could not replicate these results and therefore could not perform significance testing.
        \item[**] Time complexity became unmanageable
    \end{tablenotes}
\end{threeparttable}
}

\caption{Simple Annotation Benchmarking. The best score in each column is underlined, and all other scores with no statistical significant difference from the best are bolded. Merge methods are indicated by ``-M'' suffix. Overall there is not much consistent difference between aggregation methods on these simple real datasets.}
\label{table:simple_results}

\end{table*}

\begin{table}[t]
\begin{center}
\begin{tabular}{l|lllllllllll}
\toprule
Model &     J1 & T1 & T2 & Affect & PICO & NER & BB \\
\midrule
RU     &  0.189 & 0.1784 & 0.1663 & 5.3727 & 0.5673
& 0.6088
& 0.5697 \\
\midrule
TMV & - & - & - & - & 0.6448 & - & -  \\
CHMM & - & - & - & - & \textbf{0.6885} & 0.6669 &  -  \\
BVHP & - & - & - & - & - & - & 0.6426  \\
\midrule
HRRASA & 0.2595 & 0.2422 & \textbf{0.2818} & - & - & - & -  \\
BAU* & 0.2157 & 0.2374 & 0.2362 & 6.4551 & 0.6718
& 0.7675
& 0.6437  \\
SAD* & 0.2626 & 0.2376 & 0.2283 & 6.5493 & 0.6523
& 0.7700
& 0.6603  \\
MADD* & 0.2611 & 0.2370 & 0.2349 & \textbf{6.9870} & 0.6626
& 0.7671
& 0.6587  \\
MAS* & 0.2460 & 0.2408 & 0.2411 & 6.4925 & \textbf{0.6857}
& 0.7782
& 0.6617  \\
\midrule
SSL-BL* & 0.2655 & 0.2423 & 0.2372 & 6.6627 & \textbf{0.6992}
& 0.7688
& 0.6673  \\
SMAS* & \textbf{0.2919} & \textbf{0.2509} & 0.2659 & \textbf{6.9159} & \textbf{0.6941}
& 0.7707
& 0.6606  \\
\midrule
PSR-MAS* & - & - & - & - & \textbf{0.6905}
& \textbf{0.7887}
& \textbf{0.6860}  \\
PDMRR-MAS* & - & - & - & - & 0.6725
& -
& 0.6751  \\
\midrule
PSR-ORC & - & - & - & - & 0.7348
& 0.7913
& 0.7085  \\
PDMRR-ORC & - & - & - & - & 0.6904
& -
& 0.7236  \\
Select-ORC & 0.4934 & 0.3674 & 0.3551 & 9.7898 & 0.8182
& 0.8210
& 0.7434  \\
\bottomrule
\end{tabular}
\caption{Complex Annotation Benchmarking. Bold represents results that are not statistically significantly different from first place (other than Oracles). Horizontal lines separate categories of methods: non-aggregation, bespoke, distance-based, semi-supervised, multi-object, and oracle. Asterisk* denotes our methods. Our semi-supervised model SMAS is the most consistent out-performer. Our partitioning-based approach for multi-object annotations works best when applicable. Overall, weighted aggregation models outperform unweighted, except in the J1 dataset.}
\label{Tab:resultscomplex}
\end{center}
\end{table}


\subsection{Aggregating Multiple Labels vs.\ Collecting Only One Label per Item}

Note that Random User (RU) represents the expected performance of collecting one label per item. Across all experiments we consistently see substantial benefit to aggregating several labels per item over RU, regardless of the aggregation method.
This effect is the largest of note, demonstrating that doing any kind of aggregation is worthwhile.


\subsection{Distance-based Aggregation vs.\ Task-specific Baselines}
\label{sec:vs-baselines}

Distance-based methods perform comparably to prior work for most simple annotation tasks and are more versatile. This finding suggests their potential practical value as a general-purpose solution for both simple and complex tasks. For complex tasks, compared to task-specific baselines (TMV, CHMM, BVHP), general-purpose distance-based models tie (with text sequences), outperform (with bounding boxes), or lack prior baselines.

Note that we consider HRRASA \cite{li2020crowd} in the same class as our own distance-based methods, since the only task-specific component is the distance function.

For the Translations experiments, HRRASA is the best overall unsupervised method. MAS gets the top spot in two out of three datasets when it is allowed semi-supervision (SMAS); HRRASA does not support semi-supervision. As HRRASA and MAS are both weighted-vote distance models, the main difference seems to stem from the distance functions: we use the inverse evaluation function (GLEU score) as the distance function for MAS, whereas HRRASA uses a distance function which contains much richer linguistic information from both text embeddings and word sequence representations (they use RoBERTa language model). In fact, \citeA{li2020crowd} also find in ablation studies that HRRASA benefits from these multiple sources of information in the distance function, suggesting potential benefit to our methods from exploring more sophisticated distance functions in future work.

\subsection{Weighted vs.\ Unweighted Methods}

In the simple labeling task experiments, the differences between weighted (BAU, MAS, MADD, DS, MACE) and unweighted (SAD, MV) vote models are not statistically significant, except for on the RTE task where the weighted methods significantly outperform (and which is thought to have greater variation in annotator ability \cite{paun2018comparing}). While specialized models are usually shown in the literature to outperform simple majority vote on the datasets they are developed for, prior benchmarking studies comparing aggregation methods on a wide variety of datasets also find that majority vote is often very competitive \cite{Sheshadri13,hung2013evaluation,zheng2017truth}. In contrast, however, with the complex labeling tasks our recommended weighted model MAS outperforms SAD in all but two examples. In the J1 dataset, SAD does significantly better, but SAD also outperforms HRRASA in that example, suggesting maybe it is a difficult task with expert workers in the minority (see ``difficult skew'' distribution in Section \ref{sec:error-skew}).

While weighted methods overall seem to perform as well or better than unweighted, the benefits of unweighted methods are in their simplicity and not needing worker identifiers.
%
%
Weighted voting methods seem most valuable when there is high variance in worker accuracy, with enough instances to estimate worker accuracy reasonably well. A reason that weighted voting methods such as MAS outperform in the complex annotation tasks could be due to this diversity in annotator ability. In practice, we do not know this in advance for a given dataset without already knowing ground truth, but adopting weighted methods like MAS is a safer choice (more likely to outperform) than unweighted methods like SAD.

\subsection{Semi-supervised Learning}


Prior work by \citeA{tang2011semi} show that increasing the number of gold labels past even a small fraction of the non-expert labels enables their semi-supervised system to significantly increase accuracy on items without known gold.
While our experiments do not vary the amount of supervision, we find that even with only 10\% supervision, SMAS is consistently at least as strong as MAS, sometimes quite significantly. {\bf SMAS also achieves stronger results than our previous semi-supervised method SSL-BL}.
The only dataset where SMAS is not significantly stronger than MAS is BB, which also shows similar scores across aggregation methods, suggesting the annotators on this task may exhibit the ``concentrated'' error distribution (not much variance in ability) described in Section \ref{sec:error-scale}.
Ultimately, results show that findings of prior semi-supervised work for simple labeling tasks \cite{wang2011managing,tang2011semi} seem to carry over to complex labeling tasks, with our task-agnostic distance-based model able to similarly exploit and benefit from semi-supervised training.
These results also support the idea that the most important concern in training accurate aggregation models is correctly estimating annotator reliability, suggesting that many of the cases where aggregation models fail (e.g. relative to majority vote) is when they incorrectly estimate annotator reliability.

\subsection{Is it Better to Select or Merge Labels, When Applicable?}

In the multi-object experiments of Text Sequences (PICO) and Bounding Boxes (BB), we consider a realistic and an oracle setting in evaluating merging (PDMRR) against selection (PSR). The realistic setting involves partitioning via (real and imperfect) clustering, whereas the oracle setting uses oracle partitioning as an upper-bound for evaluating selection vs.\ merge under ideal partitioning conditions. For the realistic setting, 
%
%
selection outperforms merging. 
With the oracle partitioning, however, the difference between selection and merging is less clear: selection outperforms for text sequences, while merging outperforms for bounding boxes, with no clear winner for keypoints. It seems wise for now to use selection after partitioning in practice, but future improvements to partitioning may ultimately yield greater benefit from merging over selection. 

\subsection{Partition Multi-object Annotations or Select All from a Single Annotator?}

For the four multi-object experiments (including simulated Keypoints from Section \ref{data:keypoints}), we have compared partitioning approaches PDMRR and PSR against general-purpose selection baselines as well as task-specific methods. Our partition-select method PSR is the consistent winner across all datasets, tying selection-MAS and task-specific CHMM in PICO while statistically significantly outperforming all baselines in NER, Bounding Boxes, and Keypoints.
Notably, our methods match the bespoke model CHMM despite only inferring worker ability parameters from the 191 items with corresponding gold in the PICO dataset, rather than all 4,800 that CHMM uses.
PDMRR, while not reaching PSR's performance, also outperforms baselines in bounding boxes while having mixed results in the other two experiments.

Overall, the results clearly show a general advantage from mixing-and-matching labels across different annotators. In particular, partitioning before selection (i.e., selecting one worker's label for each cluster) performs better than direct selection (i.e., selecting all of one worker's labels for the entire multi-object item). PSR achieves state-of-art performance across very diverse datasets.

The clustering-based partitioning approach we use shows much more promise than naive partitioning, such as at the token level (TMV). Partitioning may be further improved to attain better results, as indicated by the PDMRR-ORC and PSR-ORC performance.

\subsection{Performance of More Complex MAS vs.\ Simpler MADD}
\label{sec:masvsmadd}

MADD (Section~\ref{method:madd}) bridges a conceptual, practical, and empirical gap between models from prior literature on simple annotations vs.\ MAS (Section~\ref{method:mas}). The main difference between the MAS and MADD models is in their assumptions about the relationships between collected annotations and ground truth. MADD treats one of the given annotations as the best possible annotation, whereas MAS assumes all given annotations are deviations from some unseen latent best possible annotation. 

We find that MAS outperforms MADD in the majority of tasks, and thereby suggests the benefit of MAS's additional complexity when modeling complex annotations that can exist on an infinite space, one of the main differences from simple annotations.

\section{Discussion and Future Work}
\label{sec:future-work}

\subsection{Overall Takeaways}

The main contribution this article is based on is a general method for aggregating diverse types of complex annotations using distance functions. Almost all distance-based methods show a very substantial improvement on complex tasks vs.\ collecting only a single label (RU baseline). While this may not be surprising based on prior aggregation literature, it is a point worth emphasizing for requesters of complex annotations, where collection of multiple annotations per item is relatively rare. Of the aggregation methods tested, our MAS model is particularly effective in inferring weightings based on annotator reliability. As shown in the novel work here, this method provides modest improvement over unweighted aggregation on average, and significant improvement when there are at least five annotators per item and there is an even distribution of worker error or larger fraction of low-error workers than high-error ones. In addition, for multi-object annotations such as bounding boxes, we show that our PSR method provides a significant boost in accuracy. Finally, the general technique we describe for semi-supervised learning allows provides more consistent performance and improves over the method in our initial work \cite{braylan2020modeling}.













\subsection{Assessing Alternative Distance Functions}

The aggregation models presented clearly require having some distance function, and intuition suggests that ``better'' distance functions should yield better performance (e.g., aligning choice of distance function with the evaluation metric being optimized). As discussed in Section~\ref{sec:vs-baselines}, for the Translations experiments the HRRASA \cite{li2020crowd} method is able to benefit from a distance function that uses rich linguistic information from various natural language representation models, while otherwise functioning very similarly to our methods. It would be worth exploring the benefits of adding richer relevant information to distance functions more generally across these experiments.

It would also be interesting to investigate interactions between choice of distance function and MAS likelihood function. We have used normal likelihood (Equation \ref{eq:likelihood}), corresponding to square loss, but many other kernel functions could be explored \cite{lv2009positional}. With transformations of the data or the likelihood function, when appropriate, the MAS model should be capable of improvement akin to other regression models.

\subsection{Exploring Alternative Decompositions}

The general idea of aggregation via decomposition is to decompose a complex annotation type into lower-level primitives, which may be easier to operate upon (independently) than the original complex annotation. In this work, we achieve generality by utilizing standard object serialization \cite{haverlock1998object}, assuming  any complex annotation is implemented in software as a data structure which can be automatically decomposed into its constituent primitives. This means that the underlying representation used depends on a design choice by the programmer in how the complex annotation was encoded as a data structure in software. We have assumed a representation in terms of numeric primitives, but this may or may not align with the representation chosen by the programmer.  In our case, we are the programmer and so control the representation, but this may not always be the case, and other use cases for the collected annotations may drive other representations, potentially requiring a conversion between alternative underlying representations.   

While \citeA{parameswaran2016optimizing} suggested decomposing bounding boxes into individual pixels, we noted that pixel-level aggregation of overlapping bounding boxes typically yields non-rectangular regions, which would require further post-processing to actually induce the desired, aggregated boxes. In contrast, our approach decomposed bounding boxes into upper-left and lower-right numeric vertices such that aggregation produces two other vertices that still define a valid bounding box. \citeA{nguyen2017aggregating} similarly decompose text sequences into individual tokens, whereas we assume left and right positional indices. 

Such examples suggest flexibility in choice of representation and corresponding aggregation operators that might apply to that representation.
A given complex label type might be decomposed into alternative, equivalent primitive representations, but despite this equivalence, different representations permit different aggregation operators, which in turn yield different results. Consequently, exploring alternative primitive representations to best implement this decomposition-based aggregation for different complex annotation types remains an interesting area for future work.

\subsection{Limitations}


\subsubsection{Multi-modal Annotation Spaces}

One idea for future work is to extend MAS to support complex tasks without assuming the annotation space is isotropic and unimodal (Section \ref{formula:mas}). This could extend MAS beyond {\em objective} tasks to also support {\em subjective} tasks \cite{tian2012learning}, which permit a space of wider and more uneven valid responses. For example, subjective tasks may have very different valid responses pertaining to differing schools of thought, annotator preferences, or ambiguity in the task objectives.
Tasks for recognizing sentiment, hate speech, sarcasm, humor, poetry, or political discourse are several such examples \cite{paun2022statistical}.
Even for tasks like translation, there is subjectivity in that multiple answers could all be considered correct.
While a unimodal space assumes a single correct ground truth, allowing a model to identify multiple clusters of annotations could better support tasks with such subjective responses.

\subsubsection{Selection but no Merge or Partition for More Complex Data}

Several complex annotation types are only supported by our selection-based approach (i.e., do not support merge or partition). For example, free text or tree structures are not decomposable into primitives which can simply be merged, nor is it straightforward to treat such labels as multiple objects that can be partitioned through clustering and recombined with a union operator. Future work might also consider various other complex label types which our methods can be extended to, such as semantic segmentation or image polygons. 

\section{Conclusion}


Our work provides a unified, general framework for modeling and aggregating annotations across diverse, complex annotation tasks. We have proposed distance-based aggregation models for complex annotations beyond binary, categorical, or ordinal variables. Our Multidimensional Annotation Scaling (MAS) model and related methods bypass the challenge of having to define task-specific probabilistic models for each new type of complex annotation by modeling annotation distances, which can often be easily induced from existing evaluation metrics. Such distance-based methods can be used with little alteration across a wide variety of tasks, including translation, text sequence annotation, bounding boxes, keypoints, ranked lists, syntax trees, as well as numerical and categorical labeling.

In this article, we investigate three further research questions beyond our prior work. First, how do properties of a complex annotation dataset affect aggregation performance? Second, how should task owners navigate these many choices to make the most out of their collected complex annotations? Finally, how can future researchers better test that an aggregation model for complex annotations is performing in line with expectations? To understand how various factors impact accuracy and to inform model selection, we conduct large-scale simulation studies and broad experiments on real, complex datasets. Regarding testing, we introduce the concept of “unit tests” for aggregation models and present a suite of tests to ensure that a model is not mis-specified and exhibits expected behavior.

Beyond investigating these research questions above, we discuss the foundational concept and nature of annotation complexity, present a new aggregation model as a conceptual bridge between traditional models and our own, and contribute a new general semi-supervised learning method for complex label aggregation that outperforms prior work. Ours is the only semi-supervised aggregation method we know of for complex annotations. We also conduct the most extensive benchmarking study across both simple and complex annotation tasks that we are aware of. 


We find that distance-based models perform as well or better than task-specific bespoke models. We also show how inability to accurately infer annotator reliability can impair weighted models vs.\ unweighted models, and that weighted models fare best when there are at least five annotations per item and the worker distribution is either highly diverse in skill or at least not dominated by lower-skill workers. The fact that weighted models tend to perform best on the real datasets reflects that these attributes of the worker distribution are typical in practice.

Looking ahead, human labels remain the fuel of supervised learning. As we seek to grow AI capabilities to accomplish ever-more complex modeling and prediction tasks, we will need quality annotations for those tasks \cite{aroyo2022data}. We can expect more varied performance across annotators as annotation task difficulty increases, and as diverse crowd-sourced annotators are engaged to tackle ever-more challenging annotation tasks.
With the extensions to our research explored in this article, we provide a robust, general method for  automatically and effectively aggregating multiple complex annotations across diverse annotation task types.

\section*{Acknowledgments}

We thank the reviewers for their feedback, the crowd workers for the data they contributed, and Sooyong Lee and Abhishek Sridhar for their assistance. This work was supported in part by National Science Foundation grant No.\ 1253413, by Amazon, and by {\em Good Systems}\footnote{\url{http://goodsystems.utexas.edu/}},
a UT Austin Grand Challenge to develop responsible AI technologies. This work was completed prior to Omar Alonso joining Amazon. Any opinions, findings, and conclusions or recommendations expressed by the authors are entirely their own and do not represent those of the sponsoring agencies.



\vskip 0.2in
\bibliography{bibliography}
\bibliographystyle{theapa}

\end{document}